\documentclass[manuscript,screen, nonacm]{acmart}

\usepackage{makecell}
\usepackage{pifont}
\usepackage{adjustbox}
\usepackage{graphicx}
\usepackage{lscape}
\usepackage{subcaption}
\usepackage[many]{tcolorbox}
\usepackage{url}
\usepackage{hyperref}
\usepackage{booktabs}
\usepackage{todonotes}
\usepackage[capitalize,nameinlink]{cleveref}

\usepackage{xcolor}
\usepackage{pifont}
\usepackage{enumitem}

\newboolean{showcomments}
\setboolean{showcomments}{true}
\ifthenelse{\boolean{showcomments}}
 { \newcommand{\mynote}[2]{
      \fbox{\bfseries\sffamily\scriptsize#1}
        {\small$\blacktriangleright$\textsf{\emph{#2}}$\blacktriangleleft$}}}
        { \newcommand{\mynote}[2]{}}

\newcommand{\find}[1]{
\begin{tcolorbox}[leftrule=0.4mm,rightrule=0mm,toprule=0mm,bottomrule=0mm,left=0.0pt,right=0.0pt,top=0pt,bottom=0pt]
\em #1
\end{tcolorbox}
}
\AtBeginDocument{%
  }

\setcopyright{acmlicensed}
\copyrightyear{2025}
\acmYear{2025}
\acmDOI{XXXXXXX.XXXXXXX}
\acmISBN{TBD}

\begin{document}

\title{Beyond Real Faces: Synthetic Datasets Can Achieve Reliable Recognition Performance without Privacy Compromise}
\author{Paweł Borsukiewicz}
\email{pawel.borsukiewicz@uni.lu}
\affiliation{
  \institution{University of Luxembourg}
  \city{Luxembourg}
  \country{Luxembourg}
}

\author{Fadi Boutros}
\email{fadi.boutros@igd.fraunhofer.de}
\affiliation{
  \institution{Fraunhofer IGD}
  \city{Darmstadt}
  \country{Germany}
}

\author{Iyiola E. Olatunji}
\email{emmanuel.olatunji@uni.lu}
\affiliation{
  \institution{University of Luxembourg}
  \city{Luxembourg}
  \country{Luxembourg}
}

\author{Charles Beumier}
\email{charles.beumier@mil.be}
\affiliation{
  \institution{Royal Military Academy}
  \city{Brussels}
  \country{Belgium}
}

\author{Wendkûuni Christian OUEDRAOGO}
\email{wendkuuni.ouedraogo@uni.lu}
\affiliation{
  \institution{University of Luxembourg}
  \city{Luxembourg}
  \country{Luxembourg}
}

\author{Jacques Klein}
\email{jacques.klein@uni.lu}
\affiliation{
  \institution{University of Luxembourg}
  \city{Luxembourg}
  \country{Luxembourg}
}

\author{Tegawendé F. Bissyandé}
\email{tegawende.bissyande@uni.lu}
\affiliation{
  \institution{University of Luxembourg}
  \city{Luxembourg}
  \country{Luxembourg}
}

\authorsaddresses{%
\textbf{Authors' addresses:} \{pawel.borsukiewicz, emmanuel.olatunji, wendkuuni.ouedraogo, jacques.klein, tegawende.bissyande\}@uni.lu, fadi.boutros@igd.fraunhofer.de, charles.beumier@mil.be\newline
\textbf{Correspondence:} tegawende.bissyande@uni.lu}

\renewcommand{\shortauthors}{Borsukiewicz et al.}

\begin{abstract}
The deployment of facial recognition systems has created an ethical dilemma: achieving high accuracy requires massive datasets of real faces collected without consent, leading to dataset retractions and potential legal liabilities under regulations like GDPR. While synthetic facial data presents a promising privacy-preserving alternative, the field lacks comprehensive empirical evidence of its viability. This study addresses this critical gap through extensive evaluation of synthetic facial recognition datasets.

We present a systematic literature review identifying 25 synthetic facial recognition datasets (2018-2025), combined with rigorous experimental validation. Our methodology examines seven key requirements for privacy-preserving synthetic data: identity leakage prevention, intra-class variability, identity separability, dataset scale, ethical data sourcing, bias mitigation, and benchmark reliability. Through experiments involving over 10 million synthetic samples, extended by a comparison of results reported on five standard benchmarks, we provide the first comprehensive empirical assessment of synthetic data's capability to replace real datasets.


Best-performing synthetic datasets (VariFace, VIGFace) achieve recognition accuracies of 95.67\% and 94.91\% respectively, surpassing established real datasets including CASIA-WebFace (94.70\%). While those images remain private, publicly available alternatives Vec2Face (93.52\%) and CemiFace (93.22\%) come close behind.
Our findings reveal that they ensure proper intra-class variability while maintaining identity separability. Demographic bias analysis across four ethnic groups shows that, even though synthetic data inherits limited biases, it offers unprecedented control for bias mitigation through generation parameters.

These results establish synthetic facial data as a scientifically viable and ethically imperative alternative for facial recognition research. By eliminating consent violations while maintaining competitive performance, synthetic datasets provide a sustainable path forward for the field.
\end{abstract}





\maketitle

\section{Introduction}
Facial recognition technology processes billions of faces daily across security, commerce, and personal devices. Its capability largely depends on training with large-scale images, which enable the development of robust models.
Yet, this reliance on vast data collection poses fundamental privacy, ethical, and bias concerns, as the images used are often gathered without an individual's explicit consent~\cite{smith2022ethical}.
Consequently, researchers have retracted many well-established datasets~\cite{parkhi2015vgg, cao2018vggface2, guo2016ms} in response to these legal and ethical challenges~\cite{wang2024beyond, beltran2023privacy}.

To address this privacy compromise, synthetic facial recognition datasets have emerged as a promising alternative, garnering significant interest in recent years~\cite{boutros2023synthetic}. 
Early approaches often augmented real datasets with synthetic data~\cite{boutros2022sface, qiu2021synface}, yet yielded limited performance gains.
However, continual methodological improvements have enabled models to be trained entirely on synthetic data, achieving increasingly competitive benchmark results. Recent studies demonstrate that these fully synthetic models can now rival or even surpass~\cite{yeung2024variface} the performance of models trained on real data. This observed progress demonstrates that synthetic data can not only ensure reliable performance but also mitigate inherent ethical and legal concerns of real datasets, including demographic biases and label noise, while inherently preserving privacy.

Despite the inherent privacy benefits of synthetic data, its adoption in both research and industry remains limited, primarily due to the absence of standardized benchmarks and comprehensive validation. This work addresses this gap by presenting a systematic review and extensive evaluation of current synthetic facial recognition datasets. We critically assess the state-of-the-art to benchmark its current capabilities and highlight specific weaknesses that require further research.
The contributions of this work are as follows.
\begin{itemize}[topsep=0pt, partopsep=0pt, itemsep=0pt, parsep=0pt]
    \item Identification of the core ethical and legal limitations of real-face datasets and how synthetic datasets resolve them.
    \item A systematic literature review that identifies and benchmarks 25 synthetic facial recognition datasets and provides a historical analysis of the evolution of synthetic dataset generation methodologies.
    \item A novel framework and practical guidelines for synthetic facial recognition datasets, establishing key requirements and a standardized protocol for their evaluation.
     \item An in-depth evaluation of existing datasets, yielding novel quantitative insights on intra-class variance, identity separability, and identity leakage.
    \item A discussion of compelling future research directions, informed by our analysis of the current landscape.
\end{itemize}


The remainder of this paper is structured as follows.~\Cref{sec:motivation} details the critical limitations of real-face datasets that motivate this review.~\Cref{sec:literature-methodology} outlines our systematic methodology for literature collection, scope, and dataset selection.~\Cref{sec:key-requirements} introduces key requirements for privacy-preserving synthetic datasets and surveys their primary generation methods and relevant competitive benchmarks.~\Cref{sec:evaluation} provides a comprehensive evaluation of the field's progress against these requirements, highlighting both significant advancements and persistent challenges. Finally,~\Cref{sec:conclusion} concludes the paper and discusses avenues for future work.
\section{Motivation}
\label{sec:motivation}

Consider the foundation upon which modern facial recognition is built: massive datasets scraped from the web, enabling models that process billions of faces daily. This paradigm is now collapsing. As shown in~\Cref{tab:realdatabases}, a majority of the most prominent real-face datasets, including VGGFace~\cite{parkhi2015vgg}, VGGFace2~\cite{cao2018vggface2}, and MS-Celeb-1M~\cite{guo2016ms}, are no longer available from their original sources. This systemic retraction is not merely an inconvenience; it is a direct consequence of irreconcilable ethical, legal, and technical limitations that render the real-data paradigm unsustainable.
The core weaknesses of real-face datasets are threefold:
\begin{itemize}[topsep=0pt, partopsep=0pt, itemsep=0pt, parsep=0pt]
\item \textbf{Ethical and legal limitations:} The scale required for state-of-the-art performance has historically been achieved by scraping images from the web~\cite{parkhi2015vgg, cao2018vggface2, taigman2015web, wang2018imdb_face} \textbf{without explicit consent}. This practice is ethically questionable~\cite{smith2022ethical} and clashes directly with a tightening global regulatory landscape, including the EU's GDPR~\cite{GDPR2016a} and laws like Illinois' BIPA~\cite{BIPA}, which treat facial data as a protected biometric identifier~\cite{wang2024beyond, beltran2023privacy}.

\item \textbf{Inherent demographic biases:} Web-scraped datasets often reflect and amplify real-world demographic imbalances, leading to documented biases in gender~\cite{albiero2020does, buolamwini2018gender}, race~\cite{wang2019rfw}, and age. While synthetic datasets are not immune to bias~\cite{huber2024bias, whitney2024real}, their generation process offers a more direct and controllable path to mitigation~\cite{kortylewski2018can, kortylewski2019analyzing}.

\item \textbf{Pervasive label noise and volatility:} The mass collection of images introduces significant label noise that is costly to rectify~\cite{wang2018imdb_face}. Even benchmark datasets like LFW contain mislabeled images and inconsistent identity groupings~\cite{bae2023digiface}. Furthermore, distribution methods that rely on volatile web links lead to broken URLs and corrupted data integrity over time~\cite{wang2018imdb_face}.
\end{itemize}

\begin{table}[h!]
    \centering
    \caption{Real datasets used for facial recognition training. Status refers to the distribution by original authors at the time of writing. "Private" label refers to datasets that have never been made public.
    Note that the "Unavailable" datasets can still be in use as they are frequently redistributed online.}
    \begin{tabular}{cccccc}
\toprule
Dataset & \# Identities & \# Images & Images/ID & Venue & Status \\ \midrule
CelebFaces \cite{DBLP:conf/iccv/SunWT13} & 10k & 0.2M & 20 & ICCV'13 & Available \\
CASIA-WebFace \cite{yi2014casia} & 10k & 0.5M & 47 & arXiv'14 & Unavailable \\
DeepFace \cite{DBLP:conf/cvpr/TaigmanYRW14} &  4k & 4.4M & 1,100 & CVPR'14 & Private \\
Facebook \cite{taigman2015web} &  10M & 500M & 50 & CVPR'15 & Private \\
FaceNet \cite{schroff2015facenet} &  8M & 200M & 25 & CVPR'15 & Private \\
VGGFace \cite{parkhi2015vgg} & 2k & 2.6M & 1,000 & BMVC'15 & Unavailable \\ 
MS1M \cite{guo2016ms} & 0.1M & 10M & 100 & ECCV'16 & Unavailable\\ 
UMDFaces \cite{DBLP:conf/icb/BansalNCRC17} & 8k & 0.3M & 45 & IJCB'17 & Unavailable \\
MegaFace2 \cite{DBLP:conf/cvpr/NechK17} & 0.6M & 4.7M & 7 & CVPR'17 & Unavailable\\ 
VGGFace2 \cite{cao2018vggface2} & 9k & 3.3M & 363 & FG'18 & Unavailable\\ 
IMDb-Face \cite{wang2018imdb_face} & 59k & 1.7M & 29 & ECCV'18 & Available \\
MS1M-Glint \cite{deepglint_trillionpairs} & 87k & 3.9M & 44 & - & Unavailable\\
MS1MV2 \cite{deng2019iresnet50} & 85k & 5.8M & 68 & CVPR'19 & Available\\
MillionCelebs \cite{zhang2020global} &  0.6M & 18.8M & 30 &  CVPR'20 & Unavailable\\
WebFace260M \cite{an2021glint360k} & 4M & 260M & 65 & CVPR'21 & Unavailable\\
WebFace42M \cite{an2021glint360k} & 2M & 42M & 21 & CVPR'21 & Unavailable
\\ \bottomrule
    \end{tabular}
    \label{tab:realdatabases}
    \vspace{-3mm}
\end{table}

The crisis of data availability is formally recognized by leading conferences (e.g., CVPR~\cite{CVPRguidelines}, ICCV~\cite{ICCVguidelines}), which now caution against using retracted data. This creates a critical impediment to reproducible and responsible research. While prior surveys have discussed synthetic data as an alternative~\cite{kim202550, bauer2024comprehensive} or focused on taxonomies~\cite{boutros2023synthetic}, the field has reached a pivotal moment: recent synthetic datasets now achieve benchmark performance rivaling or surpassing models trained on real data like CASIA-WebFace~\cite{papantoniou2024arc2face, wu2024vec2face, yeung2024variface}. This rapid progress is, in large part, a response to the systemic failure of the real-data paradigm, motivates our central research question: \textit{\textbf{Can synthetic datasets now deliver the reliability and performance required to become the new standard for facial recognition?}} Our work provides this comprehensive analysis, establishing synthetic datasets not merely as an ethical alternative, but as a competitive and superior foundation for the future of facial recognition. 

\section{Methodology} 
\label{sec:literature-methodology}
This section details our approach to identifying relevant research on synthetic facial recognition datasets, including our selection criteria, search methodology, and analysis of publication trends.

\subsection{Scope and Inclusion Criteria}
\label{sec:scope-inclusion-criteria}
Recent advances in generative models~\cite{chakraborty2024GAN_survey, yang2023diffusion_survey} have led to a proliferation of facial image generation tools. However, many focus on identity transformation or enhancement e.g. via beautification~\cite{liu2023facechain, li2025diffusion, ulusan2025generating} rather than creating distinct synthetic identities for recognition tasks. To ensure that our analysis remains focused on synthetic data relevant to facial recognition and privacy research, we adopt the following inclusion criteria:

\begin{itemize}
    \item \textbf{Synthetic Identities.} The paper must introduce and evaluate at least one dataset comprising entirely synthetic facial images of artificial identities.
    \item \textbf{Privacy Preservation.} The work must explicitly address privacy preservation in facial recognition.
    \item \textbf{Recognition Utility.} The dataset is designed for training or evaluating general-purpose facial recognition models.
\end{itemize}
These criteria led us to exclude specialized datasets such as synthetic child faces~\cite{farooq2025childdiffusion, falkenberg2024child} and ISO/ICAO-compliant mugshots~\cite{borghi2025tono, di2024onot}. Although these studies address important practical challenges, they fall outside the general-purpose scope that this survey aims to characterize.

\subsection{Search Strategy}
We adopted a structured, multi-stage process to identify and validate relevant works.
\begin{itemize}
     \item \textbf{Database Search.} We queried major scientific databases (arXiv\footnote{https://arxiv.org/}, DBLP\footnote{https://dblp.org/}, ACM Digital Library\footnote{https://dl.acm.org/}, IEEE XPlore\footnote{https://ieeexplore.ieee.org/Xplore/home.jsp}, Elsevier Online Library\footnote{https://www.elsevier.com/}, Springer Link\footnote{https://link.springer.com/}, Wiley Online Library\footnote{https://onlinelibrary.wiley.com/}) using keyword combinations including: \textit{synthetic, biometric, privacy-preserving, artificial, face, facial, recognition, dataset, image, generation}. Google Scholar was used as a complementary source for cross-validation.
     
    \item  \textbf{Manual Screening.} Retrieved papers were manually examined to exclude works not satisfying our inclusion criteria as stated in~\Cref{sec:scope-inclusion-criteria}.
    \item \textbf{Reference Tracking:} We performed a snowballing procedure~\cite{wohlin2014snowballing} by examining references of relevant papers to identify additional studies missed by initial searches.
 \end{itemize}

\begin{figure}[htbp]
    \centering
    \begin{subfigure}[b]{0.45\textwidth}
        \centering               \includegraphics[width=1.0\textwidth]{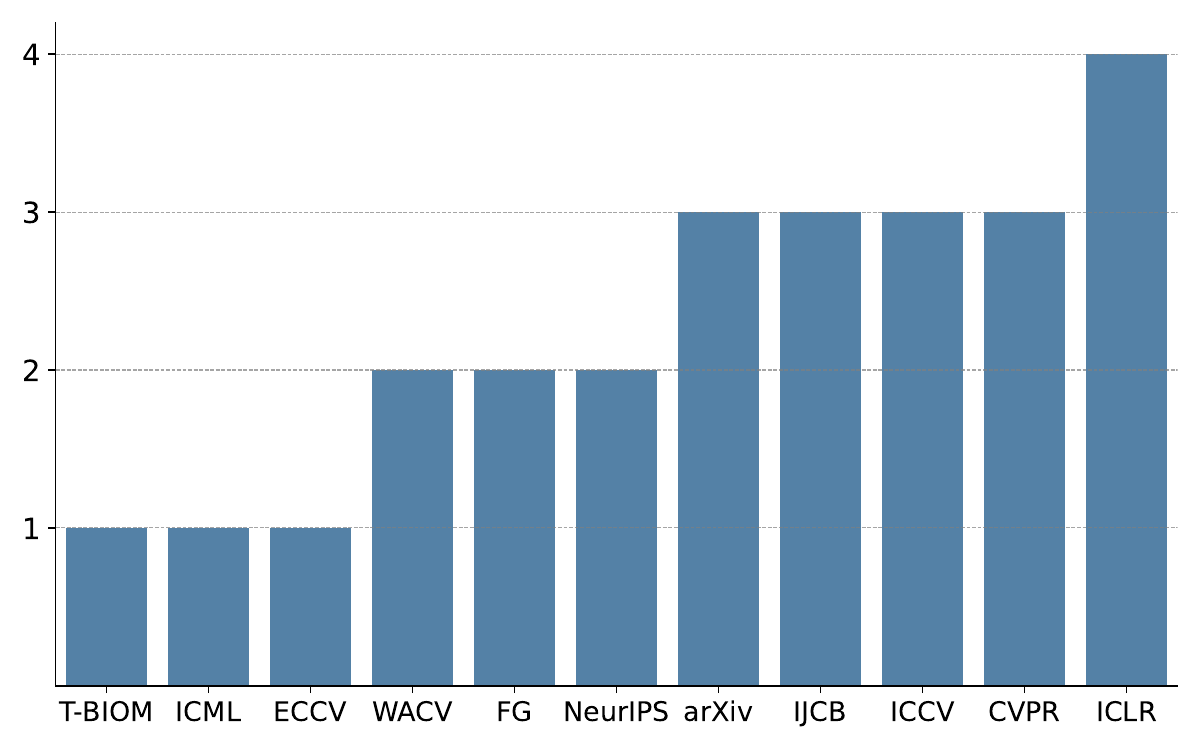} 
        \caption{Summary of publications by venues}
        \label{fig:Venues}
    \end{subfigure}
    \hfill
    \begin{subfigure}[b]{0.45\textwidth}
        \centering
        \includegraphics[width=1.0\textwidth]{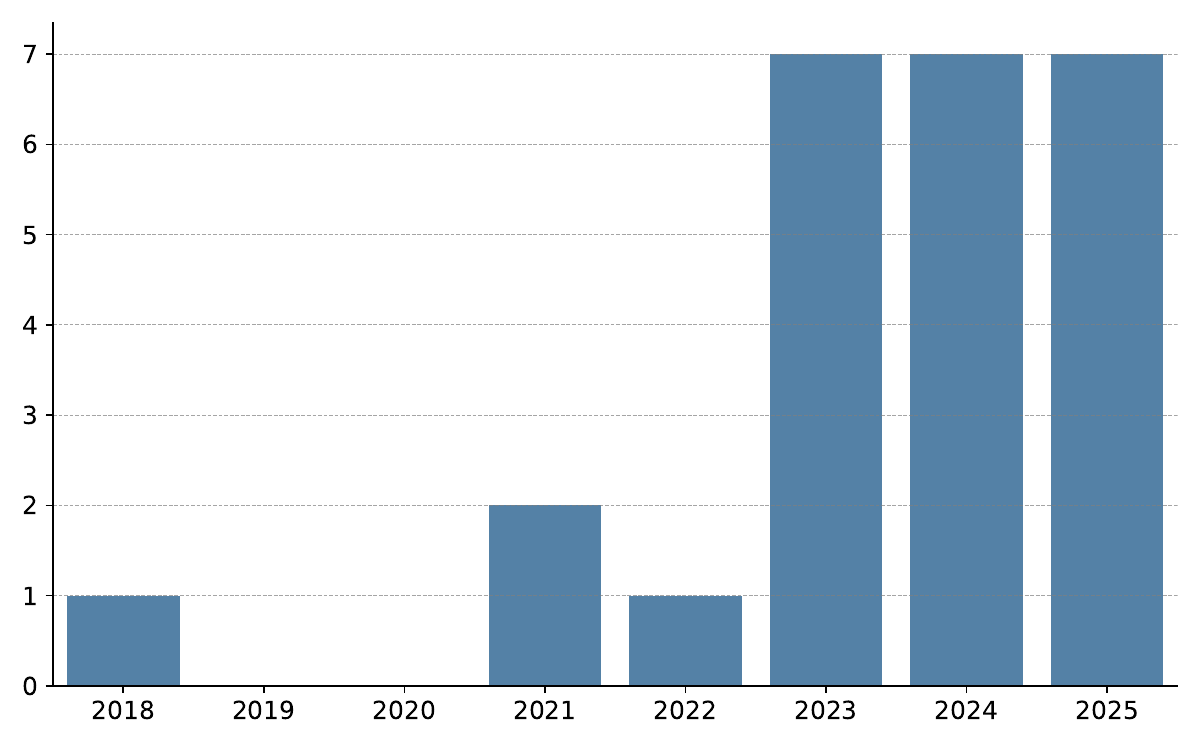} 
        \caption{Summary of publications by year}
        \label{fig:TimeOfPublication}
    \end{subfigure}
    \caption{Publications overview}
    \label{fig:twoside}
\end{figure}

\begin{figure}[h]
    \centering
    \includegraphics[width=1.0\textwidth]{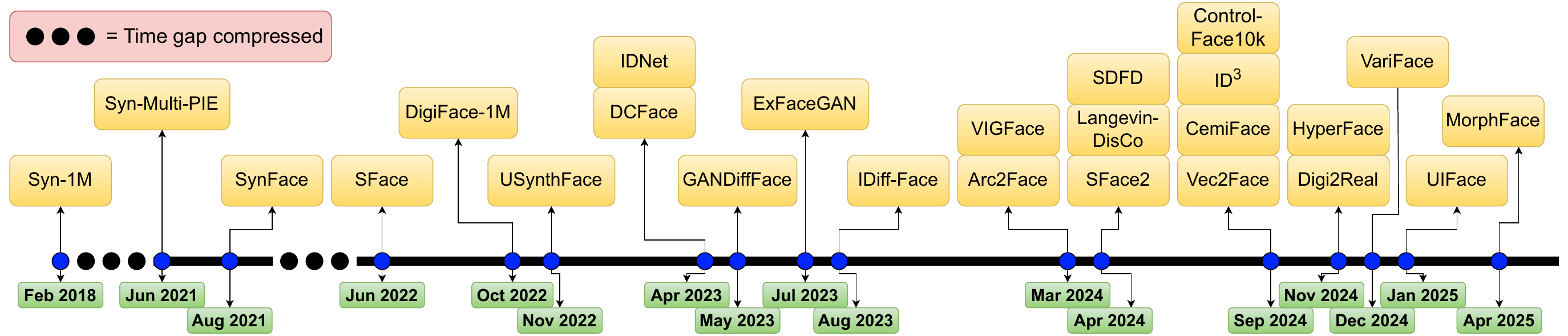} 
    \caption{Timeline of first public appearance of the identified synthetic datasets. Datasets released in the same month are aligned vertically.}
    \label{fig:Timeline}
\end{figure}

Following this systematic protocol, we identified 25 synthetic facial recognition datasets. All searches and screenings were conducted in June 2025 to ensure consistency and reproducibility. The analysis reveals several notable patterns: (i) publications are distributed across multiple venues without a dominant publication venue (\Cref{fig:Venues}), (ii) the publication frequency shows a clear upward trajectory in recent years (\Cref{fig:TimeOfPublication}) highlighting its relevance, (iii) many studies first appeared on arXiv before formal publication. \Cref{fig:Timeline} shows the first public appearance of each dataset, where works released within the same month share a vertical alignment.


\section{Synthetic Dataset Analysis Framework}
\label{sec:key-requirements}

To establish a comprehensive overview of the field, we categorize synthetic facial recognition datasets into two distinct groups based on accessibility: those publicly available for direct download (\Cref{tab:datasets_synth}) and those requiring special access or currently unavailable (\Cref{tab:frameworks_synth}). We classify datasets obtainable through structured access requests as publicly available, acknowledging that even controlled distribution represents a form of accessibility for research purposes.

To ensure comprehensive coverage, we contacted authors of all papers without publicly released datasets, though some correspondence proved unsuccessful due to inactive email addresses. Among responsive authors, common reasons for non-release included recent paper acceptance or ongoing peer review processes.



\begin{table}[h]
 \centering
 \small
 \caption{Publicly available synthetic facial recognition datasets. Access to datasets may require authors' approval. The availability and the size indicated at the time of writing may be subject to change. Provided resolution indicates the final preprocessed size of released images and can differ from the size obtained during generation.}
 \label{tab:datasets_synth}
\begin{tabular}{ccccc}
        \toprule
        Dataset & Venue & \makecell{Generation method} &  \makecell{ \#Images[Identities] \\ in the public dataset} & Resolution \\ \midrule
        Syn-Multi-PIE~\cite{colbois2021synmultipie} & IJCB'21 & GAN &  0.18M [10k]~\cite{SynMulti-PIE_datasetURL} & 112x112 \\ 
        
        SynFace~\cite{qiu2021synface} & ICCV'21 & GAN & 1M [10k]~\cite{SynFace_datasetURL} & 96x112 \\ 
        
        SFace~\cite{boutros2022sface} [\ding{70}] & IJCB'22 & GAN & 1.89M [10.5k]~\cite{SFace_datasetURL} & 112x112 \\ 
        
        DigiFace-1M~\cite{bae2023digiface} & WACV'23 & 3D model & 1.22M [110k]~\cite{DigiFace1M_datasetURL} & 112x112 \\ 
        
        IDNet~\cite{kolf2023idnet} & CVPR'23 & GAN & 1.06M  [10.5k]~\cite{IDNet_datasetURL} & 112x112 \\ 
        
        DCFace~\cite{kim2023dcface} & CVPR'23 & Diffusion & 0.55M [10k]~\cite{DCFace_datasetURL} & 112x112 \\ 

        GANDiffFace~\cite{melzi2023gandiffface} & ICCV'23 & GAN + Diffusion & 0.54M [10.1k]~\cite{GANDiffFace_datasetURL} & 512x512 \\ 

        IDiff-Face~\cite{boutros2023idiff} [\ding{70}] & ICCV'23 & Diffusion & 0.55M [10.5k]~\cite{IDiff-Face_datasetURL} & 112x112 \\ 
        
        SDFD~\cite{baltsou2024sdfd} & FG'24 & Diffusion & 0.01M [10k]~\cite{SDFD_datasetURL} & 768x768 \\ 
        
        SFace2~\cite{boutros2024sface2} [\ding{70}] & T-BIOM'24 & GAN & 1.05M [10.6k]~\cite{SFace_datasetURL} & 112x112 \\ 

        CemiFace~\cite{sun2024cemiface} & NeurIPS'24 & Diffusion & 0.55M [10k]~\cite{DCFace_datasetURL} & 112x112 \\ 
        
        Langevin-DisCo~\cite{geissbuhler2024Langevin} & ICML'25 & GAN &  1.95M [30k]~\cite{Langevin_datasetURL} & 112x112 \\ 
        
        Vec2Face~\cite{wu2024vec2face} & ICLR'25 & GAN & 15M[300k]~\cite{Vec2Face_datasetURL} & 112x112 \\ 
        
        HyperFace~\cite{shahreza2024hyperface} & ICLR'25 & GAN & 3.2M [50k] ~\cite{HyperFace_datasetURL} & 112x112 \\ 
        
        ControlFace10k~\cite{nzalasse2025sig} & ICPR'25 & Diffusion & 0.01M [3.3k]~\cite{ControlFace10K_datasetURL} & 512x512 \\ 
        
        Digi2Real-20K~\cite{george2025digi2real} & WACV'25 & 3D model + Diffusion & 0.4M[20k]~\cite{Digi2Real_datasetURL} & 112x112 \\ \bottomrule
\end{tabular}

[\ding{70}] Requires acceptance by the author through a structured process.
\end{table}

\begin{table}[h]
\centering
\small
 \caption{Datasets that were not made publicly available. Dataset and code availability indicated at the time of writing may be subject to change.}
\label{tab:frameworks_synth}
\begin{tabular}{cccc}
  \toprule
  Dataset & Venue & \makecell{Generation method} & \makecell{Code availability}\\
  \midrule

  Syn-1M~\cite{kortylewski2018syn1m} & arXiv'18 & 3D model & No \\
  
  USynthFace~\cite{boutros2023unsupervised} & FG'23 & GAN & Yes~\cite{USynthFace_datasetURL} \\
    
  ExFaceGAN~\cite{boutros2023exfacegan} & IJCB'23 & GAN & Yes~\cite{ExFaceGAN_datasetURL} \\

  Arc2Face~\cite{papantoniou2024arc2face} & ECCV'24 & Diffusion & Yes~\cite{Arc2Face_datasetURL} \\

  VIGFace~\cite{kim2024vigface} & arXiv'24 & Diffusion & No~\cite{VIGFace_datasetURL} \\

  ID$^3$~\cite{li2024id3} & NeurIPS'24 & Diffusion & No~\cite{ID3_datasetURL}\\

  VariFace~\cite{yeung2024variface} & arXiv'24 & Diffusion & No
  \\

  UIFace~\cite{lin2025uiface} & ICLR'25 & Diffusion & Yes~\cite{morphfaceANDuiface_datasetURL} \\

  MorphFace~\cite{mi2025morphFace} [\ding{70}]
  & CVPR'25 & Diffusion + 3D model & No~\cite{morphfaceANDuiface_datasetURL} \\
  
  \bottomrule
\end{tabular}

[\ding{70}] We were able to obtain the dataset through direct request to the authors.
\end{table}

\subsection{Requirements for Viable Synthetic Data}

To evaluate the viability of synthetic facial data as an alternative to real images, we define seven key requirements that extend beyond benchmark accuracy. These criteria provide a comprehensive framework for assessing both the advantages and current limitations of synthetic datasets in privacy-preserving facial recognition.


Specifically, for synthetic data to serve as a practical replacement for web-scraped repositories while maintaining privacy, we propose the following requirements:
\begin{itemize}
    \item \textbf{R1: Identity Leakage Prevention} - Newly created synthetic identities must be visually distinct from their training data sources to address legal and ethical concerns~\cite{wang2024beyond, beltran2023privacy} and prevent identity leakage~\cite{tinsley2021face}. Similarity scores of the synthetic images when compared with top samples from the training dataset should not clearly identify specific real identities, and original images should not be reverse-engineerable from synthetic samples.

    \item \textbf{R2: High Intra-class Variability} - Robust facial recognition models requires natural variation in external factors such as pose and lighting that could influence the style of each individual~\cite{mi2025morphFace}. Therefore, synthetic samples must exhibit sufficient yet natural diversity to properly challenge models during training~\cite{sun2024cemiface}.
    \item \textbf{R3: Identity Separability} - While maintaining intra-class diversity, identities must remain sufficiently distinct. That is, the variability of images cannot exceed a point when the images are so diverse that they do not belong to the same identity. Initial identity representations should be well-separated~\cite{shahreza2024hyperface}, with mated and non-mated comparisons following distributions observed in real datasets~\cite{melzi2023gandiffface}.

    \item \textbf{R4: High Image Count} - State-of-the-art performance requires large-scale data~\cite{masi2016we}. Therefore, synthetic generation frameworks should be capable of producing numerous distinct identities with multiple samples per identity.
    \item \textbf{R5: Ethical Data Sourcing} - To fully address ethical concerns, methods should either avoid real training data entirely or use exclusively consented data. While the latter is currently more feasible, real-data-independent methods represent the ideal future direction.
    \item \textbf{R6 - Bias Mitigation} - Since synthetic identity generation provides greater control over input and output data, one should strive to address demographic biases~\cite{perera2023analyzing} based on factors such as age, gender, ethnicity. Providing more balanced distribution, higher accuracy scores across typically underrepresented groups can be achieved and the probability of subsequent discrimination reduced. 
    \item \textbf{R7: Synthetic Benchmarks} - A complete synthetic pipeline requires evaluation without real data. While currently less reliable than established real benchmarks~\cite{qiu2021synface}, synthetic benchmark development remains crucial for privacy-preserving evaluation.    
\end{itemize}

\subsection{Generation methods and Trends}
\begin{figure}[h]
    \centering
    \includegraphics[width=.5\textwidth]{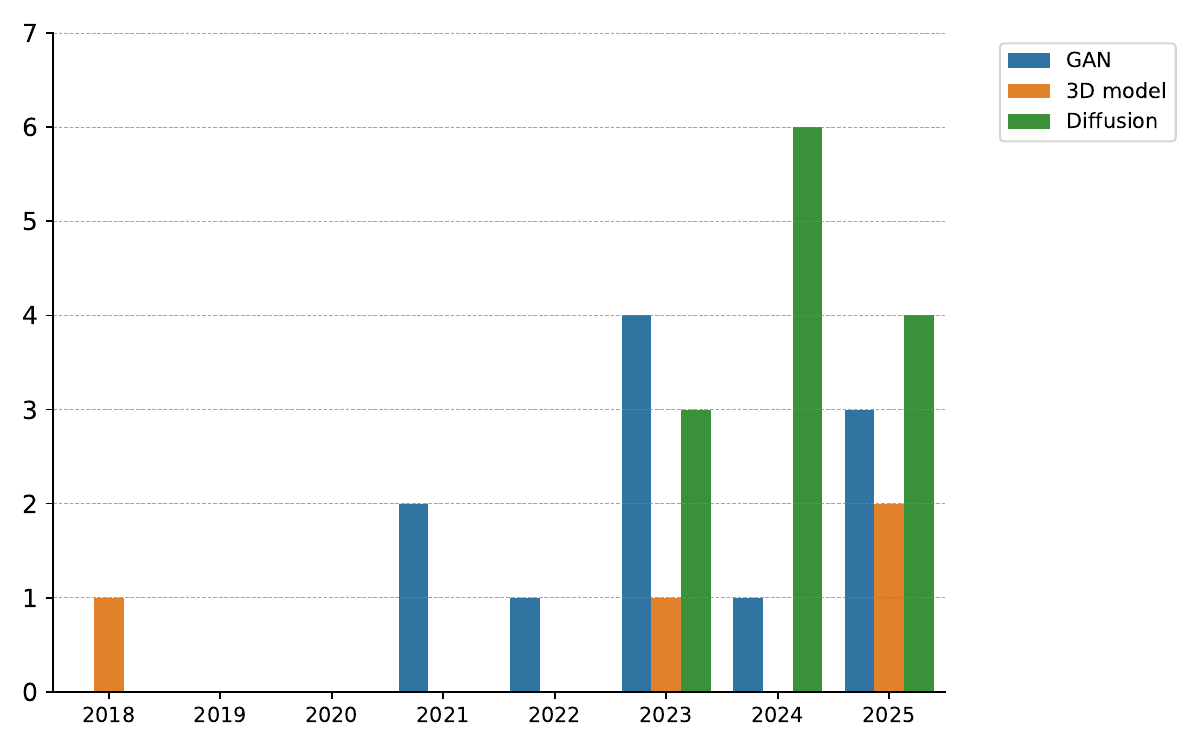} 
    \caption{Dataset generation method by year}
    \label{fig:MethodByYear}
\end{figure} 

Synthetic face generation primarily employs three approaches: Generative Adversarial Networks (GANs)~\cite{deng2020discofacegan, karras2020styleGAN2ADA, alaluf2022stylegan3, karras2020stylegan2}, Diffusion Models~\cite{ho2020ddpm, ruiz2023dreambooth, rombach2022high}, and 3D models~\cite{wood2021fake, feng2021learning}. Our analysis of method adoption over time (\Cref{fig:MethodByYear}) reveals a clear trend: a shift from GANs toward diffusion models in recent years. In the following, we examine each of these approaches in detail.

\noindent \textbf{Generative Adversarial Networks (GANs)} 
employ two neural networks: a generator that synthesizes images and a discriminator that distinguishes real from generated samples. This adversarial training process drives quality improvement, with the generator learning to produce increasingly realistic images. GANs commonly incorporate random augmentation~\cite{cubuk2020randaugment} techniques to enhance accuracy~\cite{boutros2023exfacegan, bae2023digiface, kolf2023idnet, boutros2023unsupervised}. 
Notable GAN methods include DiscoFaceGAN~\cite{deng2020discofacegan}, which generates synthetic identities with disentangled latent representation by incorporating 3D priors and contrastive losses into its learning procedure. It provides a control over pose, illumination and expression. Another line of work is StyleGAN~\cite{karras2019stylegan}, a style-based generator where generated samples are adjusted to a given style. Building upon this work, several variants exist. For instance, StyleGAN2~\cite{karras2020stylegan2} improved the quality of synthesized images by introducing adjustments resulting in smoother mappings and redesigned normalization to remove artifacts. StyleGAN2-ADA~\cite{karras2020styleGAN2ADA} introduced adaptive discriminator augmentation (ADA) technique to stabilize learning procedure when the training data is scarce. However, these methods suffer from over-reliance on absolute pixel positions. 
StyleGAN3~\cite{karras2021styleGAN3} addressed this over-reliance issue by changing internal representations. Applied modification boost rotation and translation, allowing for generation of higher quality samples.

\noindent \textbf{Diffusion models} are parameterized Markov chains~\cite{ho2020ddpm} that generate images through a gradual denoising process.
During training, they learn to reverse Gaussian noise applied to images, typically guided by text prompts. Key advances include Latent Diffusion Models (LDM)~\cite{rombach2022high}, which operate in a compressed latent space for improved efficiency. This two-stage approach first encodes images into a latent representation, then trains a denoising model on these representations, maintaining image quality while significantly reducing computational requirements.
DreamBooth~\cite{ruiz2023dreambooth}, which fine-tunes pre-trained diffusion models to generate images of specific identities from just a few reference samples, enabling strong identity preservation.
3D models are the least popular method of synthetic dataset generation. 
A general face object is generated based on scan collection. Further, it is covered with elements such as hair, accessories or clothing based on predefined options.
As presented in~\cite{wood2021fake}, one can generate datasets that lead to state-of-the-art results in face parsing and landmark localization. 
Arguably, generation cost estimated in the study as \$7.2k per 100k images rendered makes the method expensive.
More recent work~\cite{mi2025morphFace} has been based on DECA 3D morphable face model (3DMM)~\cite{feng2021learning}, built on top of FLAME~\cite{li2017learning}, which is used to synthesize a 3D mesh of vertices to represent shape or expression as a part of facial geometry. DECA itself adds appearance descriptions in form of texture and illumination. Additionally, it allows style control via numerical parameters.


\subsection{Benchmark Competitions}
Emerging facial recognition competitions based on synthetic datasets are a direct consequence of the identified need for ethical, privacy-preserving approaches.

Synthetic Data for Face Recognition Competition (SDFR)~\cite{shahreza2024sdfr} has proposed 2 challenges. Primarily participants had to train a model with up to 1M synthetic images using iResNet50~\cite{deng2019iresnet50} backbone. Second task lifted the restrictions regarding dataset size and architecture used. However, in both categories competitors were not allowed to use any real data with labeled identities. For that reason, datasets trained on FFHQ~\cite{karras2019ffhq} were allowed, while the ones trained on CASIA-WebFace~\cite{yi2014casia} were prohibited. Submissions were evaluated on the set of 5 standard benchmark datasets: LFW~\cite{huang2008lfw}, CFP-FP~\cite{sengupta2016cfp_fp}, CPLFW~\cite{zheng2018cplfw}, AgeDB~\cite{moschoglou2017agedb}, and CALFW~\cite{zheng2017calfw}; extended by two IARPA Janus
Benchmarks -- IJB-B~\cite{whitelam2017ijbb} and IJB-C~\cite{maze2018ijbc}. All the participating teams have decided to at least partially utilize IDiffFace~\cite{boutros2023idiff} dataset, which was according to the literature the highest performing option adhering to challenge rules at the time. The winning  accuracies, observed on a dataset of 1.7M pictures, were only slightly superior to the results obtained by the original IDiffFace paper, where only 0.5M images were used for training.

Two editions of FRCSyn challenge  ~\cite{melzi2024frcsyn_challenge_1st, deandres2024frcsyn_challenge_2nd} and their subsequent FRCSyn-onGoing variants ~\cite{melzi2024frcsynOngoing1, deandres2025frcsynOngoing2} have addressed two issues -- bias and challenging conditions -- in facial recognition. The first competition was based only on synthetic data from DCFace~\cite{kim2023dcface} and GANDiffFace~\cite{melzi2023gandiffface}, and real data from CASIA-WebFace~\cite{yi2014casia} and FFHQ~\cite{karras2019ffhq}. 
Experiment results have shown that the best results were obtained for the mixture of synthetic and real data.
Contrary, second FRCSyn iteration introduced 3 subcategories: constrained (500k images limit) and unconstrained synthetic data, as well as combination of real (CASIA-WebFace~\cite{yi2014casia}) and constrained synthetic data (500k images limit). The results of the 2nd challenge have shown significant improvement over the 1st edition, to a degree that some of the teams have managed to outperform real datasets with synthetic data, stressing their remarkable potential.

\subsection{Synthetic Datasets Overview}

The datasets presented in this section were identified based on the paper selection criteria and scope. Being entirely synthetic, they were specifically designed to address privacy risks attributed to the use of facial recognition technology and can be leveraged for model training or evaluation.

\def\WidthImage{1.0\textwidth}
\def\WidthFigure{0.3\textwidth}

\begin{figure}[htbp]
    \centering
    \begin{subfigure}[b]{\WidthFigure}
        \centering
        \includegraphics[width=\WidthImage]{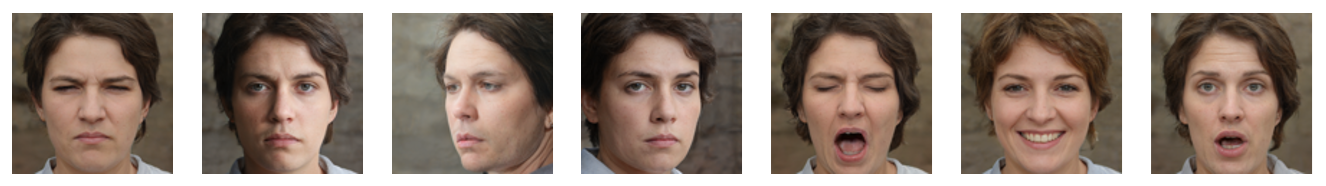} 
        \caption{ Syn-Multi-PIE }
        \label{fig:SynMultiPie-samples}
    \end{subfigure} 
    \hfill
    \begin{subfigure}[b]{\WidthFigure}
        \centering
        \includegraphics[width=\WidthImage]{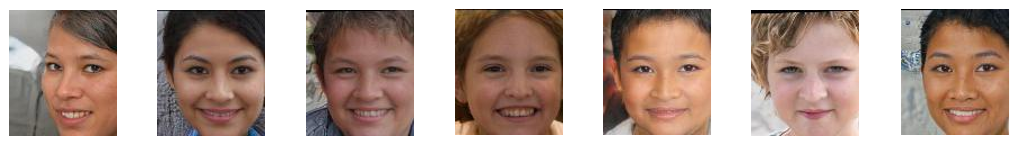} 
        \caption{ SynFace }
        \label{fig:SynFace-samples}
    \end{subfigure}
    \hfill
    \begin{subfigure}[b]{\WidthFigure}
        \centering
        \includegraphics[width=\WidthImage]{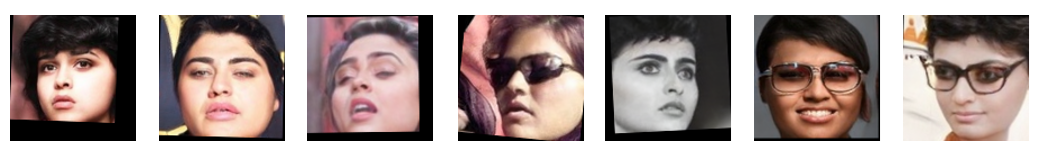} 
        \caption{ SFace }
        \label{fig:SFace-samples}
    \end{subfigure}
    \vfill
    \begin{subfigure}[b]{\WidthFigure}
        \centering
        \includegraphics[width=\WidthImage]{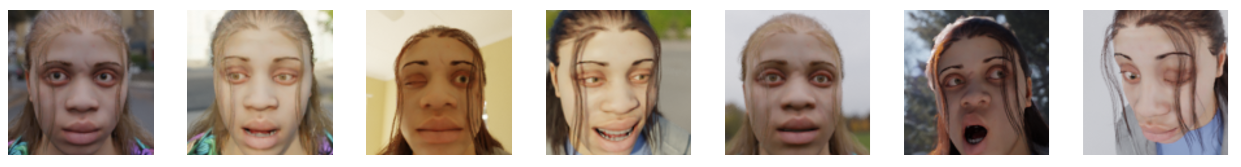} 
        \caption{ DigiFace-1M }
        \label{fig:Digi1M-samples}
    \end{subfigure}
    \hfill
    \begin{subfigure}[b]{\WidthFigure}
        \centering
        \includegraphics[width=\WidthImage]{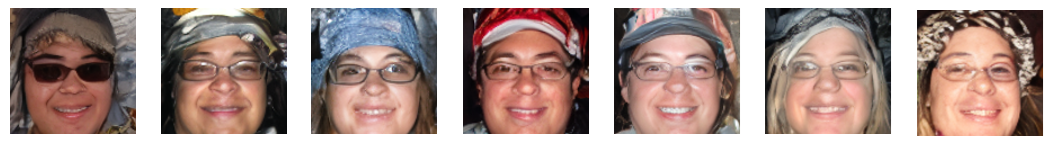} 
        \caption{ IDiff-Face }
        \label{fig:IDiffFace-samples}
    \end{subfigure}
    \hfill
    \begin{subfigure}[b]{\WidthFigure}
        \centering
        \includegraphics[width=\WidthImage]{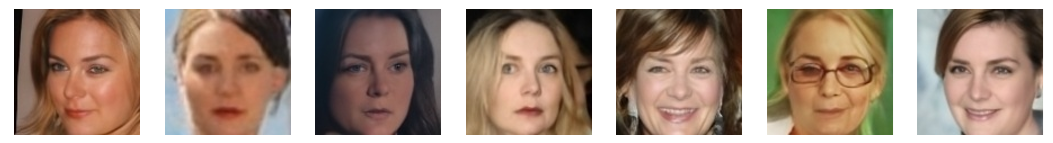} 
        \caption{ DCFace }
        \label{fig:DCFace-samples}
    \end{subfigure}
    \vfill
    \begin{subfigure}[b]{\WidthFigure}
        \centering
        \includegraphics[width=\WidthImage]{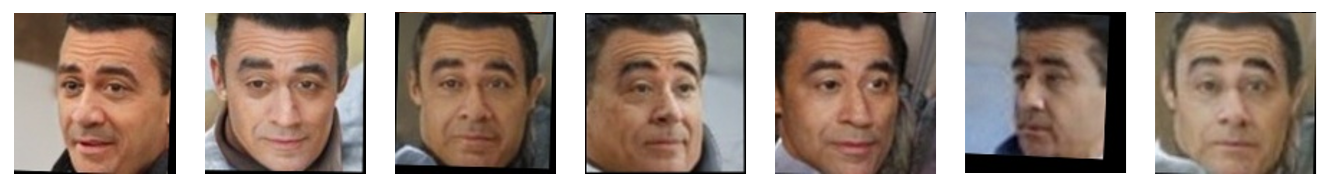} 
        \caption{ IDNet }
        \label{fig:IDNet-samples}
    \end{subfigure}
    \hfill
    \begin{subfigure}[b]{\WidthFigure}
         \centering
         \includegraphics[width=\WidthImage]{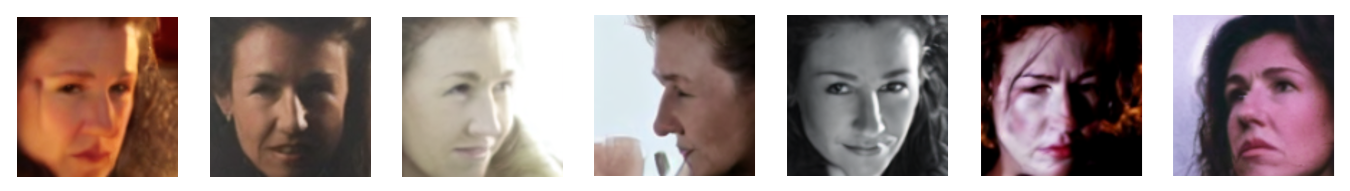} 
         \caption{ MorphFace }
         \label{fig:MorphFace-samples}
    \end{subfigure}
    \hfill
    \begin{subfigure}[b]{\WidthFigure}
        \centering
        \includegraphics[width=\WidthImage]{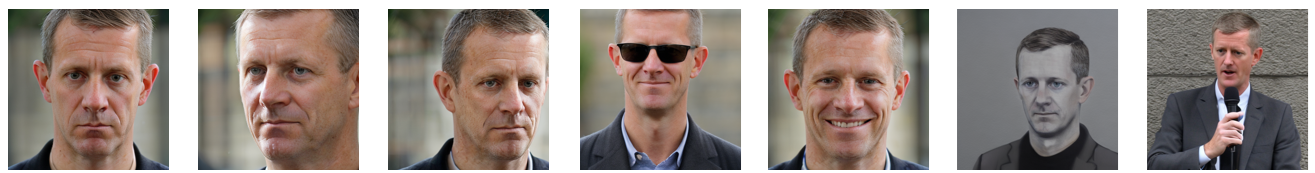} 
        \caption{ GANDiffFace }
        \label{fig:GANDiffFace-samples}
    \end{subfigure}
    \vfill
   \begin{subfigure}[b]{\WidthFigure}
        \centering
        \includegraphics[width=\WidthImage]{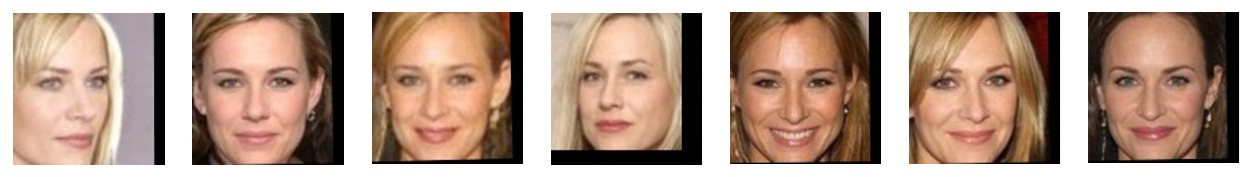} 
        \caption{ SFace2 }
        \label{fig:SFace2-samples}
    \end{subfigure}
    \hfill
    \begin{subfigure}[b]{\WidthFigure}
        \centering
        \includegraphics[width=\WidthImage]{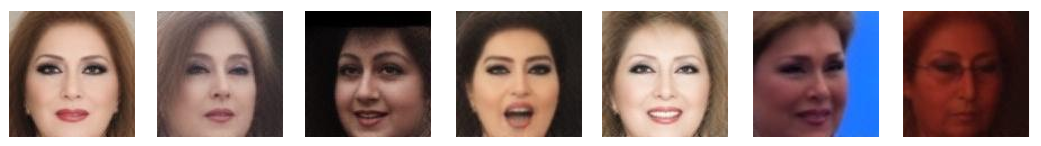} 
        \caption{ Vec2Face }
        \label{fig:Vec2Face-samples}
    \end{subfigure}
    \hfill
    \begin{subfigure}[b]{\WidthFigure}
        \centering
        \includegraphics[width=\WidthImage]{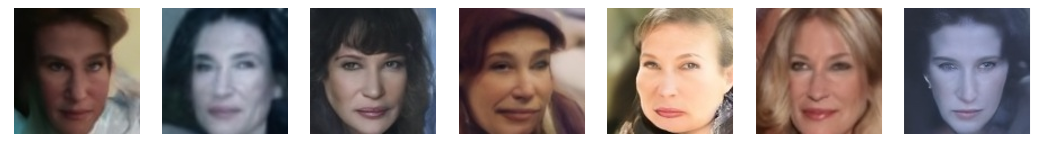} 
        \caption{ CemiFace }
        \label{fig:CemiFace-samples}
    \end{subfigure}
    \vfill
    \begin{subfigure}[b]{\WidthFigure}
        \centering
        \includegraphics[width=\WidthImage]{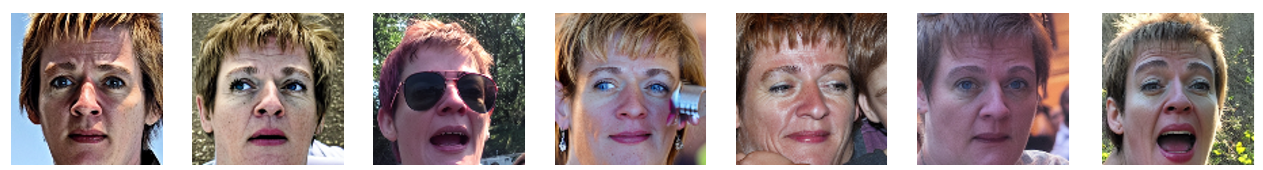} 
        \caption{ HyperFace }
        \label{fig:HyperFace-samples}
    \end{subfigure}
    \hfill
    \begin{subfigure}[b]{\WidthFigure}
         \centering
         \includegraphics[width=\WidthImage]{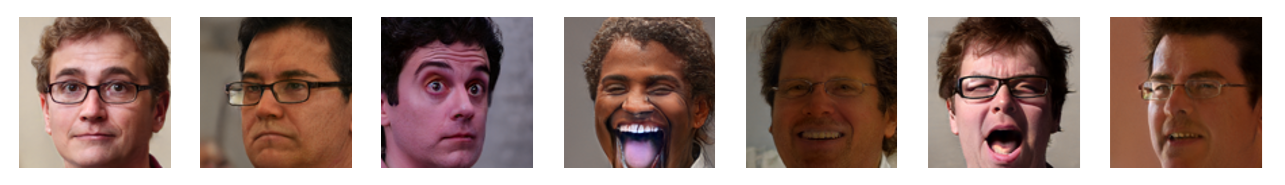} 
         \caption{ Langevin-DisCo }
         \label{fig:Langevin-samples}
     \end{subfigure}
    \hfill
    \begin{subfigure}[b]{\WidthFigure}
         \centering
         \includegraphics[width=\WidthImage]{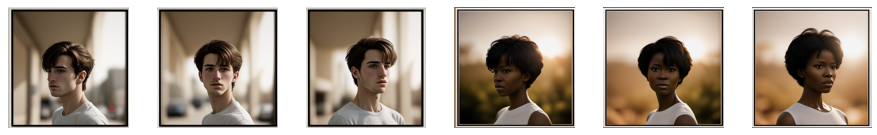} 
         \caption{ ControlFace10k (2 identities) }
         \label{fig:ControlFace10k-samples}
     \end{subfigure}
    \vfill
    \begin{subfigure}[b]{\WidthFigure}
         \centering
         \includegraphics[width=\WidthImage]{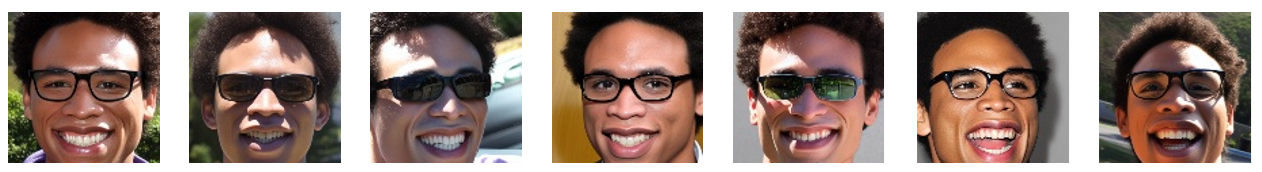} 
         \caption{ Digi2Real }
         \label{fig:Digi2Real-samples}
     \end{subfigure}
    \caption{Exemplary images of an identity across synthetic datasets}
    \label{fig:ImageExamplesAll}
\end{figure}



\noindent \textbf{Syn-1M}~\cite{kortylewski2018syn1m} was introduced in an early work to assess the effectiveness of training facial recognition models with partially and fully synthetic data. 3D Morphable Model~\cite{blanz2023morphable} is used to synthesize images based on the input data collected via 200 high-resolution face scans. While benchmark results were far from the ones obtained with real data, the paper has presented the research gap in the domain of fully synthetic facial recognition datasets.

\noindent \textbf{Syn-Multi-PIE}~\cite{colbois2021synmultipie} was created using StyleGAN2~\cite{karras2020stylegan2}. Its aim was to imitate the Multi-PIE dataset~\cite{gross2006multipie} in a synthetic manner to evaluate its viability for robustness assessment of facial recognition systems. Contrary to the majority of datasets presented in this paper, it was not intended to be used for training, but rather as a validation dataset. Even though intra-class variability of Syn-Multi-PIE was reported to be lower than for real-life data (\Cref{fig:SynMultiPie-samples}), it was found to be a promising benchmark dataset. 


\noindent \textbf{SynFace}~\cite{qiu2021synface} is a dataset created using face generator based on DiscoFaceGAN~\cite{deng2020discofacegan} that can synthesize a very limited amount of unique identities~\cite{bae2023digiface}. It was designed specifically to address the issues of low intra-class variance and domain gap between real and synthetic face datasets. The identity mixup technique, introduced to diversify photos generated for the same identities, boosted the performance of models trained on Synface. 
Nonetheless, it also significantly increased visual differences between images of the same identity (\Cref{fig:SynFace-samples}) to a degree that it is unlikely that the human reviewer would classify them as the same person. 
Further, the usage of the domain mixup technique to address the domain gap involves the use of real images in the training procedure, which does not eliminate underlying ethical concerns. 
While the dataset presented a remarkable 0.9998 accuracy on Syn-LFW, a synthetic benchmark created in the very same study, it had significantly worse performance on the real LFW~\cite{huang2008lfw}, with subsequent studies reporting even lower accuracy than indicated in the original study~\cite{yeung2024variface}. 


\noindent \textbf{SFace}~\cite{boutros2022sface} (including SFace-10, SFace-20, SFace-40, and SFace-60 variants) was generated with StyleGAN2-ADA~\cite{karras2020styleGAN2ADA}. The datasets of 10575 identities (\Cref{fig:SFace-samples}), equivalent to the number in CASIA-WebFace~\cite{yi2014casia} was used for GAN training. The base approach for SFace-60, containing 60 images per identity, reached much inferior accuracy when compared with CASIA-WebFace on the 5 benchmark datasets. However, the authors have introduced 2 additional training methods based on knowledge transfer and combined learning with a real dataset. While one could observe impressive improvement, confirmed by benchmark scores that would surpass many more recent methods, it comes at the cost of ethical concerns that the paper was supposed to address.


\noindent \textbf{DigiFace-1M}~\cite{bae2023digiface} is a dataset released by Microsoft and can be considered a successor of non-synthetic MS-Celeb-1M~\cite{guo2016ms}, introduced in 2016. With increasing concerns regarding ethics and morality of facial recognition~\cite{ftWhosUsing}, the website hosting the dataset was ultimately taken down in 2019~\cite{Exposing.ai}, even though the motive was not clearly acknowledged~\cite{peng2021mitigating}. Having addressed the issues of bias, label noise and ethical concerns, the new dataset was created using only data from 511 consensual face scans. Aforementioned scans served as a base for creation of texture library~\cite{wood2021fake}, which could be used to large amount of identities. Nonetheless, the proposed method has a serious drawback in terms of low image realism, producing unnatural faces (\Cref{fig:Digi1M-samples}). 

\noindent \textbf{USynthFace}~\cite{boutros2023unsupervised} study has proposed an unsupervised facial recognition method trained on the dataset composed of only one unlabeled image per identity, generated using DiscoFaceGAN~\cite{deng2020discofacegan}. By performing data augmentation on source pictures, positive pairs of images are created and the facial recognition model learns via Momentum Contrast
(MoCo)~\cite{he2020moco} method.


\noindent \textbf{IDNet}~\cite{kolf2023idnet} was constructed using an unconventional GAN approach involving not two but three players. In addition to the traditionally present generator and discriminator, in this particular case StyleGAN2-Ada~\cite{karras2020styleGAN2ADA}, the study has introduced the third entity -- an identity-oriented model -- to boost intra-class variance and class separability. Similarly to the previously described GAN-based solution, it is capable of generating highly realistic images (\Cref{fig:IDNet-samples}). 
Contrary to most of the datasets, the increase in its size usually does not have a positive influence on the benchmark results, with the best average scores obtained for around 0.2M images.


\noindent \textbf{DCFace}~\cite{kim2023dcface} is a dataset generated based on a framework laying emphasis on the possibility of generating a large number of unique and diverse subjects without label noise. Combining identity and style conditions for unconditional DDPM~\cite{ho2020ddpm} helps to preserve individual-related features, while style increases intra-class variations (\Cref{fig:DCFace-samples}). Even though the style variation has been praised in subsequent study~\cite{mi2025morphFace}, serious problems have also been identified. Firstly, identity leakage issues affecting this dataset have been raised~\cite{geissbuhler2024Langevin}. Subsequently, insufficient identity separability~\cite{mi2025morphFace} has been also observed. 



\noindent \textbf{GANDiffFace}~\cite{melzi2023gandiffface} is an outcome of a study proposing a combination of GANs and Diffusion models.
It leverages StyleGAN3~\cite{alaluf2022stylegan3} to generate identities and DreamBooth~\cite{ruiz2023dreambooth} to improve intra-class variation. As stated in the paper, one significant shortcoming of the method is its computational cost, which resulted in the creation of only 700 identities in the study. Nonetheless, the publicly available dataset was extended to contain around 10k identities. Due to unusually high image quality and resolution (\Cref{fig:GANDiffFace-samples}), namely 512x512 pixels, the dataset's digital size is much higher than on average.

\noindent \textbf{ExFaceGAN}~\cite{boutros2023exfacegan} has been generated using GANs with identity information disentanglement from latent spaces. By learning identity boundary, multiple synthetic images of an identity are created without attribute classifiers. A balance between intra-class diversity and inter-class separability has been addressed via a controllable sampling technique.

\noindent \textbf{IDiff-Face}~\cite{boutros2023idiff} was created using a pre-trained autoencoder~\cite{rombach2022high} to train DDPM~\cite{ho2020ddpm} for synthetic image generation.
Choice of identity-conditioned approach allowed to obtain high intra-class diversity, while preserving relatively realistic (\Cref{fig:IDiffFace-samples}) looks. 
The addition of Contextual Partial Dropout (CPD)~\cite{nitish2014dropout}, a probability-based solution for eliminating parts of the context embedding, as a mechanism preventing identity-context overfitting, had a positive impact on intra-class variation. Nevertheless, it was reported~\cite{mi2025morphFace} that the generated images lack style variation, which helps in facial recognition model generalization.

\noindent \textbf{VIGFace}~\cite{kim2024vigface}
was generated using a technique that preassigns synthetic identities in the feature space, which are later used by a diffusion model to create synthetic images. Clear division of the feature space for real and fake samples ensures proper identity separability. The paper also introduces a quite unique approach for intra-class diversity measurement using a normalized Face Image Quality Assessment (FIQA) metric based on CR-FIQA~\cite{boutros2023fiqa}. At the same time, it praises sample consistency, a metric promoting higher mated similarity scores. Consequently, it does not fulfill the aim of mimicking real distributions~\cite{li2024id3} and in fact could result in diversity reduction, negatively affecting training difficulty and subsequently weaker model performance.

\noindent \textbf{Arc2Face}~\cite{papantoniou2024arc2face} dataset was assembled leveraging a model allowing for creation of synthetic photo-realistic images from ArcFace embeddings, independently of pose, expression or scene requirements. The dataset was created using Stable Diffusion v1.5, guided by the CLIP~\cite{radford2021clip} text encoder, on an up-sampled version of WebFace42M~\cite{zhu2022webface260m} to ensure a high quantity of input data. An ID-conditioned model allows for generating images directly from embeddings without the need for a textual prompt. 


\noindent \textbf{SDFD}~\cite{baltsou2024sdfd} was introduced in a study that has identified a focus on demographic factors as a limitation of previous works in terms of diversity of generated images. To address this issue, authors have proposed a methodology allowing for the generation of images with a wider spectrum of facial attributes and accessories 
. Having compiled a list containing a wide range of attributes, it was used to generate prompts passed to the Stable Diffusion model responsible for image generation. Ultimately, an evaluation dataset composed of 1000 high-quality samples was created. Due to the significant limitation in the form of synthesis of only one image per identity, it is more applicable in the task of demographic attribute prediction rather than general facial recognition model evaluation.

 
\noindent \textbf{Langevin-DisCo} \& \textbf{Langevin-Dispersion}~\cite{geissbuhler2024Langevin} (\Cref{fig:Langevin-samples}) are datasets based on unique methods inspired by physical phenomena observed on soft particles. In the paper, the authors have introduced three new algorithms used on top of the StyleGAN~\cite{karras2019stylegan} face generator model - Langevin, Dispersion, and DisCo - to create large synthetic datasets. Langevin optimizes inter-class diversity, while Dispersion and DisCo focus on proper intra-class variation. 
Introduced datasets performed better than their predecessors on standard datasets as well as on RFW, suggesting lesser racial bias towards Caucasian individuals. The paper has also presented an identity leakage problem within the DCFace dataset.


\noindent \textbf{SFace2}~\cite{boutros2024sface2} has been designed to address the problem of low identity discrimination presented in its predecessor SFace~\cite{boutros2022sface}. 
It leverages a filtering mechanism, added to the previously presented StyleGAN2-Ada~\cite{karras2020styleGAN2ADA}-based image synthesis process. 
Applied adjustment does not affect negatively the image quality (\Cref{fig:SFace2-samples}).
Additionally, combining samples from SFace, which are characterized by larger intra-class variability, with newly generated SFace2 images, the authors introduced the SFace2+ variants of the datasets. It has been observed that as a result of the mixture, accuracy scores on benchmarks have exceeded base versions of both SFace and SFace2. The dataset has also been tested on different learning strategies, involving the usage of ethically questionable models pre-trained on real data, which allowed it to achieve scores comparable with CASIA-WebFace~\cite{yi2014casia}. 

\noindent \textbf{ControlFace10k}~\cite{nzalasse2025sig}, based on Synthetic Identity Generation (SIG) pipeline, is an evaluation dataset. By focusing on reducing potential demographic biases and external factors, it proposes 10k images of 3.3k identities. Using fine-tuned stable diffusion model with a prompt method based on diverse combinations of names from different countries, they are capable of creating idealized, hyper-realistic images (\Cref{fig:ControlFace10k-samples}). While the authors claim that SIG eliminates reliance on existing real datasets, the scope of data used to train underlying model raises serious concerns as it is not explicitly specified. Further, the dataset was not used as a benchmark to evaluate other real or synthetic datasets, putting into question its primary function in context of applicability in real life scenarios.

\noindent \textbf{ID$^3$}~\cite{li2024id3} synthesis method utilizes IDentity-preserving-yet-Diversified Diffusion model to synthesize images. Choice of 4-terms loss function and sampling algorithm ensures identity preservation and aims to preserve real-life data distribution at the same time satisfying variability, by incorporating attribute predictors, and identity-preservation requirements.  


\noindent \textbf{CemiFace}~\cite{sun2024cemiface} was introduced based on the notion of center-based semi-hard samples. The method incorporated a similarity controlling factor into the training loss of the denoising diffusion model. 
By generating only images with purposely reduced similarity, resulting in relatively high intra-class variance, one assures the difficulty level for facial recognition model training to be the most effective~\cite{kim2022adaface, schroff2015facenet}. 
While the method outperformed the other techniques at the time of publication, it produces quite distorted images (\Cref{fig:CemiFace-samples}).


\noindent \textbf{Vec2Face}~\cite{wu2024vec2face} 
is a dataset that balances the needs for high inter-class variation and intra-class separability with a large amount of identities. Its generation procedure involving face image reconstruction from a vector allows for flexibility in terms of choice of facial attributes (\Cref{fig:Vec2Face-samples}). By applying small perturbations to the input, one achieves proper variability of output samples. The synthesized HSFace10K dataset -- 500k images of 10k identities -- was reported to be the first dataset to surpass CASIA-WebFace~\cite{yi2014casia} on CALFW~\cite{zheng2017calfw}, IJBB~\cite{whitelam2017ijbb} and IJCC~\cite{maze2018ijbc} benchmarks. Additionally, the HSFace400k dataset version, consisting of 20M images of 400k identities, greatly surpassed the sizes offered by its predecessors.


\noindent \textbf{HyperFace}~\cite{shahreza2024hyperface} paper approaches synthetic dataset generation as a hypersphere packing problem. By analyzing multiple strategies, authors have been able to find an optimization to ensure high inter-class variation by the proper positioning of embeddings. Using specifically crafted embeddings, generative models can be used to synthesize images (\Cref{fig:HyperFace-samples}). In the context of this experiment, StyleGAN~\cite{karras2019stylegan} has been used for initialization and regularization.  Even though there is a certain degree of similarity between the closest original and synthetic images, identity leakage could not be definitively confirmed or denied. However, the problem was found to be the most prominent in child images.



 
\noindent \textbf{Digi2Real}~\cite{george2025digi2real} was synthesized leveraging a novel method to tackle the problem of low realism of the DigiFace-1M~\cite{bae2023digiface} dataset by applying a pre-trained Arc2Face model combined with CLIP shift. While there are still some imperfections, image quality is significantly higher (\Cref{fig:Digi2Real-samples}) than in the base dataset. Thanks to this change and focus on intra-class variation, an improvement on benchmark scores has been observed.

\noindent \textbf{VariFace}~\cite{yeung2024variface} 
was created using a two-stage diffusion-based method. In the first stage, a demographically balanced set of identities is generated. Subsequently, a diffusion model is trained to take into consideration race and gender attributes. The second stage ensures proper inter-class variation by assigning a training score representing deviation from the base sample. This approach eliminates reliance on text prompts or classifier labels. Outliers with too low similarity score are removed. 

\noindent \textbf{UIFace}~\cite{lin2025uiface} addresses poor face recognition performance of synthetic datasets attributed to low intra-class diversity caused by context overfitting. Using the diffusion model, two sampling contexts are introduced. First, identity context allows for identity preservation, however, it negatively affects intra-class variation. Second, empty context boosts the intra-class diversity, yet for random identities. To benefit from the strengths of both approaches, a two-stage sampling technique has been introduced. 


\noindent \textbf{MorphFace}~\cite{mi2025morphFace} is an answer to the limitation of previous works in the trade-off between image consistency and style diversity. By training a diffusion-based generator on DECA~\cite{feng2021learning} 3D morphable model renderings, authors have been able to craft identity-specific styles. Additionally, a statistical sampling method based on real-life distribution was used to balance inter-class and intra-class variation. The introduction of the context blending technique has allowed for greater consistency and style variation (\Cref{fig:MorphFace-samples}). Importantly, an internal study on identity leakage has shown a low relation with the underlying training dataset. 




\subsection{Evaluation Methodology and Performance}

Commonly, synthesized images are used as a training dataset for ResNet-50~\cite{he2016resnet} models to be evaluated on LFW~\cite{huang2008lfw}, CFP-FP~\cite{sengupta2016cfp_fp}, CPLFW~\cite{zheng2018cplfw}, AgeDB~\cite{moschoglou2017agedb}, and CALFW~\cite{zheng2017calfw} benchmark datasets using approximately 0.5M images of 10k identities to match CASIA-WebFace~\cite{yi2014casia} size. In recent years, we have observed an impressive improvement in the performance of models trained entirely on synthetic data. Even though, none of the proposed approaches have yet outperformed CASIA-WebFace in a fair competition (\Cref{tab:benchmark_500k}), we have observed a few datasets coming very close behind, suggesting that it might be just a matter of time before synthetic datasets achieve superior performance.


LFW~\cite{huang2008lfw} is an unconstrained face verification benchmark and contains 13,233 images of 5749 identities. The result on LFW is reported as verification accuracy following the "unrestricted with labeled outside data" protocol using the standard 6000 comparison pairs. AgeDB~\cite{moschoglou2017agedb} is an in-the-wild dataset for age-invariant face verification evaluation, containing 16,488 images of 568 identities. The results are reported as verification accuracy for AgeDB-30 (30 years age gap) as it is the most reported and challenging subset of AgeDB, following the evaluation protocol provided by AgeDB~\cite{moschoglou2017agedb}. Cross-age LFW (CALFW) benchmark~\cite{zheng2017calfw} is based on LFW with a focus on comparison pairs with an age gap, however, not as large as AgeDB-30. Age gap distribution of the CALFW is provided in~\cite{zheng2017calfw}. It contains 3000 genuine comparisons and the negative pairs are selected of the same gender and race to reduce the effect of attributes. The Cross-Pose LFW (CPLFW) benchmark~\cite{zheng2018cplfw}, is based on LFW with a focus on comparison pairs with pose differences. CPLFW contains 3000 genuine comparisons, while the negative pairs are selected of the same gender and race. Celebrities in Frontal-Profile in the Wild (CFP-FP)~\cite{sengupta2016cfp_fp} benchmark addresses the comparison between frontal and profile faces. CFP-FP evaluation contains 3500 genuine pairs and 3500 imposter pairs. The evaluation protocol of CALFW~\cite{zheng2017calfw}, CPFLFW~\cite{zheng2018cplfw}, and CFP-FP~\cite{sengupta2016cfp_fp}, reports the face recognition performance as verification accuracy, similar to the LFW~\cite{huang2008lfw} benchmark.

\begin{table}[h]
 \centering
 \caption{Benchmark comparison of the best results reported on evaluation datasets for ResNet-50 models trained on synthetic datasets variants of sizes closest to CASIA-WebFace. The best results are indicated by bold font, the second best by underlining. Data taken from original studies unless stated otherwise. The image and identity number are approximated.} 
\label{tab:benchmark_500k}
\begin{tabular}{ccccccccc}
  \toprule
  Dataset & \#Images & \#Identities & LFW & CFP-FP & CPLFW & AgeDB & CALFW & Average \\
  \midrule
  CASIA-WebFace~\cite{yi2014casia} [\ding{70}] & 494k & 10.5k & \textbf{0.9950} & \textbf{0.9536} & \textbf{0.9007} & \textbf{0.9487} & \textbf{0.9368} & \textbf{0.9470}\\
  \hline
  
  VariFace~\cite{yeung2024variface} & 500k & 10k & \underline{0.9948} & 0.9460 & \underline{0.8962} & \underline{0.9438} & 0.9305 & \underline{0.9411} \\

  MorphFace~\cite{mi2025morphFace} & 500k & 10k & 0.9925 & 0.9411 & 0.8873 & 0.9180 & 0.9273 & 0.9332 \\

  UIFace~\cite{lin2025uiface} & 500k & 10k & 0.9927 & 0.9429 & 0.8958 & 0.9095 & 0.9225 & 0.9327 \\

  VIGFace~\cite{kim2024vigface} & 500k & 10k & 0.9902 & \underline{0.9509} & 0.8772 & 0.9095 & 0.9000 & 0.9256 \\
  
  CemiFace~\cite{sun2024cemiface} & 500k & 10k & 0.9903 & 0.9106 & 0.8765 & 0.9133 & 0.9242 & 0.9230\\
  
  Vec2Face ~\cite{wu2024vec2face} & 500k & 10k & 0.9887 & 0.8897 & 0.8547 & 0.9312 & \underline{0.9357} & 0.9200\\

  Arc2Face~\cite{papantoniou2024arc2face} & 500k & 10k & 0.9881 & 0.9187 & 0.8516 & 0.9018 & 0.9263 & 0.9173  \\

  ID$^3$~\cite{li2024id3} & 500k & 10k & 0.9768 & 0.8684 & 0.8277 & 0.9100 & 0.9073 & 0.8980 \\

  DCFace~\cite{kim2023dcface} & 500k & 10k & 0.9855 & 0.8533 & 0.8262 & 0.8970 & 0.9160 & 0.8956\\

  HyperFace~\cite{shahreza2024hyperface} & 500k & 10k & 0.9850 & 0.8883 & 0.8423 & 0.8653 & 0.8940 & 0.8950 \\

  IDiff-Face~\cite{boutros2023idiff} & 500k & 10k & 0.9800 & 0.8547 & 0.8045 & 0.8643 & 0.9065 & 0.8820\\

  Digi2Real-20K~\cite{george2025digi2real} & 400k & 20k & 0.9858 & 0.8749 & 0.8265 & 0.8472 & 0.8722 & 0.8813\\

  Langevin-DisCo ~\cite{geissbuhler2024Langevin} & 650k & 10k & 0.9707 & 0.7956 & 0.7673 & 0.8338 & 0.8905 & 0.8516 \\
  
  DigiFace-1M~\cite{bae2023digiface} & 500k & 10k & 0.9540 & 0.8740 & 0.7887 & 0.7697 & 0.7862 & 0.8345\\
  
  ExFaceGAN~\cite{boutros2023exfacegan} & 500k & 10k & 0.9350 & 0.7384 & 0.7160 & 0.7892 & 0.8298 & 0.8017  \\ 

  SFace2~\cite{boutros2024sface2} & 635k & 10.5k & 0.9462 & 0.7624& 0.7218 & 0.7437 & 0.8157 & 0.7920\\
 
  IDNet~\cite{kolf2023idnet} & 528k & 10.5k & 0.9258 & 0.7540 & 0.7425 & 0.7353 & 0.7990 & 0.7913 \\
  
  USynthFace~\cite{boutros2023unsupervised} & 400k & 400k & 0.9223 & 0.7856 & 0.7203 & 0.7162 &  0.7705 & 0.7830\\
    
  Langevin-Dispersion~\cite{geissbuhler2024Langevin} & 650k & 10k & 0.9438 & 0.6551 & 0.6575 & 0.7730 & 0.8603 & 0.7779 \\
  
  SFace~\cite{boutros2022sface} & 423k & 10.5k & 0.9143 & 0.7310 & 0.7342 & 0.6987 & 0.7692 & 0.7695\\

  
  SynFace~\cite{qiu2021synface} [\ding{70}] & 500k & 10k & 0.8407 & 0.6337 & 0.6355 & 0.5910 & 0.6937 & 0.6789\\

  \bottomrule
\end{tabular}

[\ding{70}] Results obtained from reproduced experiments~\cite{yeung2024variface} due to data unavailability in original studies.


\end{table}

However, in terms of unrestricted dataset sizes, there have been reported cases when synthetic datasets achieved superior performance on some or even all test benchmarks (\Cref{tab:benchmark_unrestricted}). While one has to be aware of an unfair advantage lying in the bigger size of the training dataset, it is also important to keep in mind that, in terms of real-life datasets, researchers for years strived to outperform preceding datasets, focusing not only on data quality but also quantity.
While analyzing the data, one can also observe that the top results were obtained by VariFace, MorphFace, UIFace, VIGFace - solutions based on diffusion or their combination with 3D models. Meanwhile, Vec2Face, obtaining the best results among GAN-based methods, outperforms multiple diffusion-based datasets, yet lags behind the best-performing ones. Nevertheless, the presented ranking should not be interpreted as a straightforward show of superiority of diffusion models over GANs, but rather as a current standing in the race between those two technologies. 

\begin{table}[h]
 \centering
 \caption{Benchmark comparison of the best results reported on evaluation datasets for ResNet-50 models trained on synthetic datasets without size restrictions, compared with CASIA-WebFace. The best results are indicated by bold font, the second best by underlining. The number of images and identities refers to the approximation of the largest dataset referenced in the original paper. Data taken from original studies unless stated otherwise.} 
\label{tab:benchmark_unrestricted}
\begin{tabular}{ccccccccc}
  \toprule
  Dataset & \#Images & \#Identities & LFW & CFP-FP & CPLFW & AgeDB & CALFW & Average \\
  \midrule
  CASIA-WebFace~\cite{yi2014casia} [\ding{70}] & 0.49M & 10.5k & \underline{0.9950} & 0.9536 & 0.9007 & \underline{0.9487} & 0.9368 & 0.9470\\
  \midrule
  
  VariFace~\cite{yeung2024variface} & 6M & 60k & \textbf{0.9960} & \underline{0.9637} & \textbf{0.9205} & \textbf{0.9568} & \textbf{0.9467} & \textbf{0.9567} \\

  VIGFace~\cite{kim2024vigface} & 6M & 120k & 0.9948 & \textbf{0.9731} & \underline{0.9112} & 0.9382 & 0.9295 & \underline{0.9491} \\

  UIFace~\cite{lin2025uiface} & 1.5M & 30k & 0.9938 & 0.9596 & 0.9067 & 0.9328 & 0.9343 & 0.9454 \\

  MorphFace~\cite{mi2025morphFace} & 1.2M & 24k & 0.9935 & 0.9477 & 0.9007 & 0.9327 & 0.9340 & 0.9417 \\
     
  Vec2Face ~\cite{wu2024vec2face} & 20M & 400k & 0.9930 & 0.9154 & 0.8770 & 0.9445 & \underline{0.9458} & 0.9352 \\

  CemiFace~\cite{sun2024cemiface} & 1.2M & 60k & 0.9922 & 0.9284 & 0.8886 & 0.9213 & 0.9303 & 0.9322\\
  
  Arc2Face~\cite{papantoniou2024arc2face} & 1.2M & 60k &  0.9892 & 0.9458 & 0.8645 & 0.9245 & 0.9333 & 0.9314  \\

  HyperFace~\cite{shahreza2024hyperface} & 3.2M & 50k & 0.9827 & 0.9224 & 0.8560 & 0.9040 & 0.9148 & 0.9160 \\

  DCFace~\cite{kim2023dcface} & 1.2M & 60k & 0.9858 & 0.8861 & 0.8507 & 0.9097 & 0.9282 & 0.9121 \\

  Langevin-DisCo ~\cite{geissbuhler2024Langevin} & 3.25M & 50k & 0.9897 & 0.8377 & 0.8152 & 0.9332 & 0.9395 & 0.9031 \\

  ID$^3$~\cite{li2024id3} & 0.5M & 10k & 0.9768 & 0.8684 & 0.8277 & 0.9100 & 0.9073 & 0.8980 \\

  IDiff-Face~\cite{boutros2023idiff} & 0.5M & 10k & 0.9800 & 0.8547 & 0.8045 & 0.8643 & 0.9065 & 0.8820\\
  
  Digi2Real-20K~\cite{george2025digi2real} & 0.4M & 20k & 0.9858 & 0.8749 & 0.8265 & 0.8472 & 0.8722 & 0.8813\\
  
  DigiFace-1M~\cite{bae2023digiface} & 1.2M & 110k & 0.9582 & 0.8877 & 0.8162 & 0.7972 & 0.8070 & 0.8532\\
 
  Langevin-Dispersion~\cite{geissbuhler2024Langevin} & 1.62M & 30k & 0.9798 & 0.7276 & 0.7157 & 0.8887& 0.9170 & 0.8458 \\
  
  SFace2+~\cite{boutros2024sface2} & 0.63M & 10.5k & 0.9560 & 0.7711 & 0.7460 & 0.7737 & 0.8340 & 0.8162\\

  ExFaceGAN~\cite{boutros2023exfacegan} & 0.5M & 10k & 0.9350 & 0.7384 & 0.7160 & 0.7892 & 0.8298 & 0.8017  \\ 
  
  IDNet~\cite{kolf2023idnet} & 0.63M & 10.5k & 0.9288 & 0.7690 & 0.7577 & 0.7478 & 0.8192 & 0.8004 \\
  
  USynthFace~\cite{boutros2023unsupervised} & 0.4M & 400k & 0.9223 & 0.7856 & 0.7203 & 0.7162 &  0.7705 & 0.7830\\

  SFace~\cite{boutros2022sface} & 0.63M & 10.5k & 0.9187 & 0.7386 & 0.7320 & 0.7168 & 0.7793 & 0.7771\\
  

  SynFace~\cite{qiu2021synface} [\ding{70}] & 1M & 10k & 0.8580 & 0.6473 & 0.6395 & 0.5765 & 0.6987 & 0.6840 \\

  \bottomrule

\end{tabular}

[\ding{70}] Results obtained from reproduced experiments~\cite{yeung2024variface} due to data unavailability in original study.


\end{table}

To better compare the capabilities of real and synthetic data, we have assembled available test data from previous studies into~\Cref{tab:comparison_with_real}. We have decided to use LFW~\cite{huang2008lfw}, CFP-FP~\cite{sengupta2016cfp_fp} and AgeDB~\cite{moschoglou2017agedb} benchmark data from InsightFace Model Zoo~\cite{InsightFaceZoo, an2021glint360k}, which offers pretrained facial recognition models, many of them based on IResNet50, allowing for a fair comparison with synthetic datasets trained on the same backbone. It can be observed that the currently existing synthetic datasets cannot compete with the top solutions, based on large real datasets, achieving over 0.99 accuracy, however, they are already outperforming some of the real datasets - namely GlintAsian~\cite{an2021glint360k} and CASIA-WebFace~\cite{yi2014casia}. Clearly, there is still room for improvement, yet already achieved results allow for training reliable and well-performing models.

\begin{table}[h]
    \caption{Accuracy comparison of IResNet50 models trained on synthetic or real data. The best results are indicated by bold font, the second best by underlining.}
    \label{tab:comparison_with_real}
    \centering
    \begin{tabular}{cccccccc}
    \toprule
        Dataset & \#Images & \#Identities & Data & LFW & CFP-FP & AgeDB & Average \\ \midrule
        Glint360K~\cite{an2021glint360k} & 17M & 360k & Real & \underline{0.9982} &\underline{0.9914} & \textbf{0.9845} & \textbf{0.9914} \\ 
        WebFace12M~\cite{zhu2022webface260m} & 12M & 600k & Real & 0.9980 & \textbf{0.9920} & 0.9810 & \underline{0.9903} \\
        MS1MV3~\cite{deng2019lightweight} & 5.2M & 93k & Real & 0.9980 & 0.9853 & \underline{0.9827} & 0.9887 \\ 
        MS1MV2~\cite{deng2019iresnet50} & 5.8M & 87k & Real & \textbf{0.9983} & 0.9808 & 0.9808 & 0.9866 \\
        MS1M-MegaFace~\cite{deng2019iresnet50} & 14.7M & 750k & Real & 0.9975 & 0.9756 & 0.9740 & 0.9824 \\
        VGGFace2~\cite{cao2018vggface2} & 3.3M & 9k & Real & 0.9955 & 0.9741 & 0.9508 & 0.9735 \\
        VariFace~\cite{yeung2024variface} & 6M & 60k & Synthetic & 0.9960 & 0.9637 & 0.9568 & 0.9722 \\
        VIGFace~\cite{kim2024vigface} & 6M & 120k & Synthetic & 0.9948 & 0.9731 & 0.9382 & 0.9687 \\ 
        CASIA-WebFace~\cite{yi2014casia} & 0.5M & 11k & Real & 0.9945 & 0.9521 & 0.9490 & 0.9652 \\
        UIFace~\cite{lin2025uiface} & 1.5M & 30k & Synthetic & 0.9938 & 0.9596 & 0.9328 & 0.9621 \\ 
        GlintAsian~\cite{an2021glint360k} & 2.9M & 180k & Real & 0.9958 & 0.9319 & 0.9540 & 0.9606 \\
        MorphFace~\cite{mi2025morphFace} & 1.2M & 24k & Synthetic & 0.9935 & 0.9477 & 0.9327 & 0.9580 \\ \bottomrule
    \end{tabular}
\end{table}

Currently observed evaluation methodologies prioritize accuracy for dataset ranking, which is not an optimal solution in many cases. For example, in terms of biometric applications, ISO standard ISO/IEC 19795-1:2021~\cite{ISO19795} presents verification procedures distinct from those observed in the literature, such as evaluation of False Match Rate (FMR), False Non-Match Rate (FNMR), False Reject Rate (FRR) and False Accept Rate (FAR)~\cite{damer2021extended}. In this context, we observe the need to adjust the current evaluation strategy to align more closely with the intended data application instead of blind reliance on accuracy.
\section{Evaluation}
\label{sec:evaluation}
In this section, we go back to our requirements for synthetic facial recognition datasets and we analyze the current state of research to showcase progress in the field as well as to stress the limitations and to highlight areas that need further work.
We have decided to enrich the currently available data with our own experiments on available synthetic datasets. We conducted them using a pretrained IResNet100~\cite{deng2019iresnet50} model. Its training procedure involved the use of the real MS1MV2 dataset. 

\def\WidthImage{1.0\textwidth}
\def\WidthFigure{0.33\textwidth}

\begin{figure}[htbp]
    \centering
    \begin{subfigure}[b]{\WidthFigure}
        \centering               \includegraphics[width=\WidthImage]{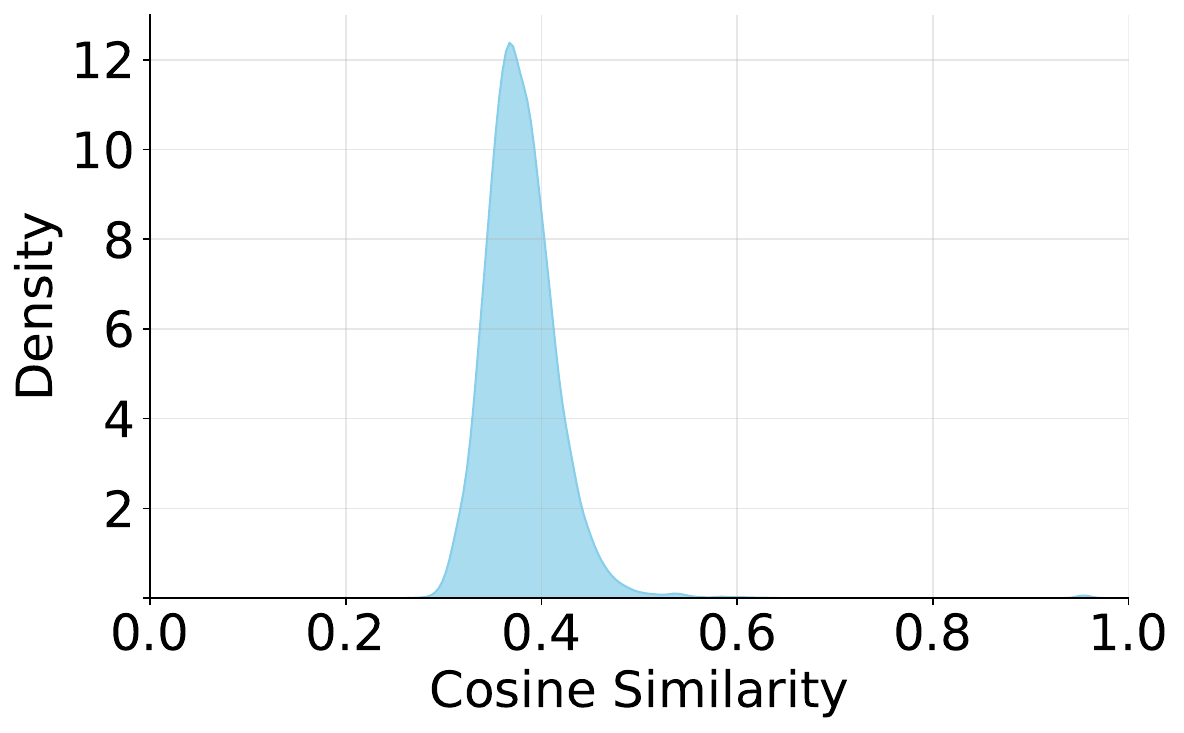} 
        \caption{CemiFace}
        \label{fig:CASIA_CemiFace}
    \end{subfigure}
    \hfill
    \begin{subfigure}[b]{\WidthFigure}
        \centering               \includegraphics[width=\WidthImage]{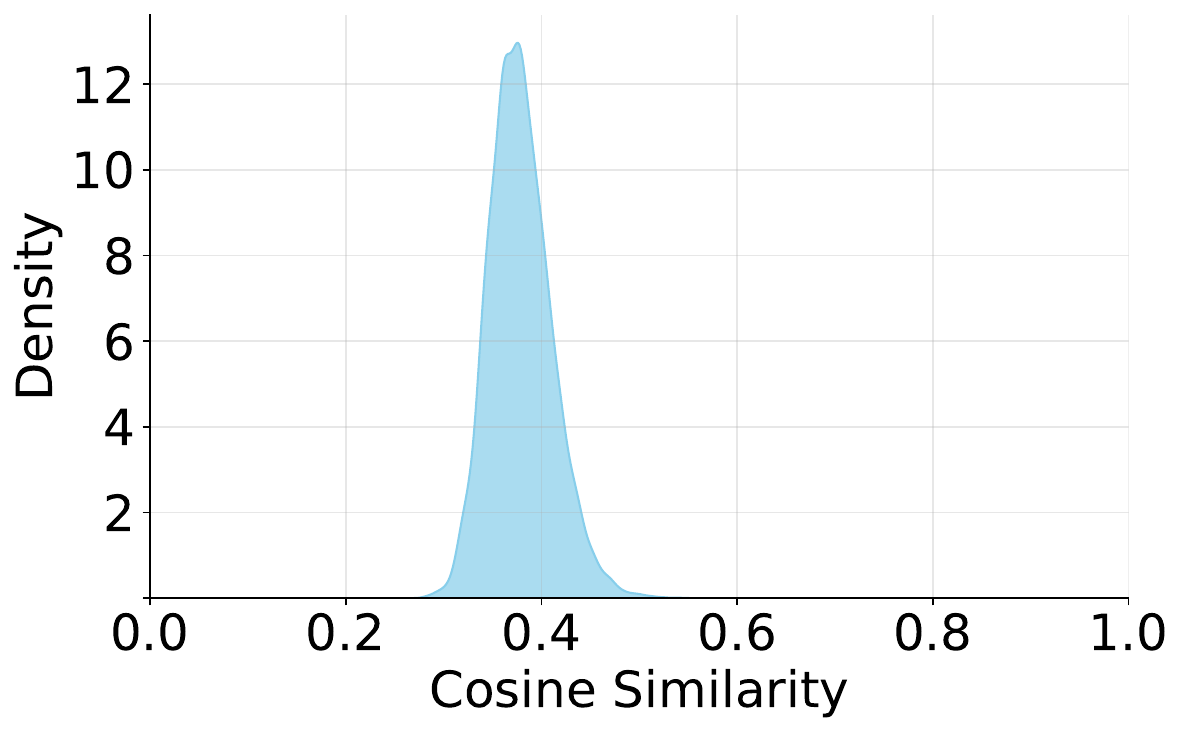} 
        \caption{DCFace}
        \label{fig:CASIA_DCFace}
    \end{subfigure}
    \hfill
        \begin{subfigure}[b]{\WidthFigure}
        \centering               \includegraphics[width=\WidthImage]{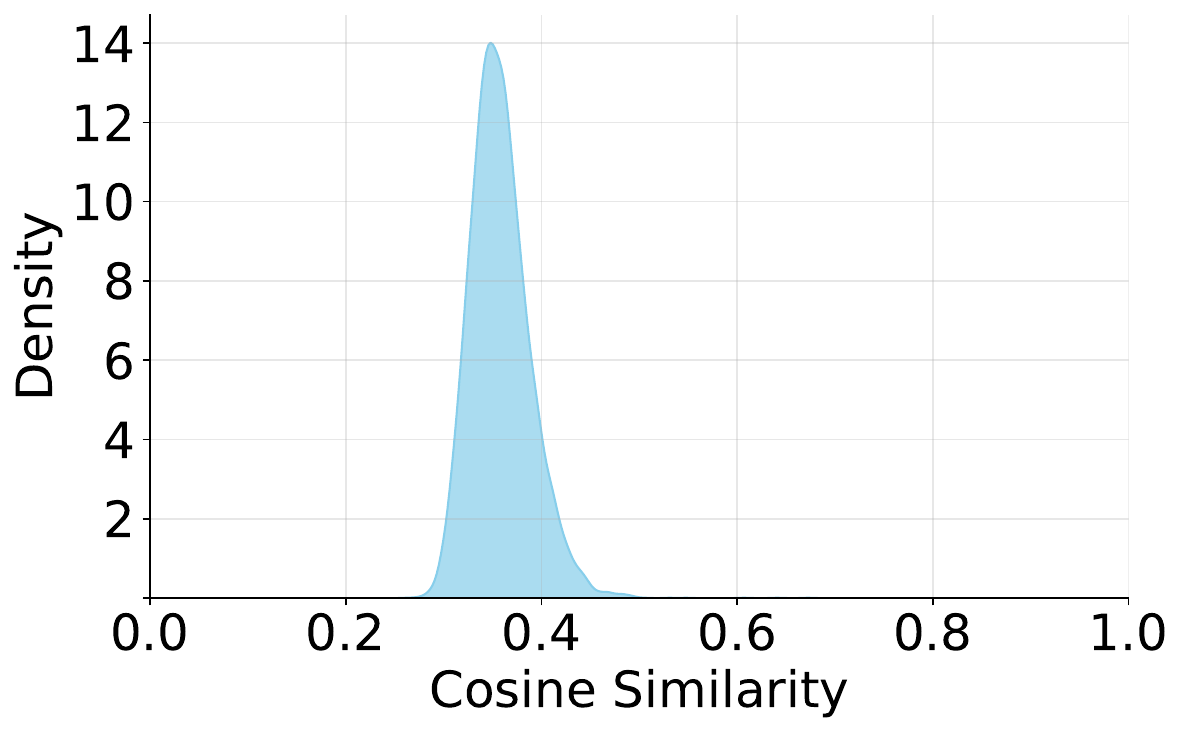} 
        \caption{DigiFace-1M}
        \label{fig:CASIA_DigiFace}
    \end{subfigure}
    \vfill
        \begin{subfigure}[b]{\WidthFigure}
        \centering               \includegraphics[width=\WidthImage]{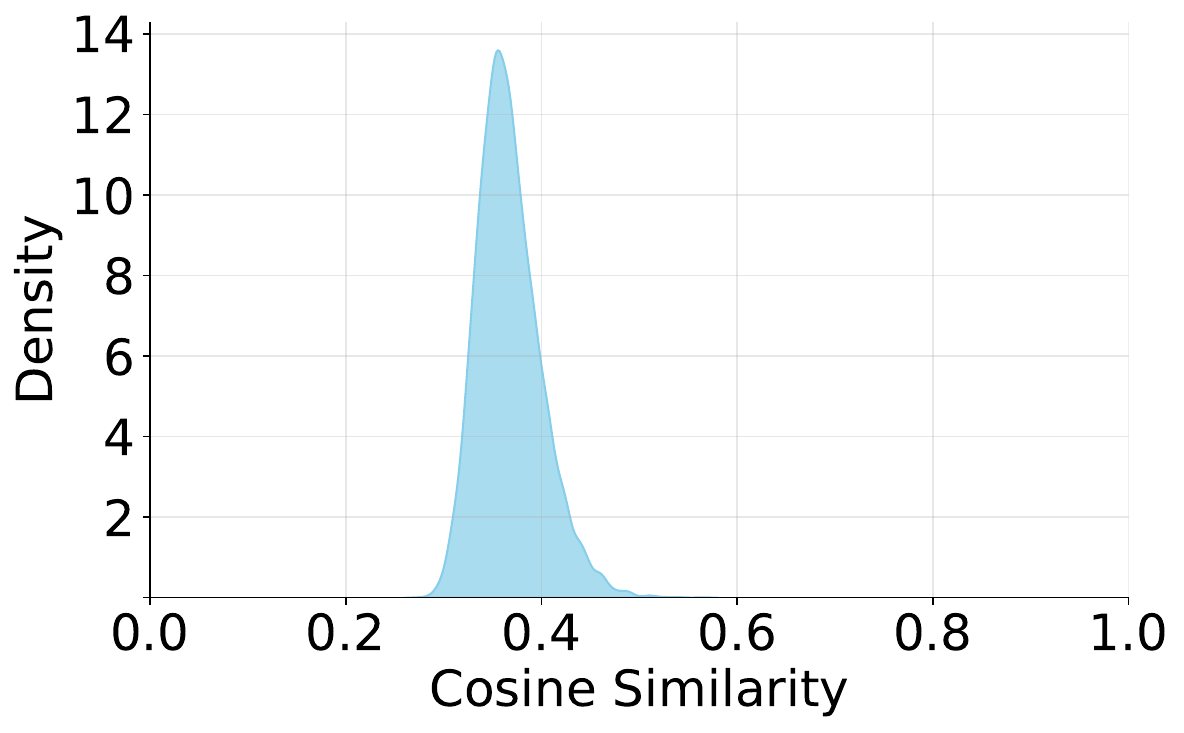} 
        \caption{Digi2Real}
        \label{fig:CASIA_Digi2Real}
    \end{subfigure}
    \hfill
        \begin{subfigure}[b]{\WidthFigure}
        \centering               \includegraphics[width=\WidthImage]{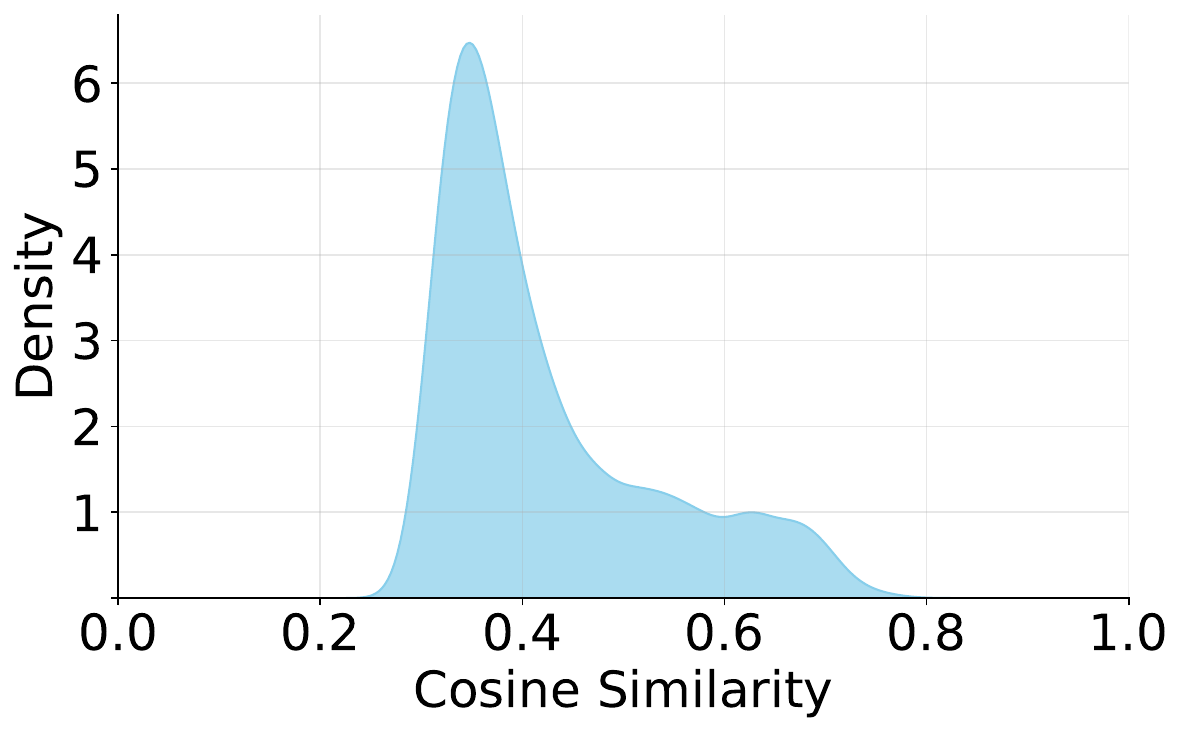} 
        \caption{GANDiffFace}
        \label{fig:CASIA_GanDiffFace}
    \end{subfigure}
    \hfill
        \begin{subfigure}[b]{\WidthFigure}
        \centering               \includegraphics[width=\WidthImage]{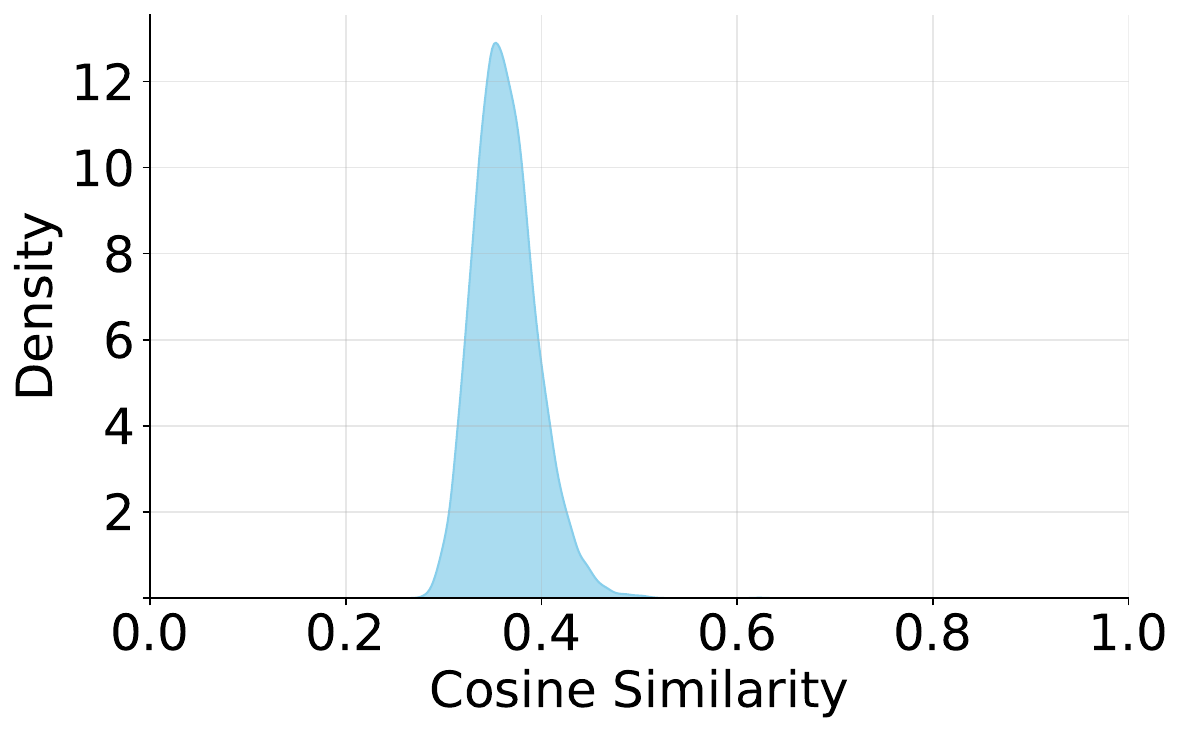} 
        \caption{HyperFace}
        \label{fig:CASIA_HyperFace}
    \end{subfigure}
    \vfill
        \begin{subfigure}[b]{\WidthFigure}
        \centering               \includegraphics[width=\WidthImage]{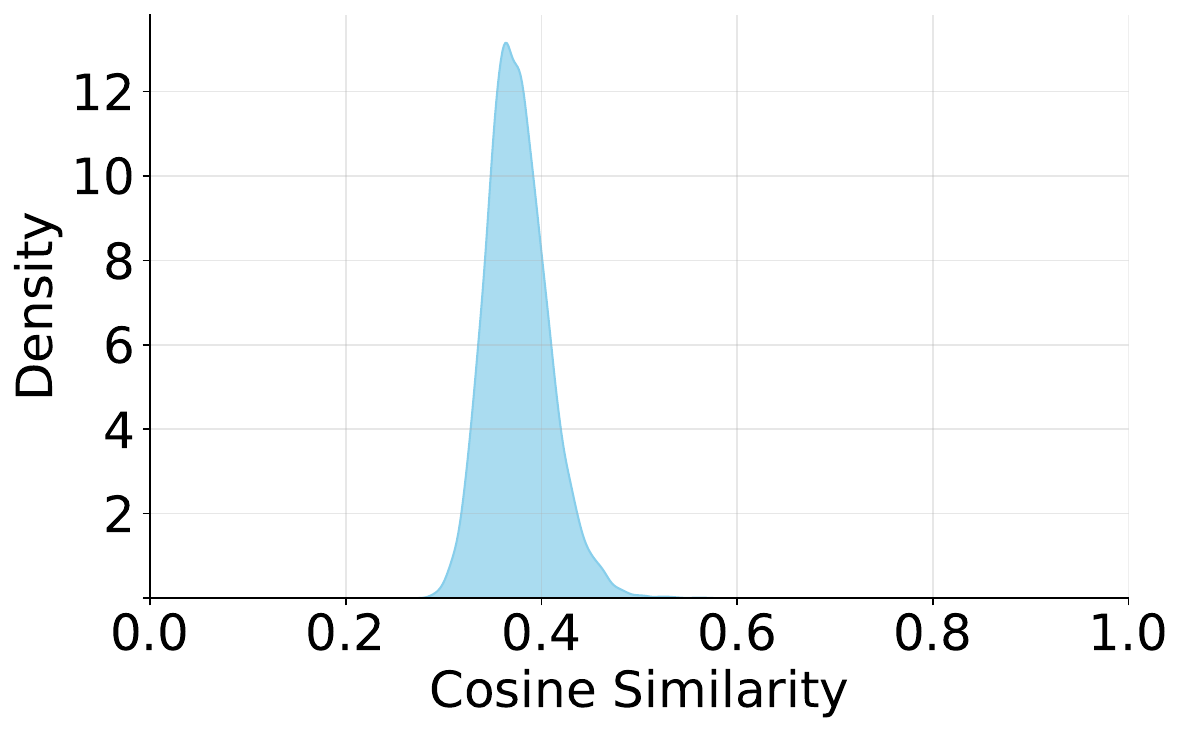} 
        \caption{Langevin-DisCo}
        \label{fig:CASIA_Langevin}
    \end{subfigure}
    \hfill
        \begin{subfigure}[b]{\WidthFigure}
        \centering               \includegraphics[width=\WidthImage]{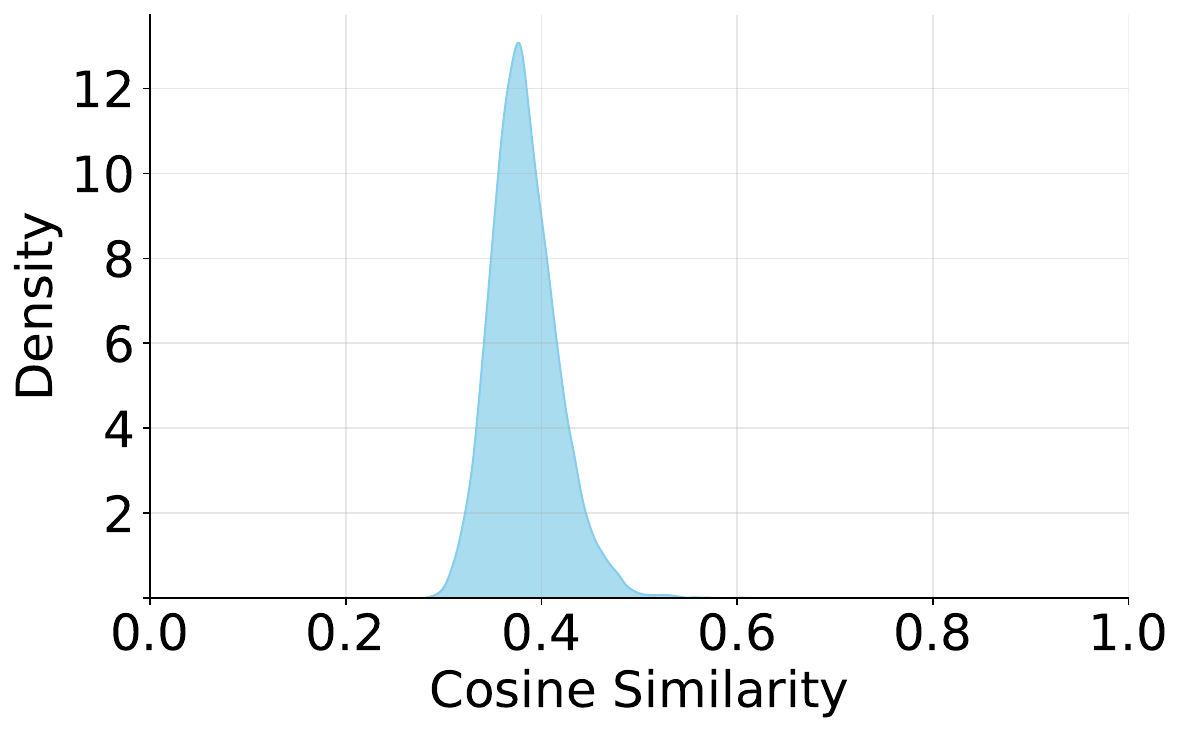} 
        \caption{MorphFace}
        \label{fig:CASIA_MorphFace}
    \end{subfigure}
    \hfill
        \begin{subfigure}[b]{\WidthFigure}
        \centering               \includegraphics[width=\WidthImage]{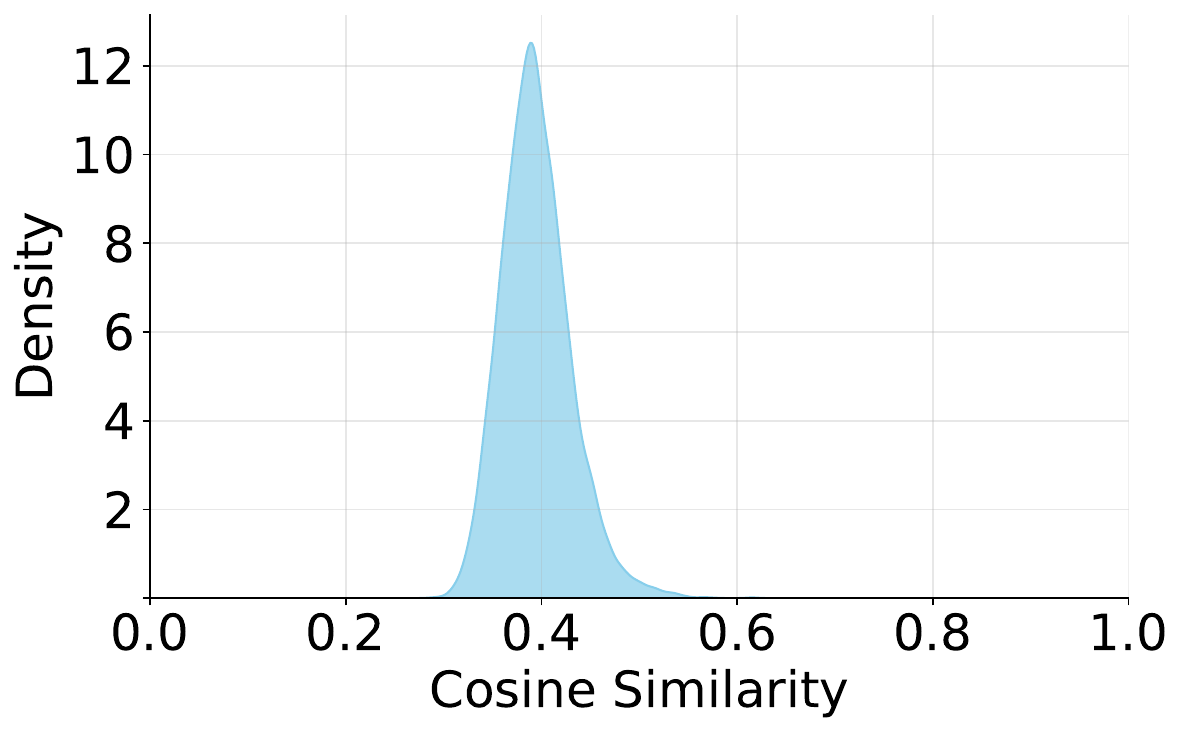} 
        \caption{SFace}
        \label{fig:CASIA_SFace}
    \end{subfigure}
    \vfill
        \begin{subfigure}[b]{\WidthFigure}
        \centering               \includegraphics[width=\WidthImage]{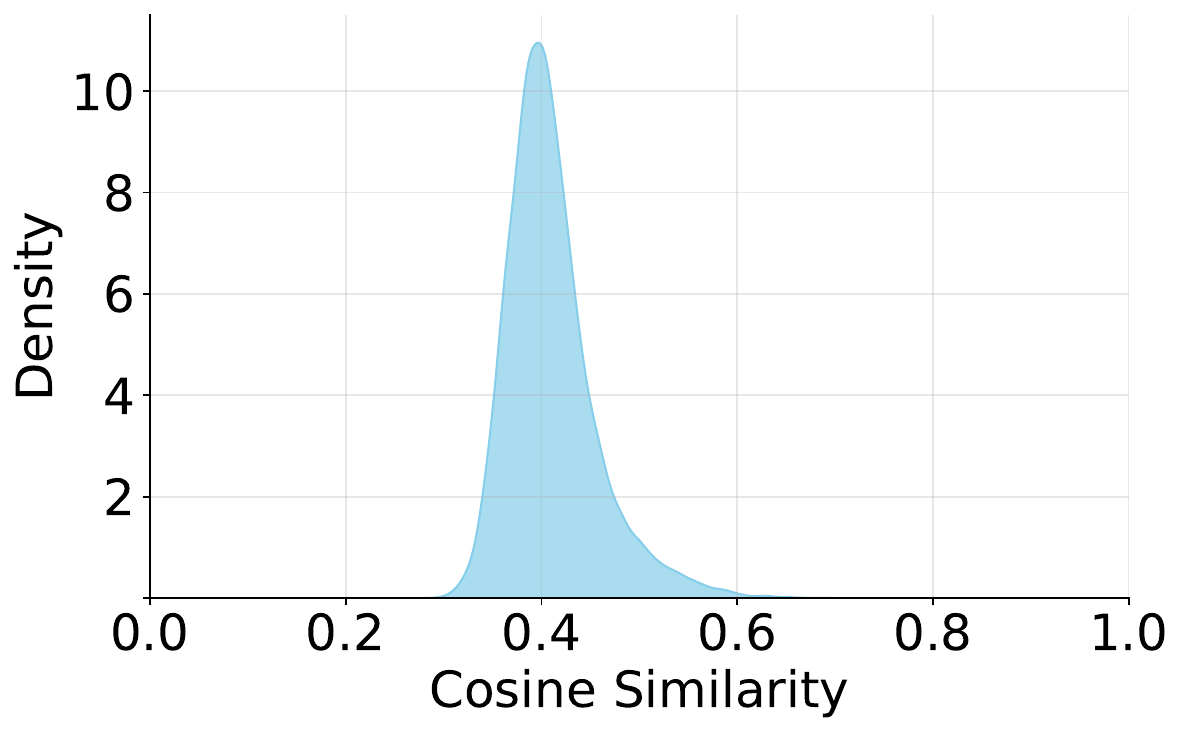} 
        \caption{SFace2}
        \label{fig:CASIA_SFace2}
    \end{subfigure}
    \hfill
        \begin{subfigure}[b]{\WidthFigure}
        \centering               \includegraphics[width=\WidthImage]{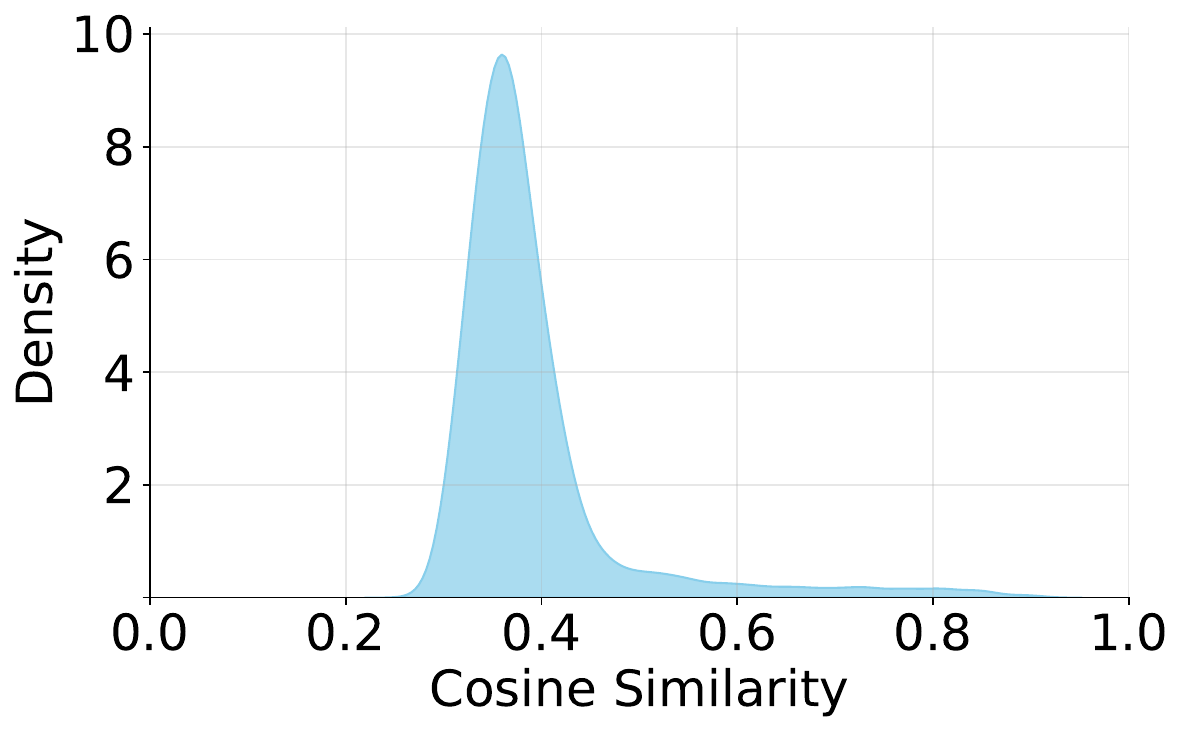} 
        \caption{Vec2Face}
        \label{fig:CASIA_Vec2Face}
    \end{subfigure}
    \hfill
        \begin{subfigure}[b]{\WidthFigure}
        \centering               \includegraphics[width=\WidthImage]{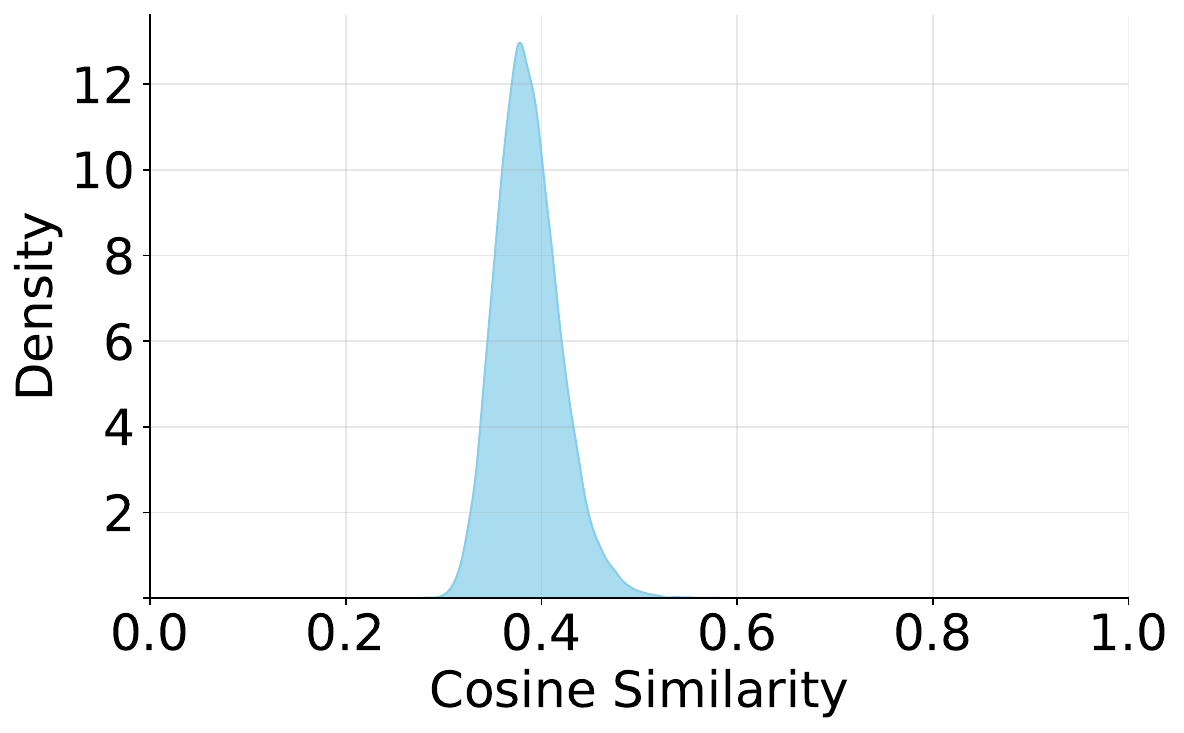} 
        \caption{SynFace}
        \label{fig:CASIA_SynFace}
    \end{subfigure}
    \vfill
        \begin{subfigure}[b]{\WidthFigure}
        \centering               \includegraphics[width=\WidthImage]{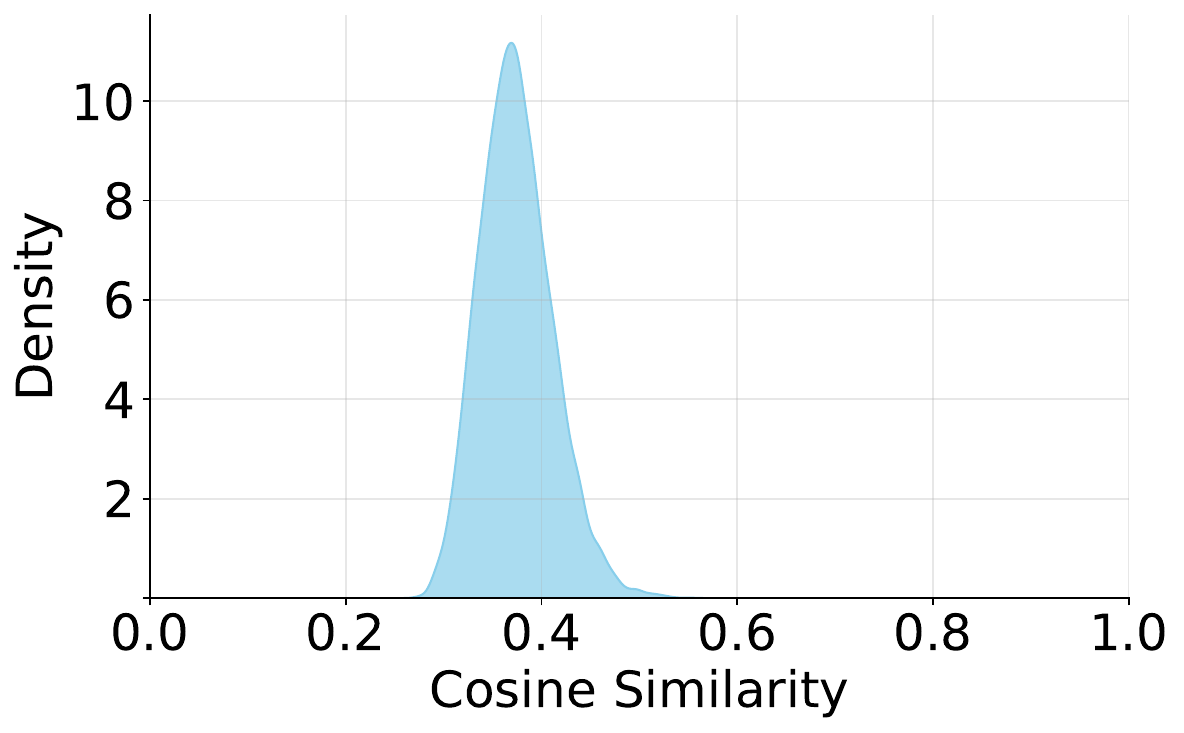} 
        \caption{IDiffFace}
        \label{fig:CASIA_IDiffFace}
    \end{subfigure}
    \hfill
        \begin{subfigure}[b]{\WidthFigure}              
        \includegraphics[width=\WidthImage]{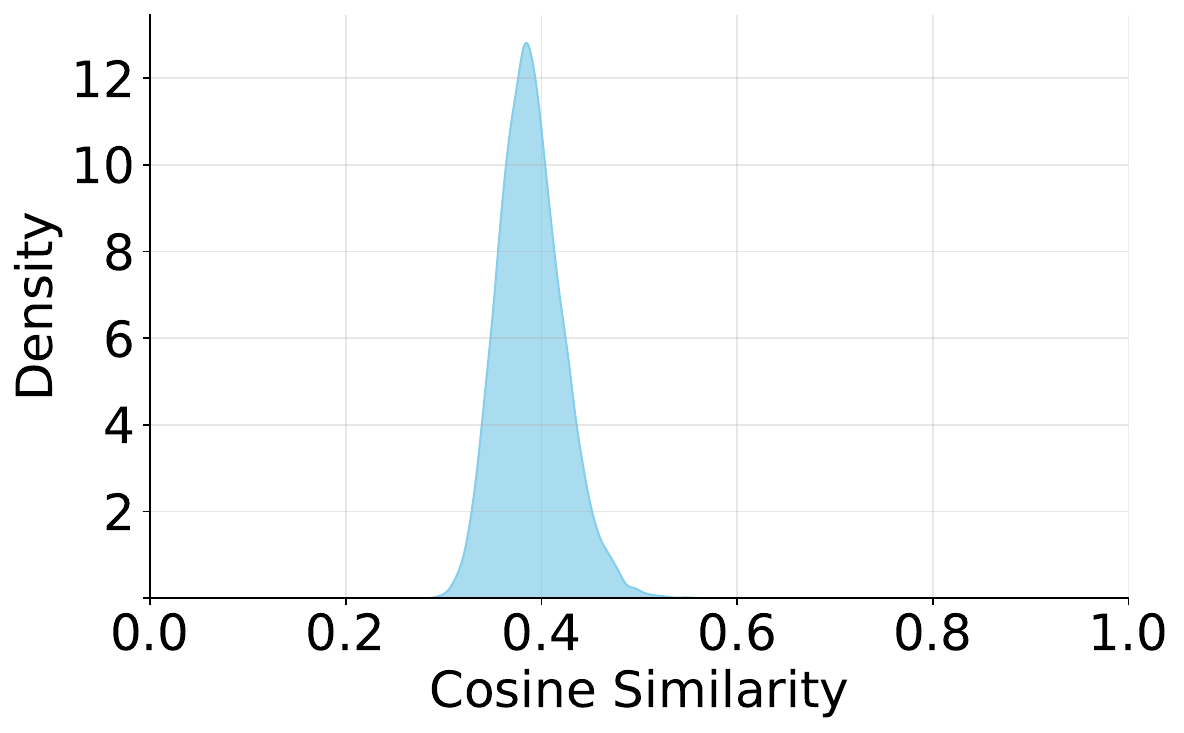} 
        \caption{IDNet}
        \label{fig:CASIA_IDNet}
    \end{subfigure}
    \hfill
        \begin{subfigure}[b]{\WidthFigure}              
        \includegraphics[width=\WidthImage]{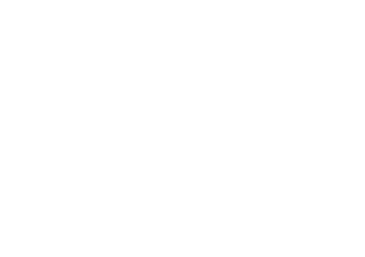} 
    \end{subfigure}

    \caption{Cosine similarities distribution of closest samples with respect to CASIA-WebFace dataset.}
    \label{fig:ClosestImage}
\end{figure}

\subsection{R1 - Identity Leakage
Prevention}
Creating a synthetic dataset, one wants to divert away from the original identities. Consequently, the aim is to observe low similarity scores when compared with the samples from the training set. An appropriate threshold should be defined based on the particular model's behavior, optimally always determining that every image from the real dataset belongs to a different identity than any synthesized sample. 
It is not an uncommon practice to address identity issues by removing samples that have too high similarity scores with other classes~\cite{papantoniou2024arc2face, wu2024vec2face}.
Vec2Face~\cite{wu2024vec2face} has reported only 0.4\% overlap of generated samples, which ultimately were discarded from the dataset. Yet, another study~\cite{kim2024vigface} has reported identity leakage and claims that the method is, in fact, failing in creating novel identities. Similarly, high image likeness between original and synthetic datasets has also been attributed to CemiFace~\cite{sun2024cemiface}.

To assess the degree of data leakage, we have checked each available dataset to find the most similar samples with respect to the most commonly used CASIA-WebFace~\cite{yi2014casia} dataset (\Cref{fig:ClosestImage}). While not all synthetic datasets used the aforementioned dataset during the training process, it is highly likely that there is a certain overlap between the used identities, especially if multiple real datasets were used in the process. We have observed that all of the tested datasets tend to have very similar centers of distributions, with the most frequent closest samples having scores below 0.4, implying proper separation from real samples. Even though this value itself is not alarming, the existence of long-tail distributions among some of the tested datasets certainly is. The problem is most clearly visible in the examples of GANDiffFace~\cite{melzi2023gandiffface} (\Cref{fig:CASIA_GanDiffFace}), Vec2Face~\cite{wu2024vec2face} (\Cref{fig:CASIA_Vec2Face}), and SFace2~\cite{boutros2024sface2} (\Cref{fig:CASIA_SFace2}).
While SFace2 and Vec2Face used CASIA-WebFace in their training process (in the case of Vec2Face, a model pretrained on Glint360k~\cite{an2021glint360k}, which contains images from CASIA-WebFace), GandDiffFace was created using only FFHQ~\cite{karras2019ffhq}. It suggests an overlap between the real datasets and is consistent with the data preprocessing approach of the CBSR team in the FRCSyn-onGoing~\cite{melzi2024frcsynOngoing1} challenge.  
Significantly smaller amounts of high-similarity outliers can also be observed in CemiFace~\cite{sun2024cemiface} (\Cref{fig:CASIA_CemiFace}) and DigiFace-1m~\cite{bae2023digiface} (\Cref{fig:CASIA_DigiFace}), suggesting that there may exist a small number of easily filterable samples where identity leakage is particularly strong. 

\def\WidthImage{1.0\textwidth}
\def\WidthFigure{0.3\textwidth}

\begin{figure}[htbp]
    \centering
    \begin{subfigure}[b]{\WidthFigure}
        \centering               \includegraphics[width=\WidthImage]{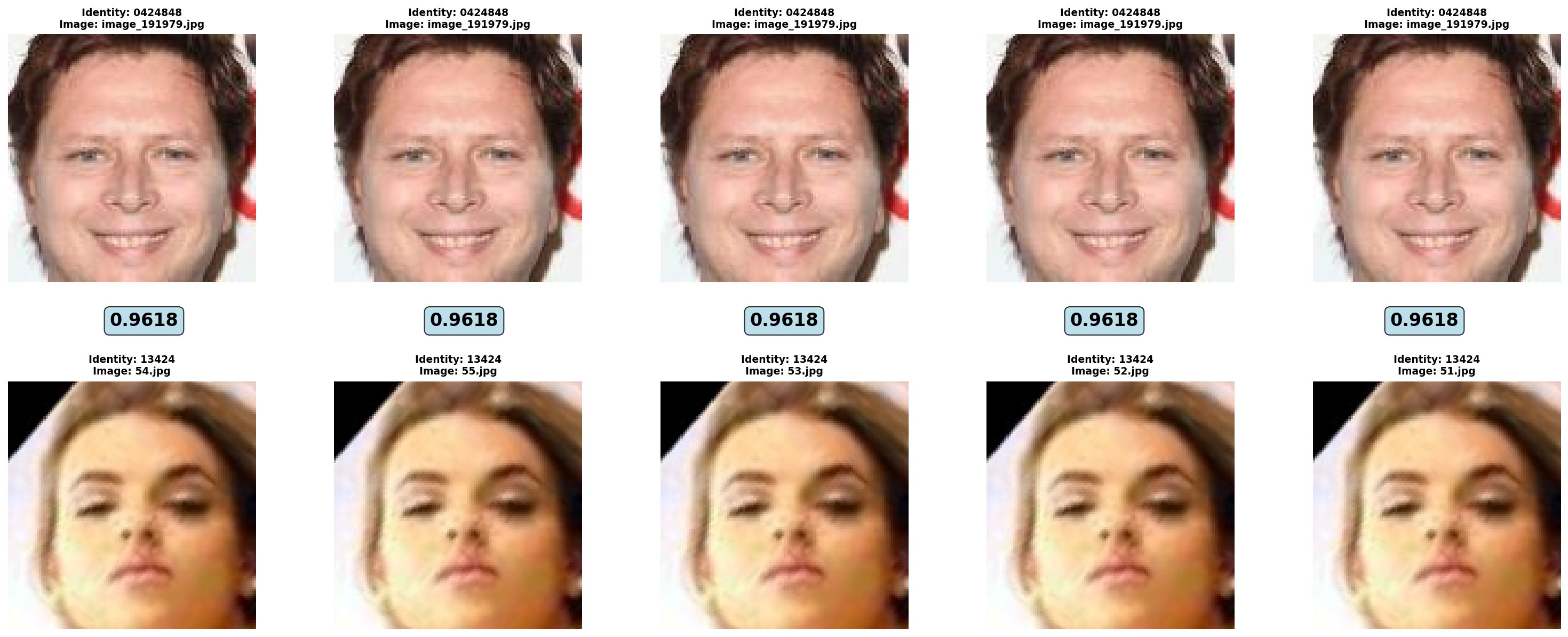} 
        \caption{CemiFace}
        \label{fig:Closest_CemiFace}
    \end{subfigure}
    \hfill
    \begin{subfigure}[b]{\WidthFigure}
        \centering               \includegraphics[width=\WidthImage]{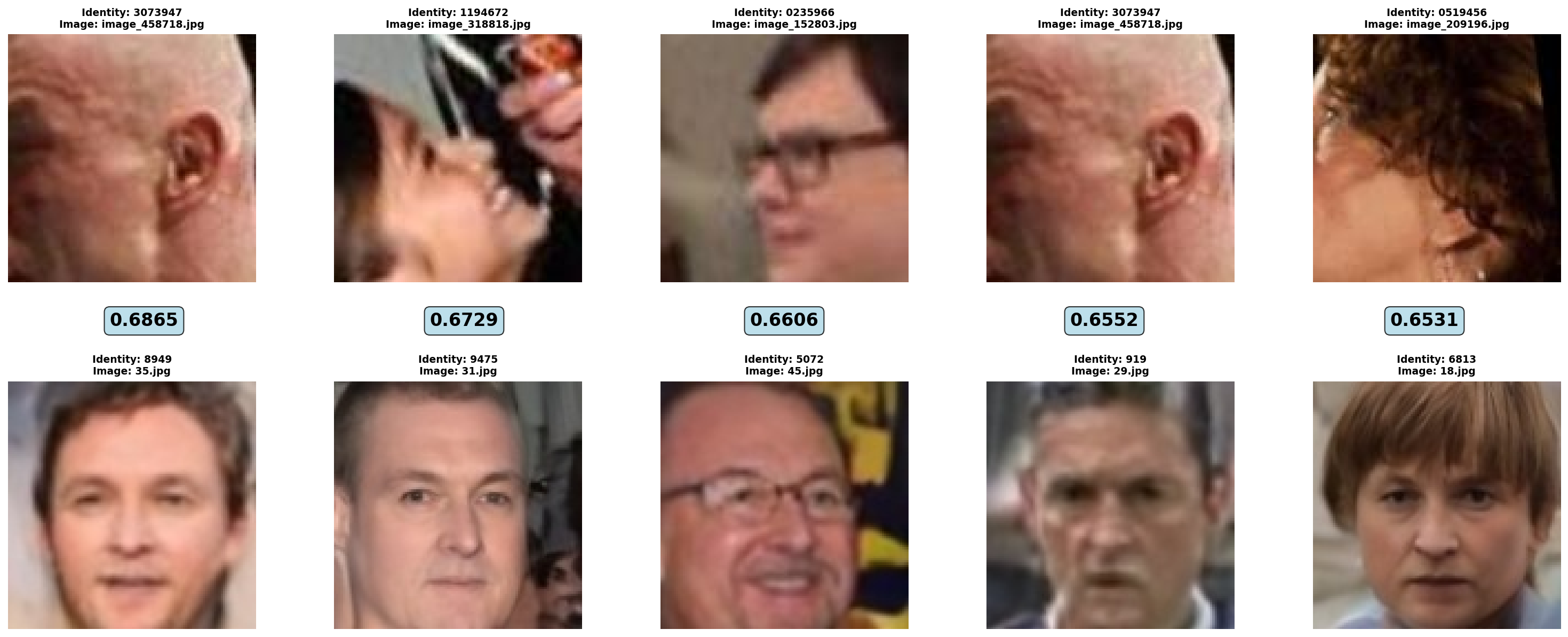} 
        \caption{DCFace}
        \label{fig:Closest_DCFace}
    \end{subfigure}
    \hfill
        \begin{subfigure}[b]{\WidthFigure}
        \centering               \includegraphics[width=\WidthImage]{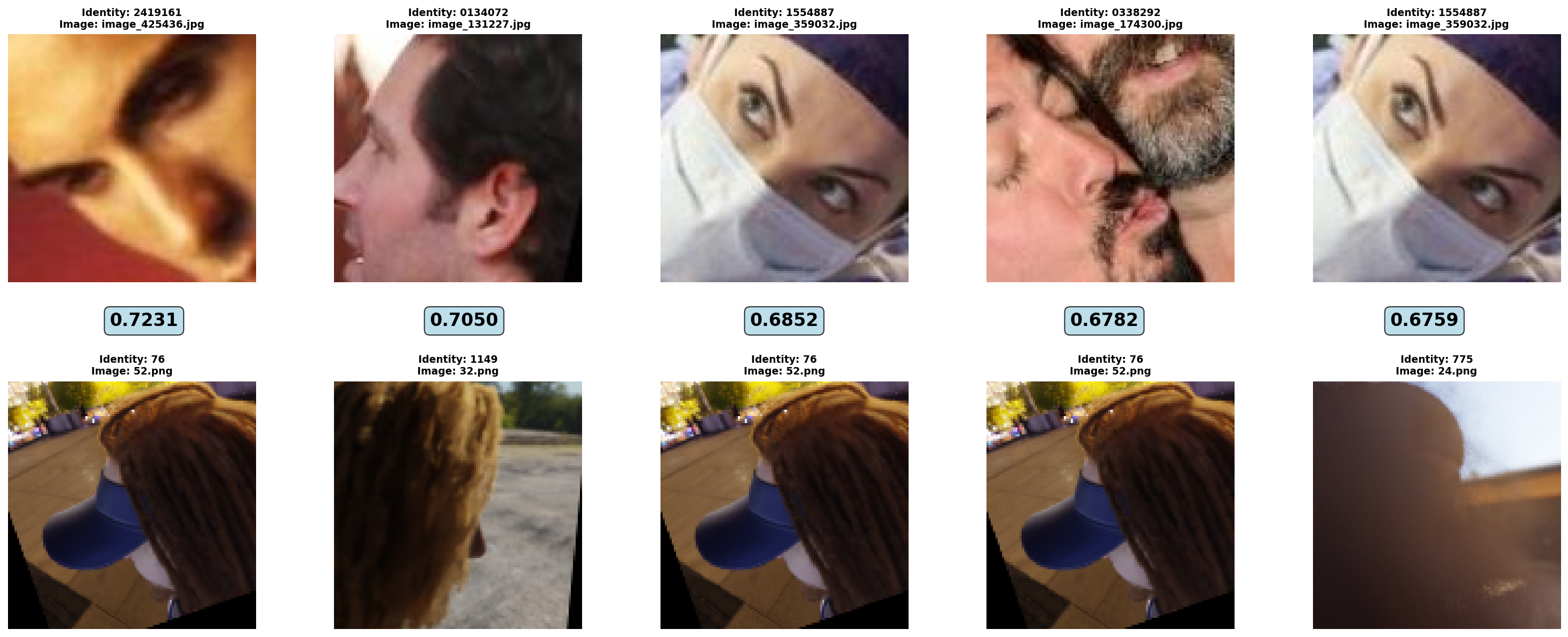} 
        \caption{DigiFace-1M}
        \label{fig:Closest_DigiFace}
    \end{subfigure}
    \vfill
        \begin{subfigure}[b]{\WidthFigure}
        \centering               \includegraphics[width=\WidthImage]{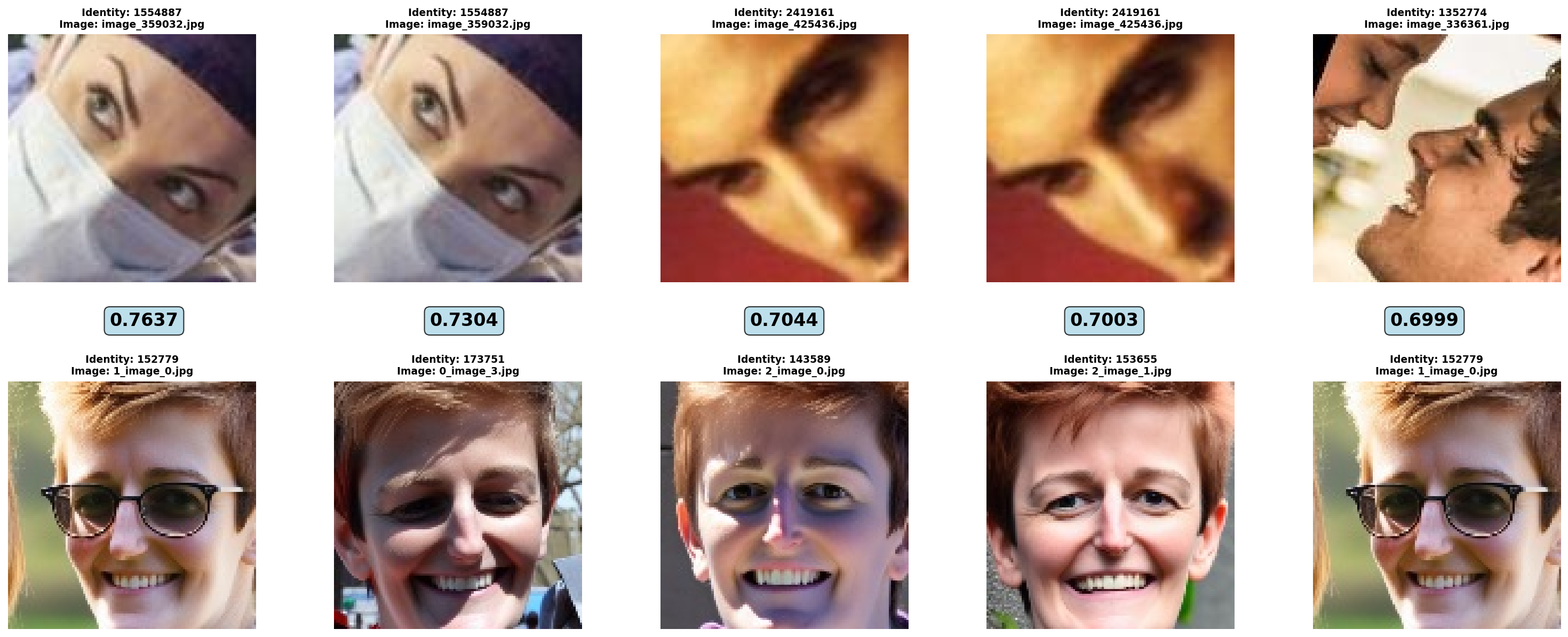} 
        \caption{Digi2Real}
        \label{fig:Closest_Digi2Real}
    \end{subfigure}
    \hfill
        \begin{subfigure}[b]{\WidthFigure}
        \centering               \includegraphics[width=\WidthImage]{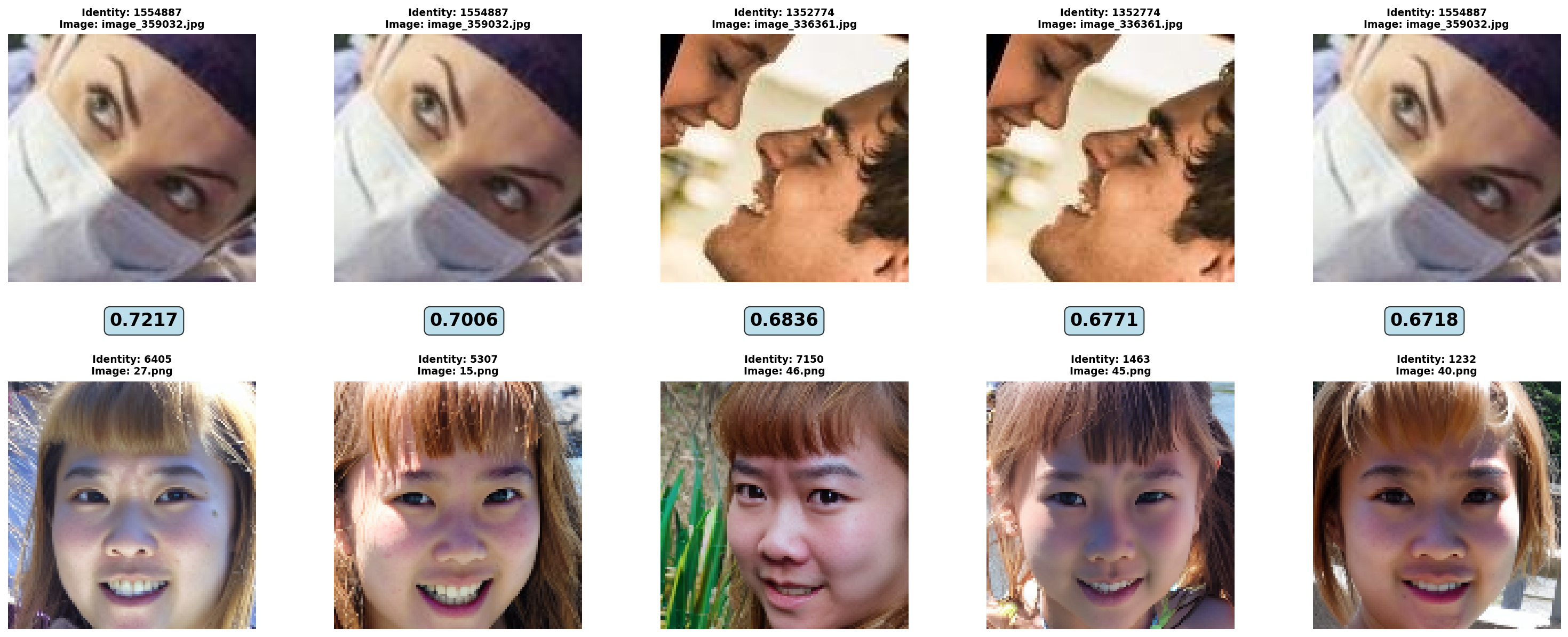} 
        \caption{HyperFace}
        \label{fig:Closest_HyperFace}
    \end{subfigure}
    \hfill
        \begin{subfigure}[b]{\WidthFigure}
        \centering               \includegraphics[width=\WidthImage]{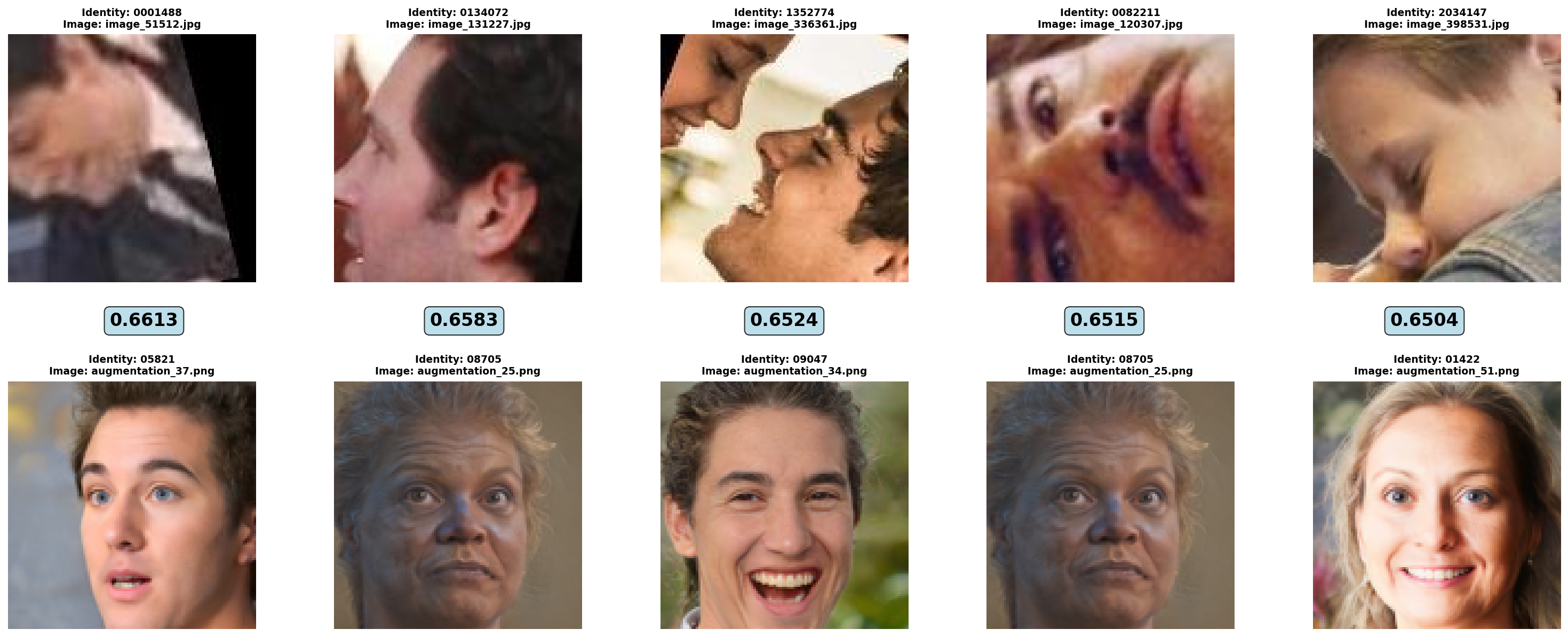} 
        \caption{Langevin-DisCo}
        \label{fig:Closest_Langevin}
    \end{subfigure}
    \vfill
        \begin{subfigure}[b]{\WidthFigure}
        \centering               \includegraphics[width=\WidthImage]{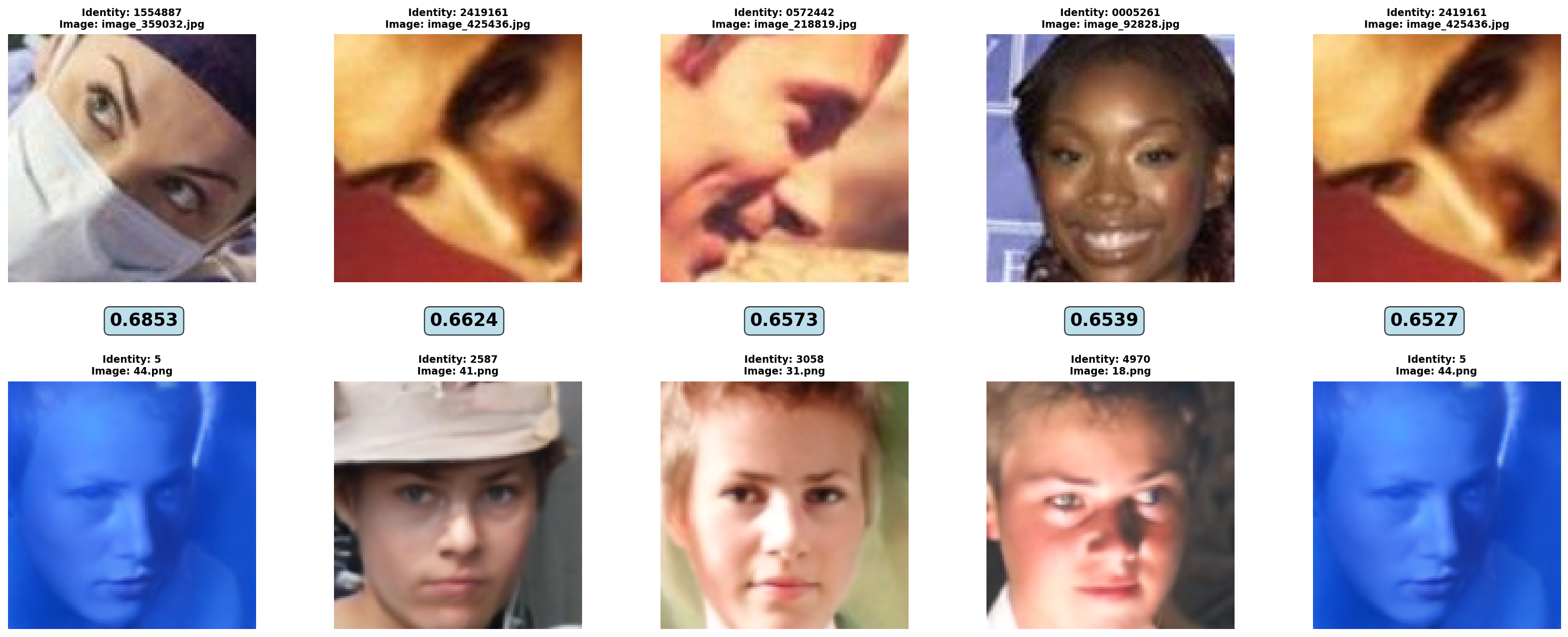} 
        \caption{MorphFace}
        \label{fig:Closest_MorphFace}
    \end{subfigure}
    \hfill
        \begin{subfigure}[b]{\WidthFigure}
        \centering               \includegraphics[width=\WidthImage]{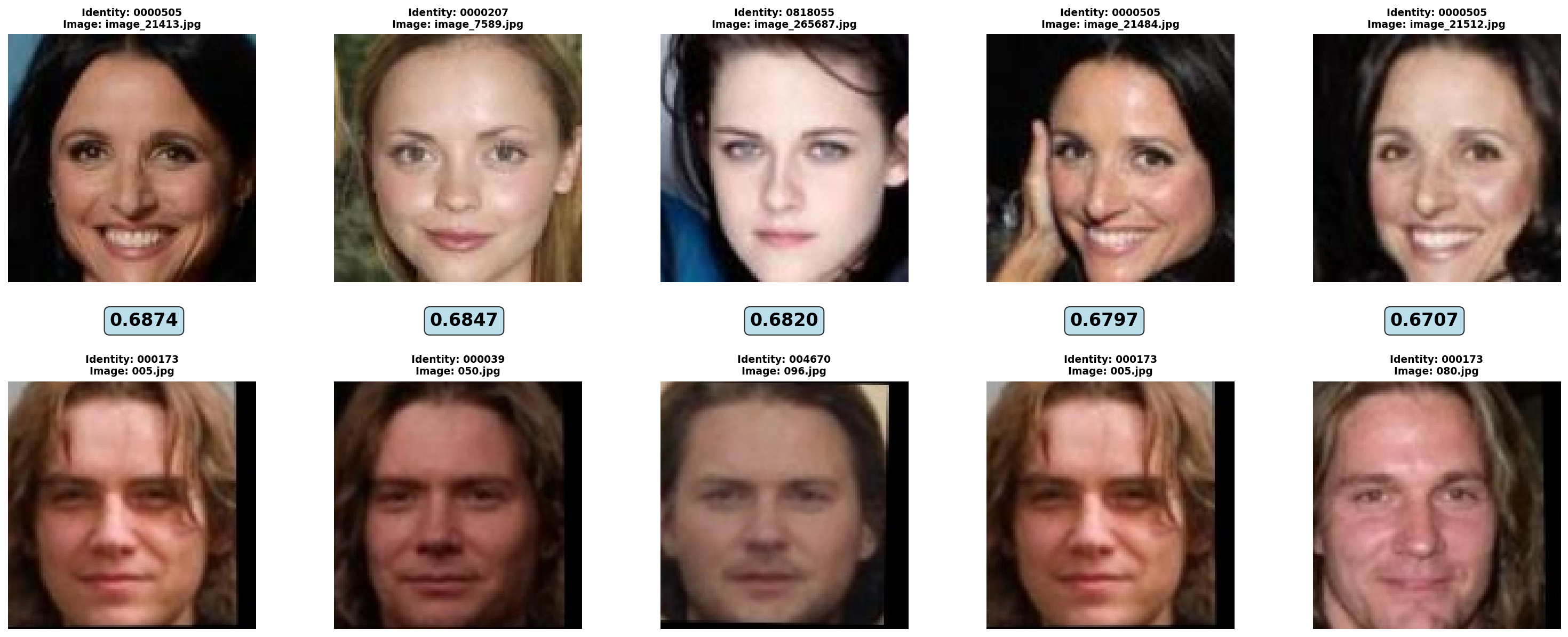} 
        \caption{SFace2}
        \label{fig:Closest_SFace2}
    \end{subfigure}
    \hfill
        \begin{subfigure}[b]{\WidthFigure}
        \centering               \includegraphics[width=\WidthImage]{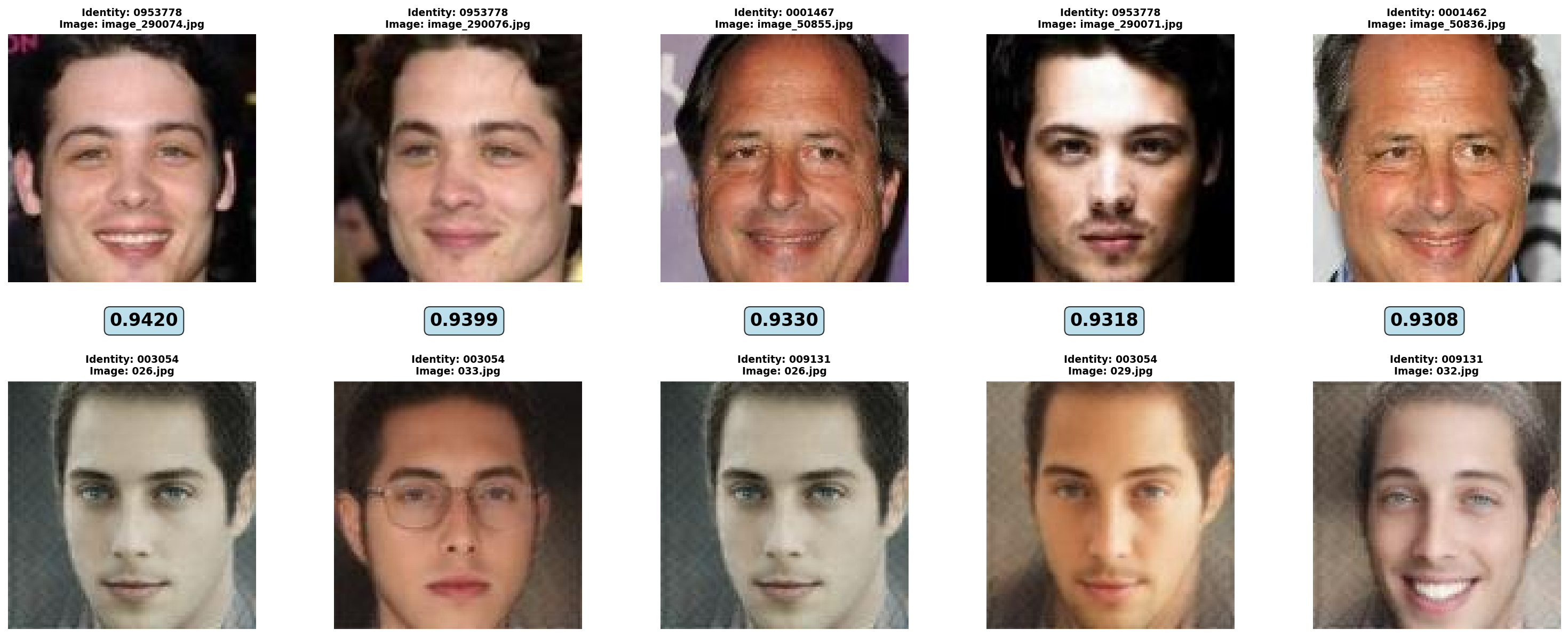} 
        \caption{Vec2Face}
        \label{fig:Closest_Vec2Face}
    \end{subfigure}
    \vfill
        \begin{subfigure}[b]{\WidthFigure}
        \centering               \includegraphics[width=\WidthImage]{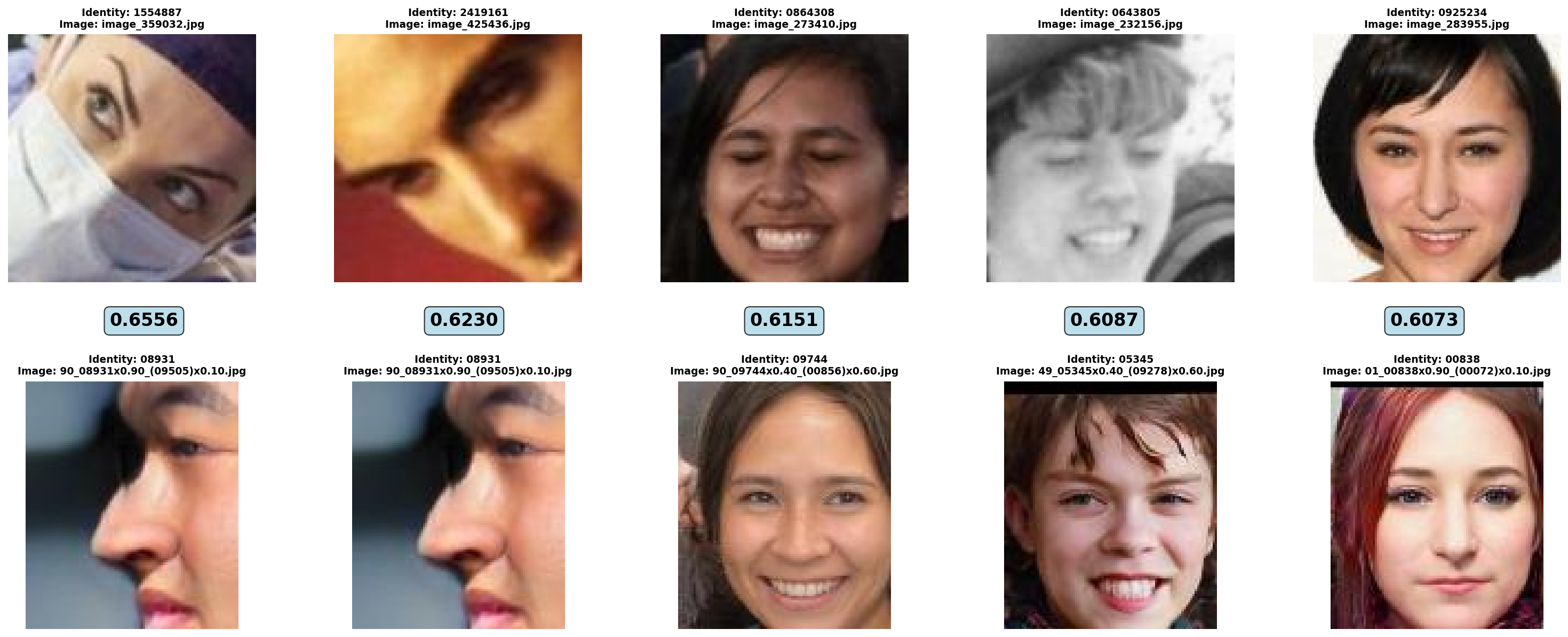} 
        \caption{SynFace}
        \label{fig:Closest_SynFace}
    \end{subfigure}
    \hfill
        \begin{subfigure}[b]{\WidthFigure}
        \centering               \includegraphics[width=\WidthImage]{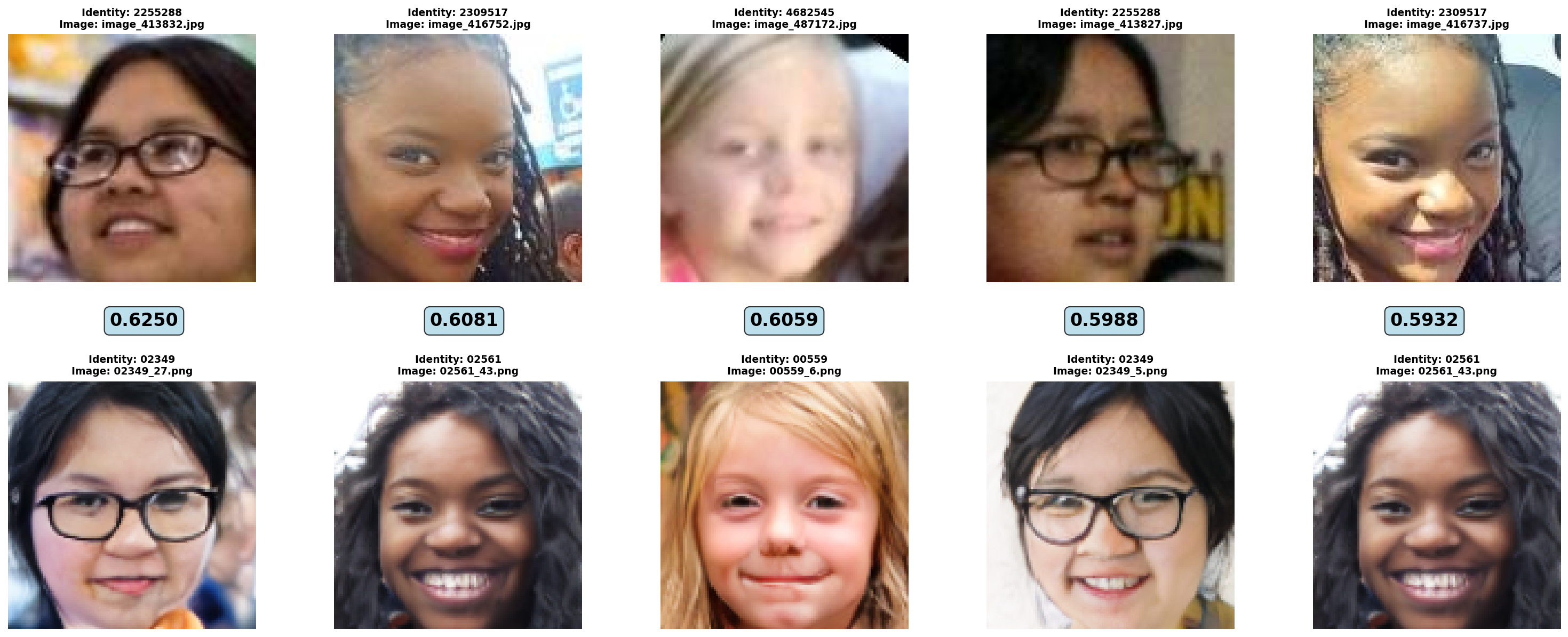} 
        \caption{IDiffFace}
        \label{fig:Closest_IDiffFace}
    \end{subfigure}
    \hfill
        \begin{subfigure}[b]{\WidthFigure}
        \centering               \includegraphics[width=\WidthImage]{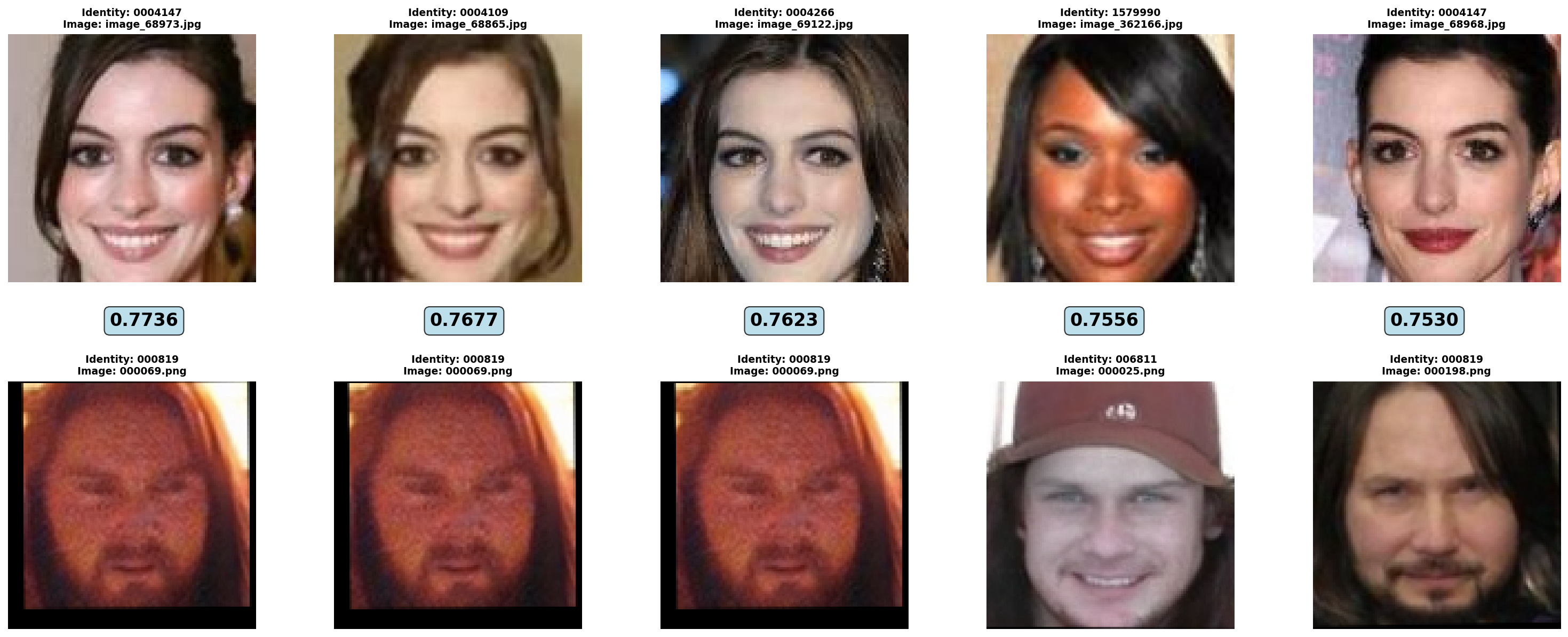} 
        \caption{SFace}
        \label{fig:Closest_SFace}
    \end{subfigure}
    \vfill
        \begin{subfigure}[b]{\WidthFigure}
        \centering               \includegraphics[width=\WidthImage]{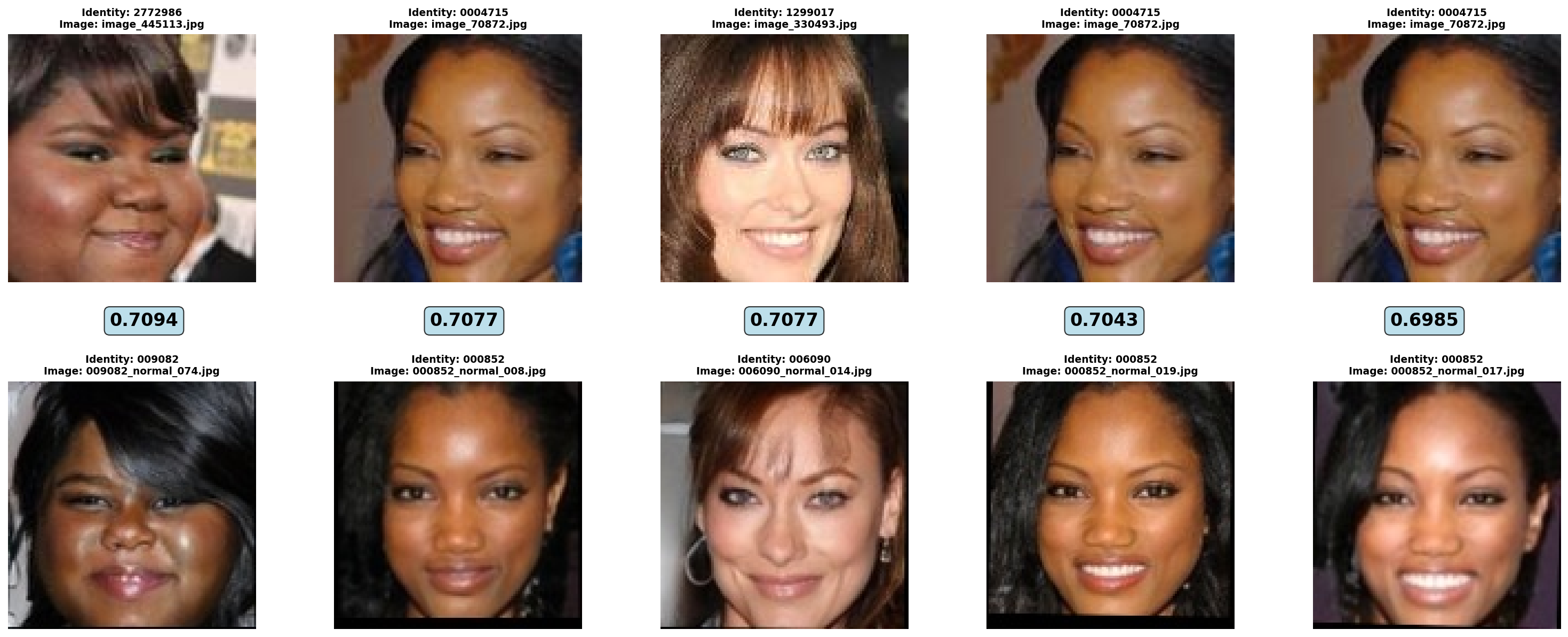} 
        \caption{IDNet}
        \label{fig:Closest_IDNet}
    \end{subfigure}
    \hfill
        \begin{subfigure}[b]{\WidthFigure}
        \centering               \includegraphics[width=\WidthImage]{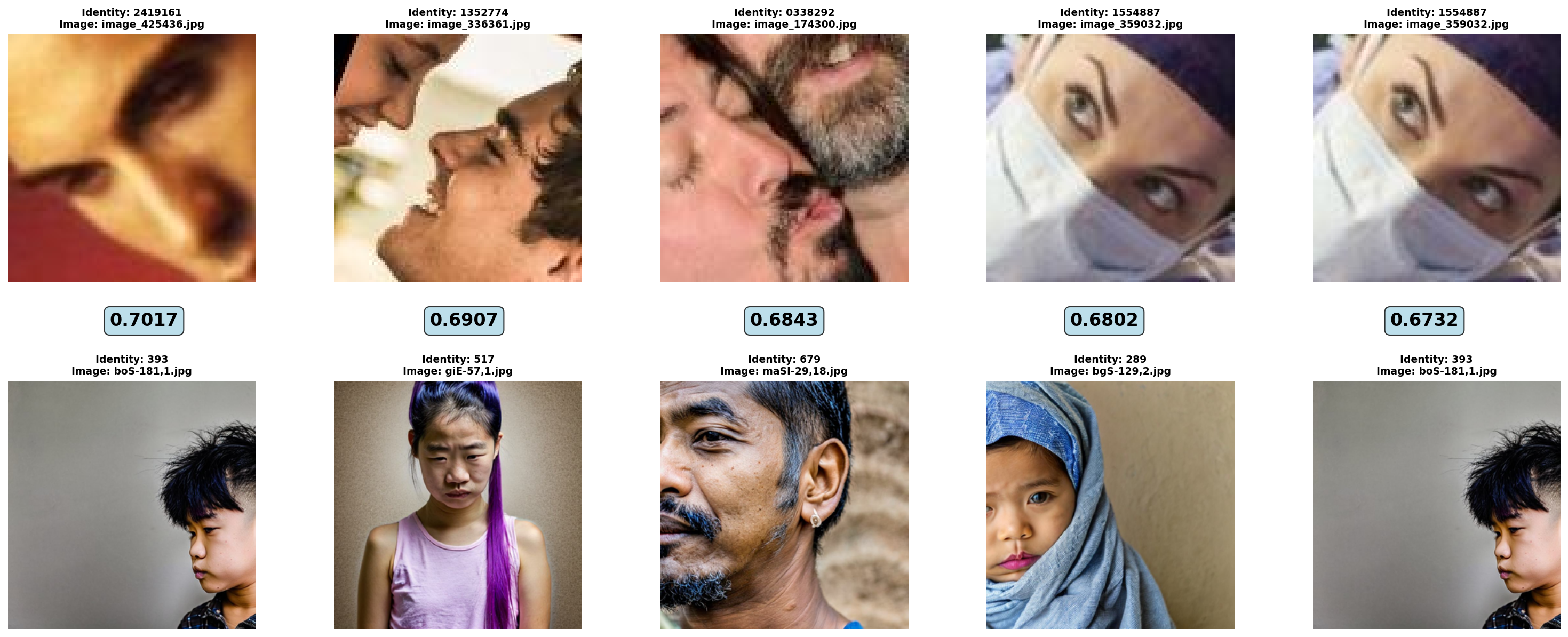} 
        \caption{SDFD}
        \label{fig:Closest_SDFD}
    \end{subfigure}
    \hfill
        \begin{subfigure}[b]{\WidthFigure}
        \centering               \includegraphics[width=\WidthImage]{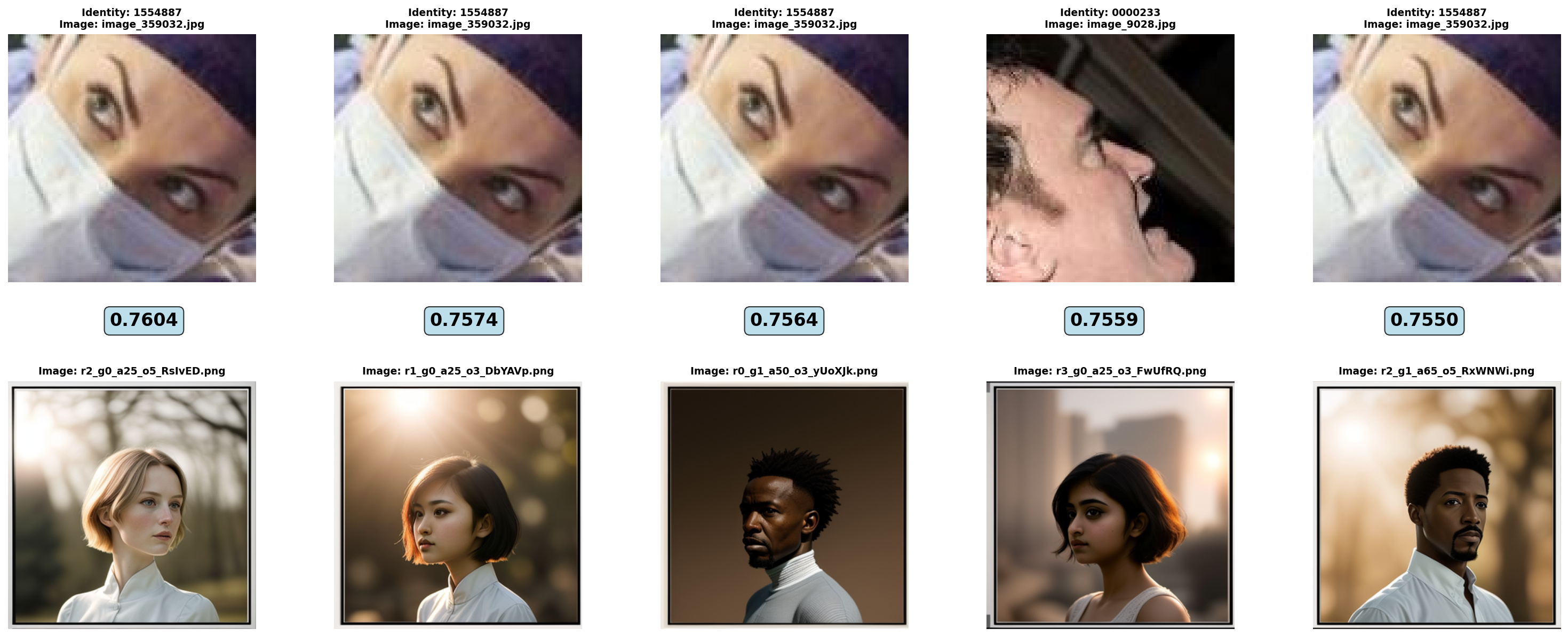} 
        \caption{ControlFace10k}
        \label{fig:Closest_ControlFace10k}
    \end{subfigure}
    \vfill
        \begin{subfigure}[b]{\WidthFigure}
        \centering               \includegraphics[width=\WidthImage]{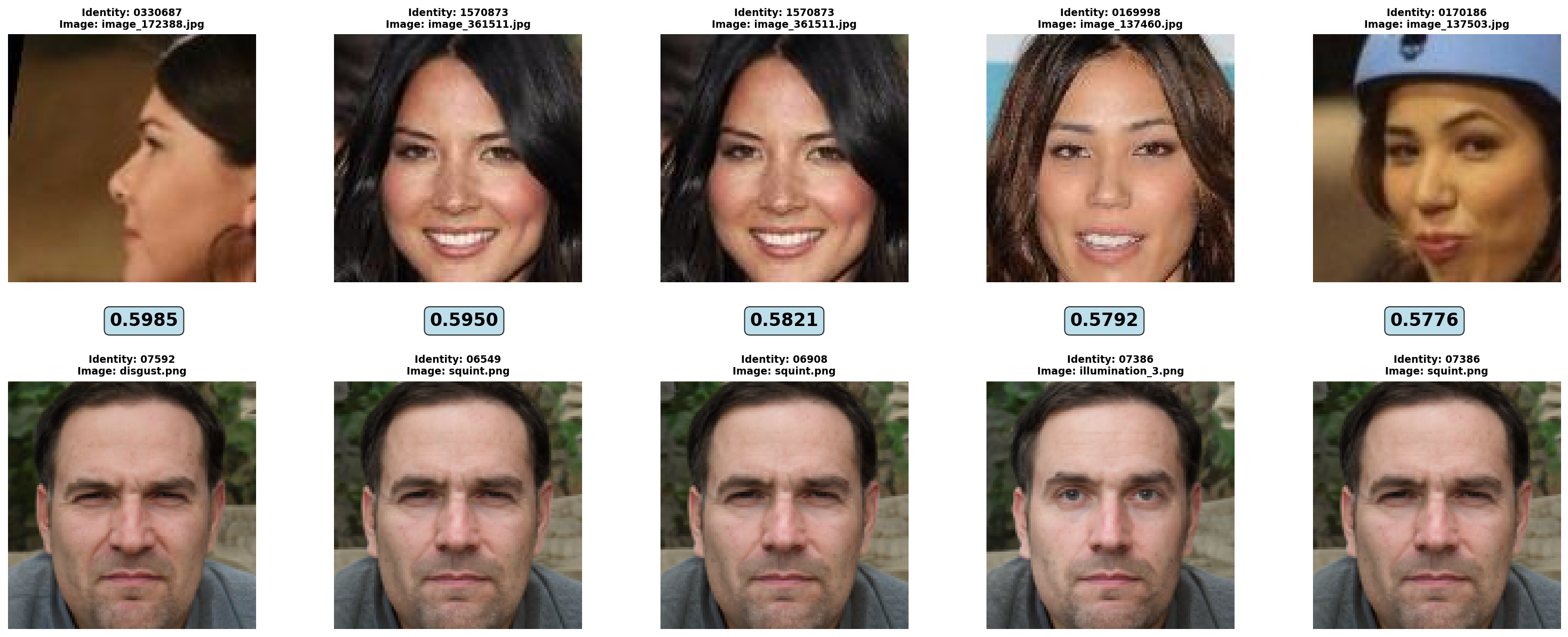} 
        \caption{Syn-Multi-Pie}
        \label{fig:Closest_Syn-Multi-Pie}
    \end{subfigure}
    \caption{Top 5 highest cosine similarity scores of synthetic samples (upper row) with respect to CASIA-WebFace (lower row) dataset}
    \label{fig:LeakageCASIA}
\end{figure}

The very leakage for the most similar samples may be evaluated in an empirical way. However, it has already been reported~\cite{shahreza2024hyperface} as a challenging task. Nonetheless, based on data presented in Figure~\ref{fig:LeakageCASIA}, there are some visible patterns. First of all, one can notice label noise in the case of CemiFace~\cite{sun2024cemiface}, where all of the most similar images are the same (\Cref{fig:Closest_CemiFace}), yet they are separate files in the scope of the same identity. On the other hand, in Syn-Multi-PIE~\cite{colbois2021synmultipie} label noise affects identities (\Cref{fig:Closest_Syn-Multi-Pie}) as all the presented synthetic samples resemble a very high degree of similarity, but are considered as different identities. 
In Vec2Face~\cite{wu2024vec2face}, one can observe that two pairs of identities are consistently paired with very high similarity scores, even though visually identities seem to be distinct (\Cref{fig:Closest_Vec2Face}). However, one has to take into account potential temporal changes as one of the compared identities differs significantly in age. Interestingly, identity's \textit{1554887} image with a facemask on, from CASIA-WebFace, appears across multiple datasets within the most similar samples (Figures~\ref{fig:Closest_DigiFace},~\ref{fig:Closest_Digi2Real},~\ref{fig:Closest_HyperFace},~\ref{fig:Closest_MorphFace},~\ref{fig:Closest_SynFace},~\ref{fig:Closest_SDFD},~\ref{fig:Closest_ControlFace10k}), indicating a potential model's bias. Two examples where the identity leakage seems to be the most apparent are IDNet~\cite{kolf2023idnet} (\Cref{fig:Closest_IDNet}) and IDiffFace~\cite{boutros2023idiff} (\Cref{fig:Closest_IDiffFace}). In case of some of the datasets, such as DigiFace-1M~\cite{bae2023digiface} (\Cref{fig:Closest_DigiFace}) or MorphFace~\cite{mi2025morphFace} (\Cref{fig:Closest_MorphFace}), the highest similarities are products of comparisons made on misformed samples.

While some of the works in the field focus on the identity leakage problem, there is very little attention paid to the idea of reverse engineering the underlying images from generated samples. Research into the viability of retrieving real data from synthetic data is a potential research direction.

\find{
{\bf\ding{45}  Key take-aways of [R1]}
\begin{itemize}[leftmargin=*]
    \item Identity leakage is difficult to be assessed precisely, yet majority of the datasets seem to retain relatively low similarity to closest real samples and they visually resemble the real identities.
    \item We have observed potential leakage in  Vec2Face, IDiffFace and IDNet. 
    \item Real datasets contain overlaps, resulting in leakage with respect to CASIA-WebFace being observed in synthetic data not directly trained on it.
\end{itemize}
}

\subsection{R2 - High Intra-class Variability}
To discuss intra-class variability of a dataset, it must offer multiple images per identity. As such, it is not measurable in the case of USynthFace~\cite{boutros2023unsupervised} that proposes only one sample per identity. Even though, one may perceive it to be free of variability-related shortcomings, it has been found~\cite{papantoniou2024arc2face} that a single image per class is not enough to allow for training a robust facial recognition model. 

\begin{figure}[htbp]
    \centering
    \begin{subfigure}[b]{0.33\textwidth}
        \centering               \includegraphics[width=1.0\textwidth]{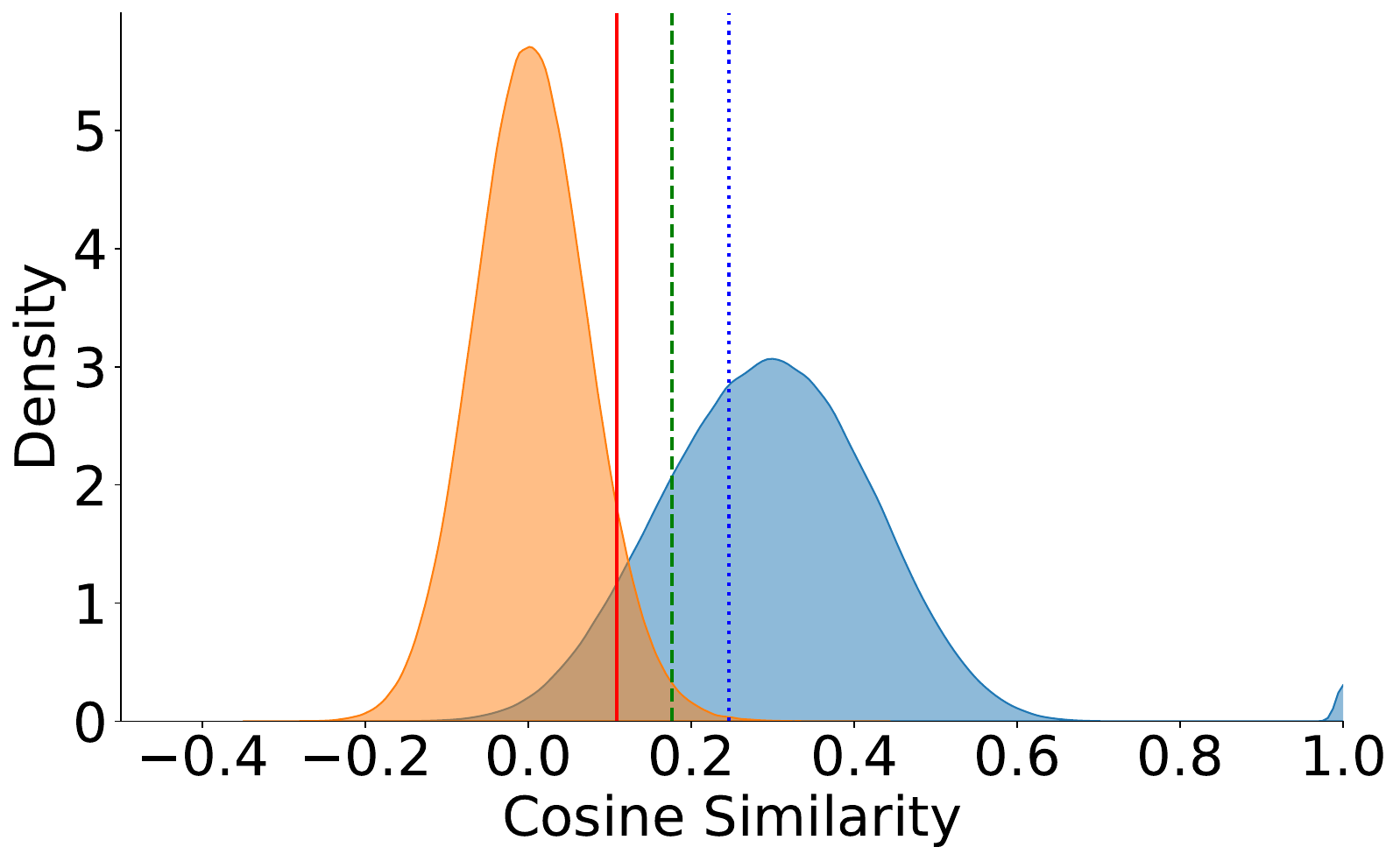} 
        \caption{CemiFace}
        \label{fig:CemiFace}
    \end{subfigure}
    \hfill
    \begin{subfigure}[b]{0.33\textwidth}
        \centering               \includegraphics[width=1.0\textwidth]{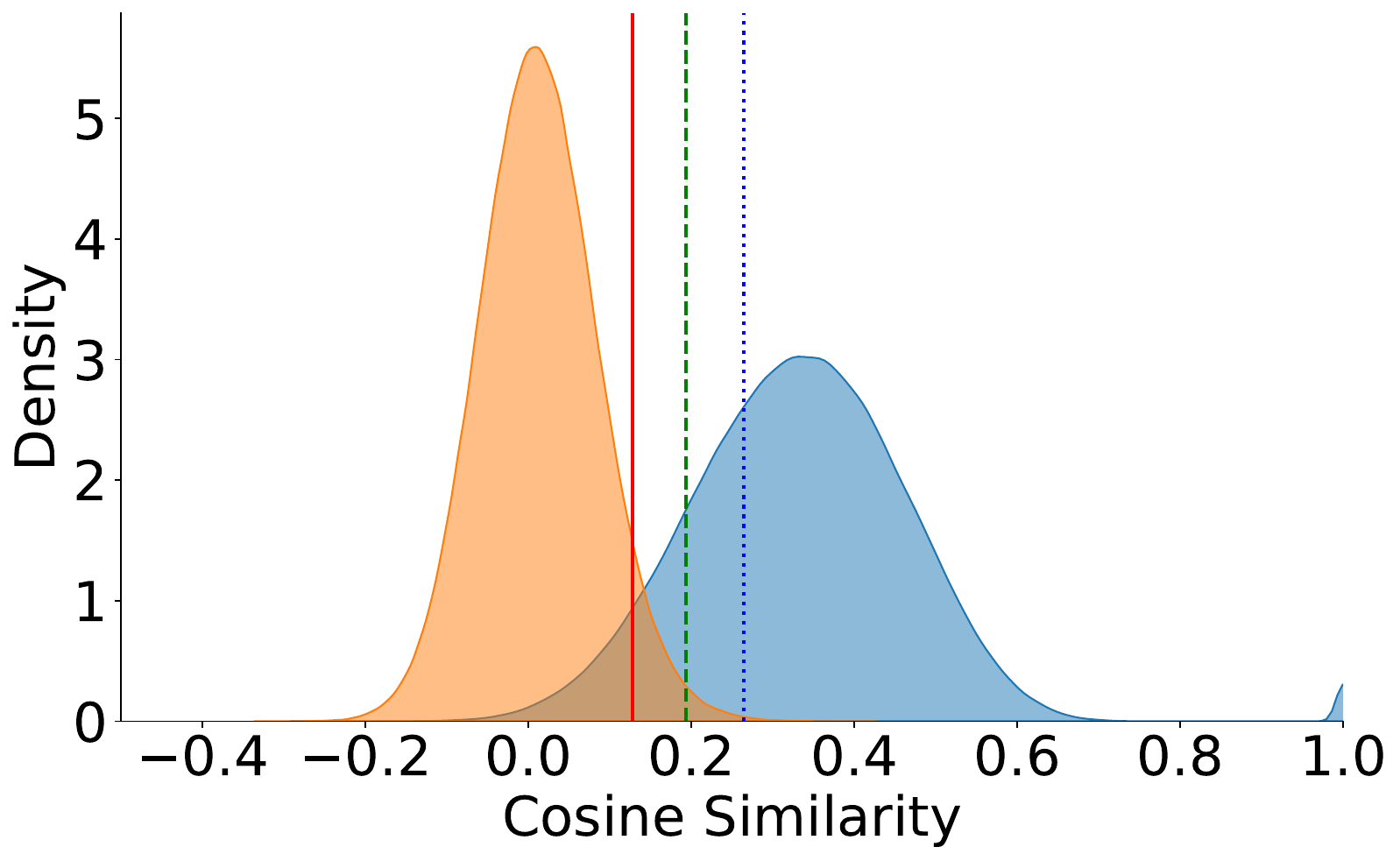} 
        \caption{DCFace}
        \label{fig:DCFace}
    \end{subfigure}
    \hfill
        \begin{subfigure}[b]{0.33\textwidth}
        \centering               \includegraphics[width=1.0\textwidth]{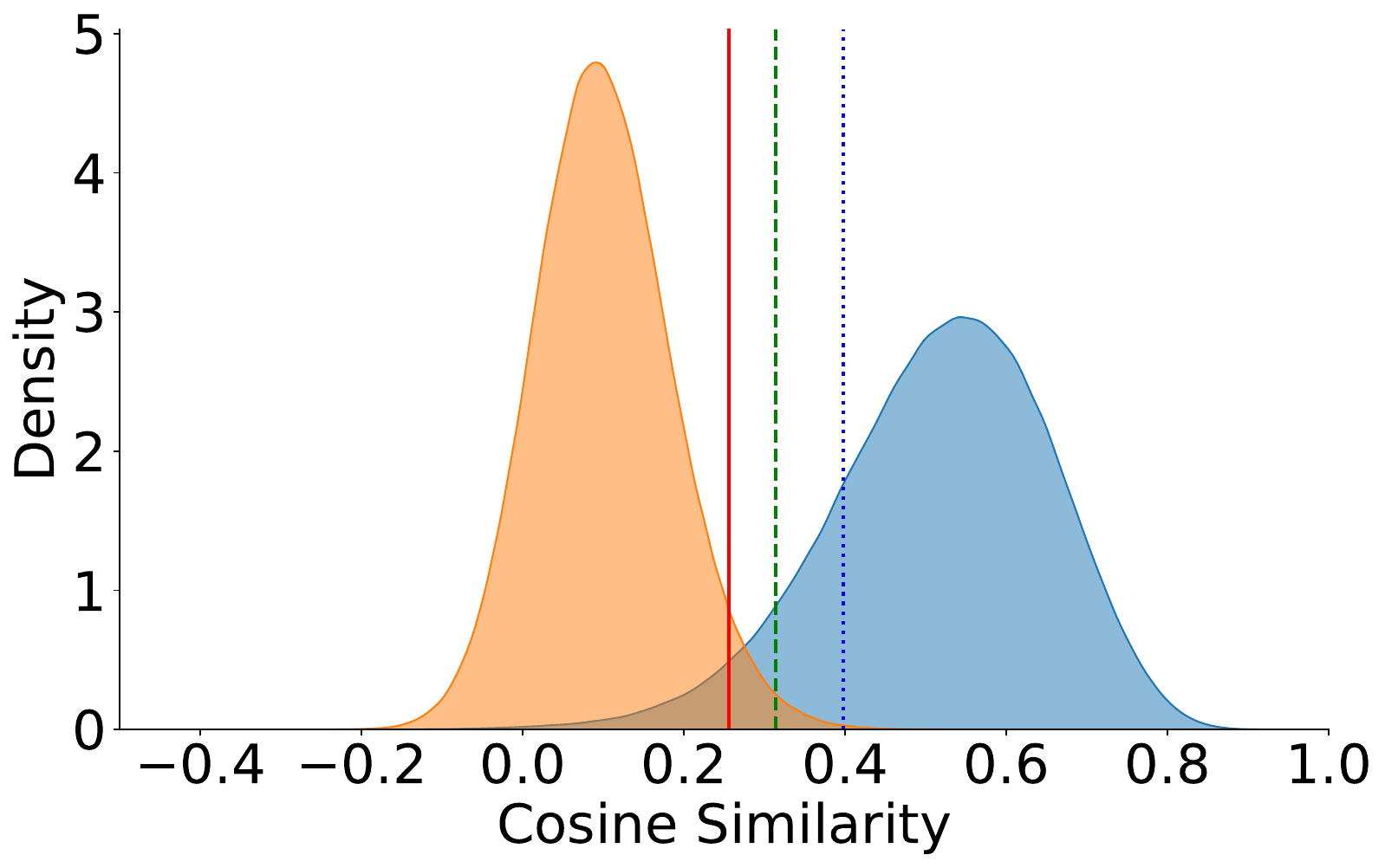} 
        \caption{DigiFace-1M}
        \label{fig:DigiFace}
    \end{subfigure}
    \vfill
        \begin{subfigure}[b]{0.33\textwidth}
        \centering               \includegraphics[width=1.0\textwidth]{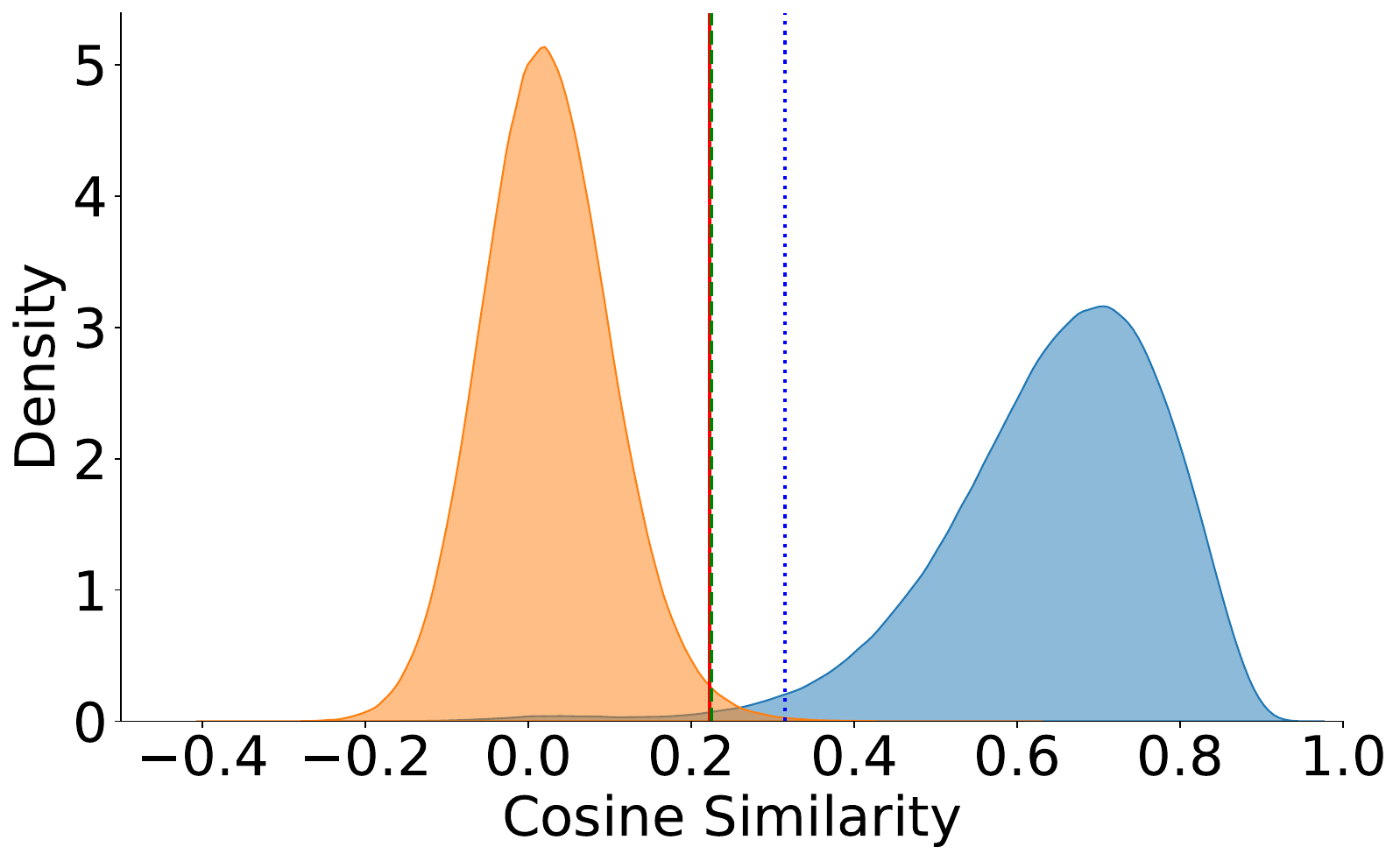} 
        \caption{Digi2Real}
        \label{fig:Digi2Real}
    \end{subfigure}
    \hfill
        \begin{subfigure}[b]{0.33\textwidth}
        \centering               \includegraphics[width=1.0\textwidth]{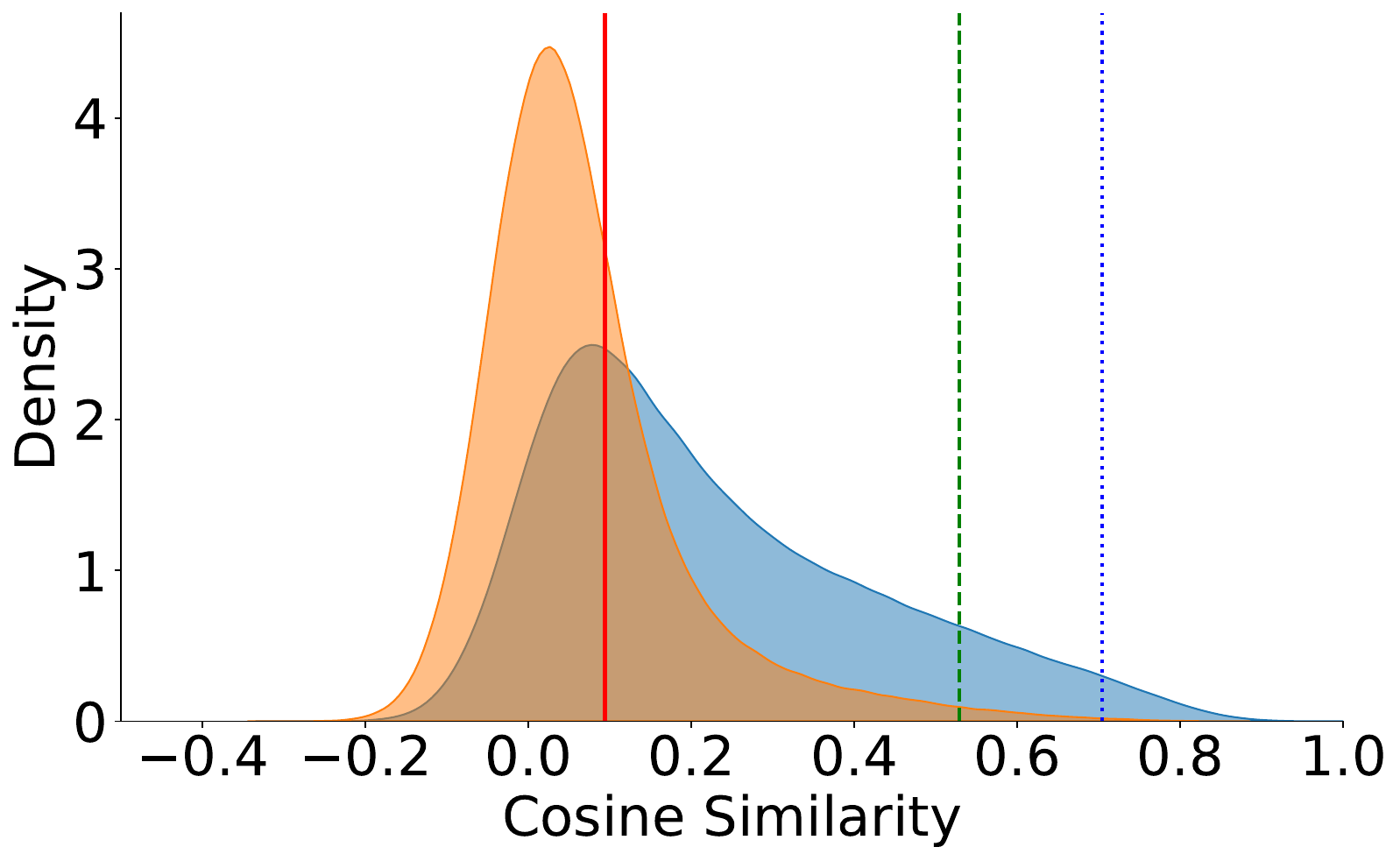} 
        \caption{GANDiffFace}
        \label{fig:GanDiffFace}
    \end{subfigure}
    \hfill
        \begin{subfigure}[b]{0.33\textwidth}
        \centering               \includegraphics[width=1.0\textwidth]{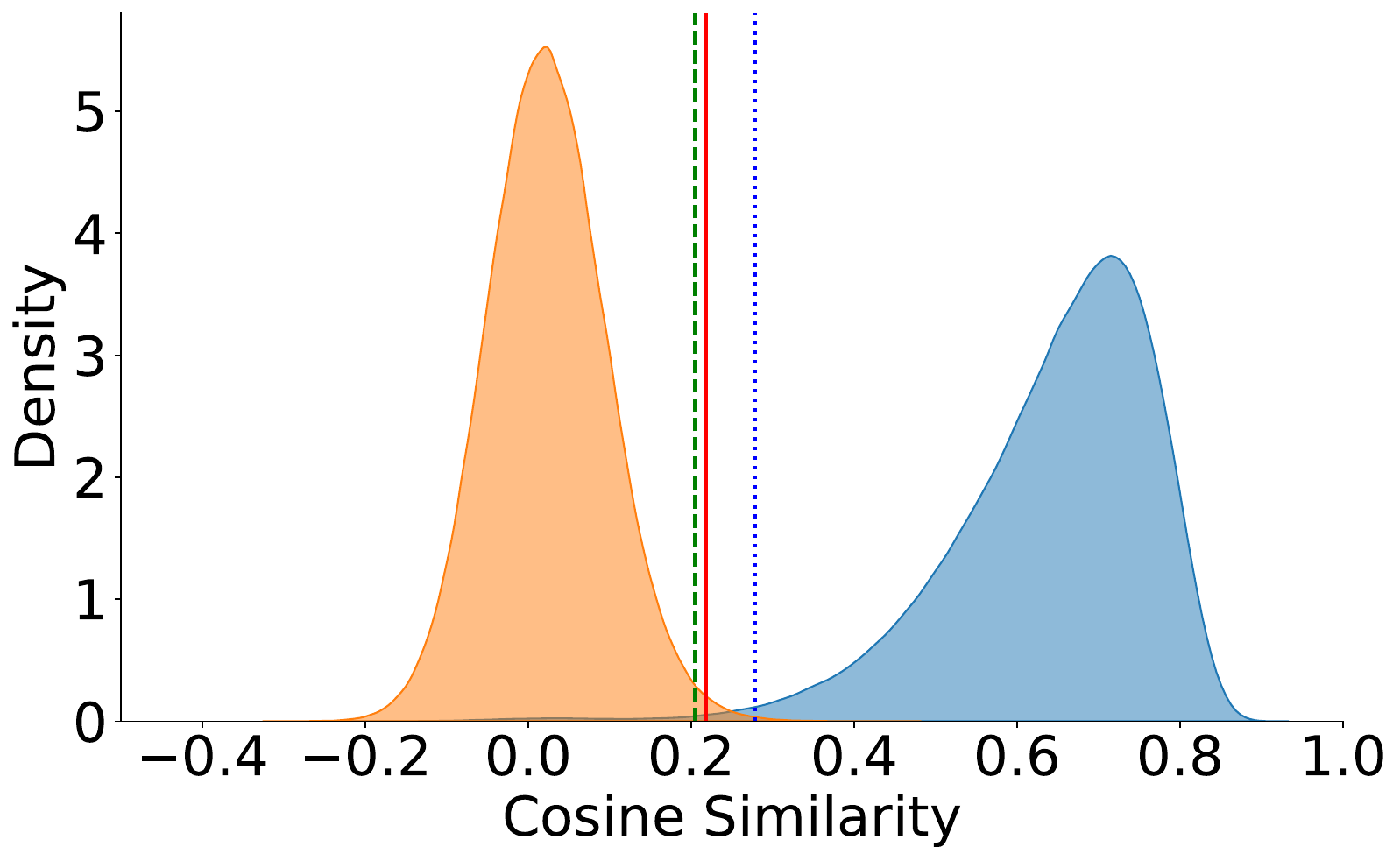} 
        \caption{HyperFace}
        \label{fig:HyperFace}
    \end{subfigure}
    \vfill
        \begin{subfigure}[b]{0.33\textwidth}
        \centering               \includegraphics[width=1.0\textwidth]{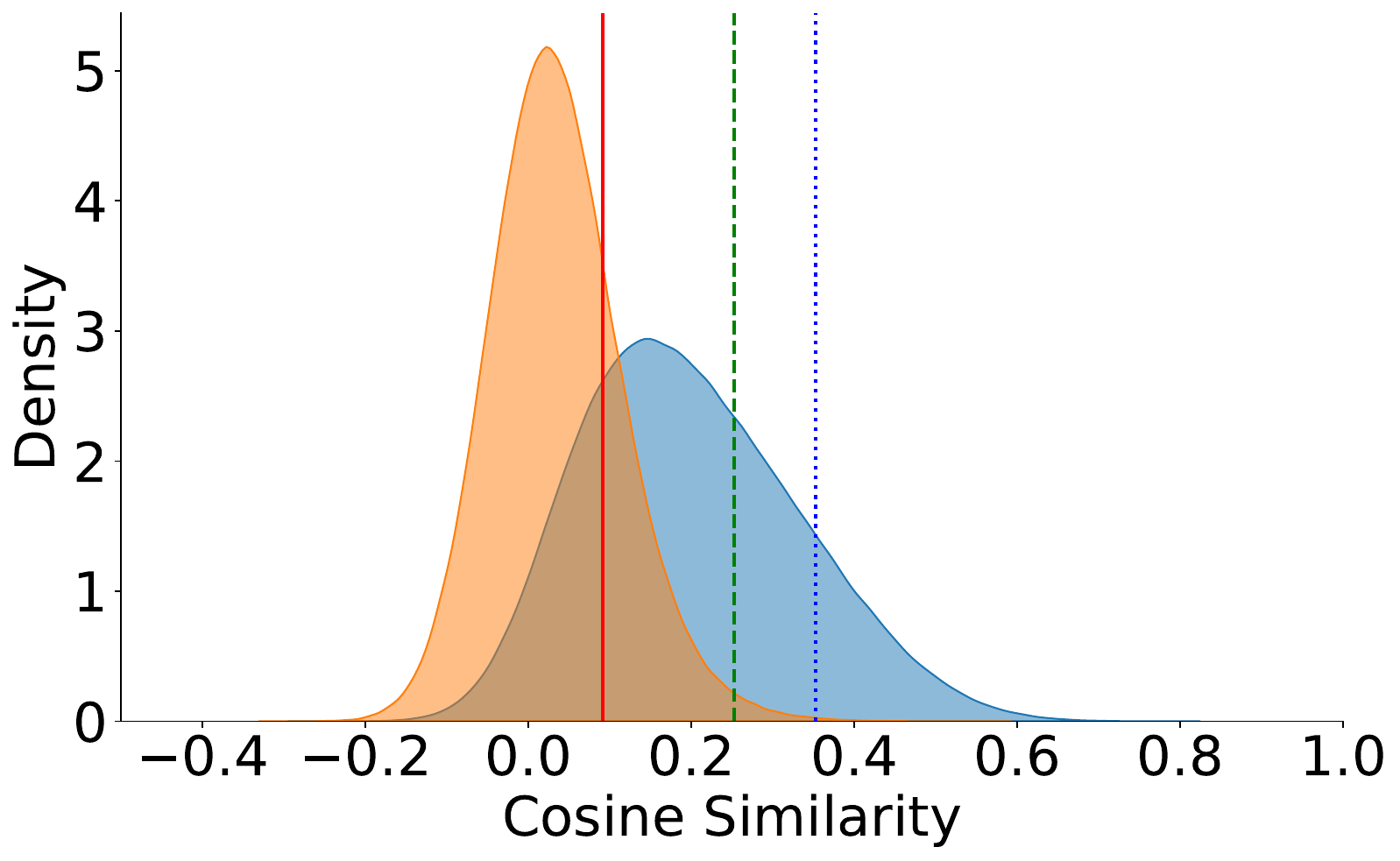} 
        \caption{Langevin-DisCo}
        \label{fig:Langevin}
    \end{subfigure}
    \hfill
        \begin{subfigure}[b]{0.33\textwidth}
        \centering               \includegraphics[width=1.0\textwidth]{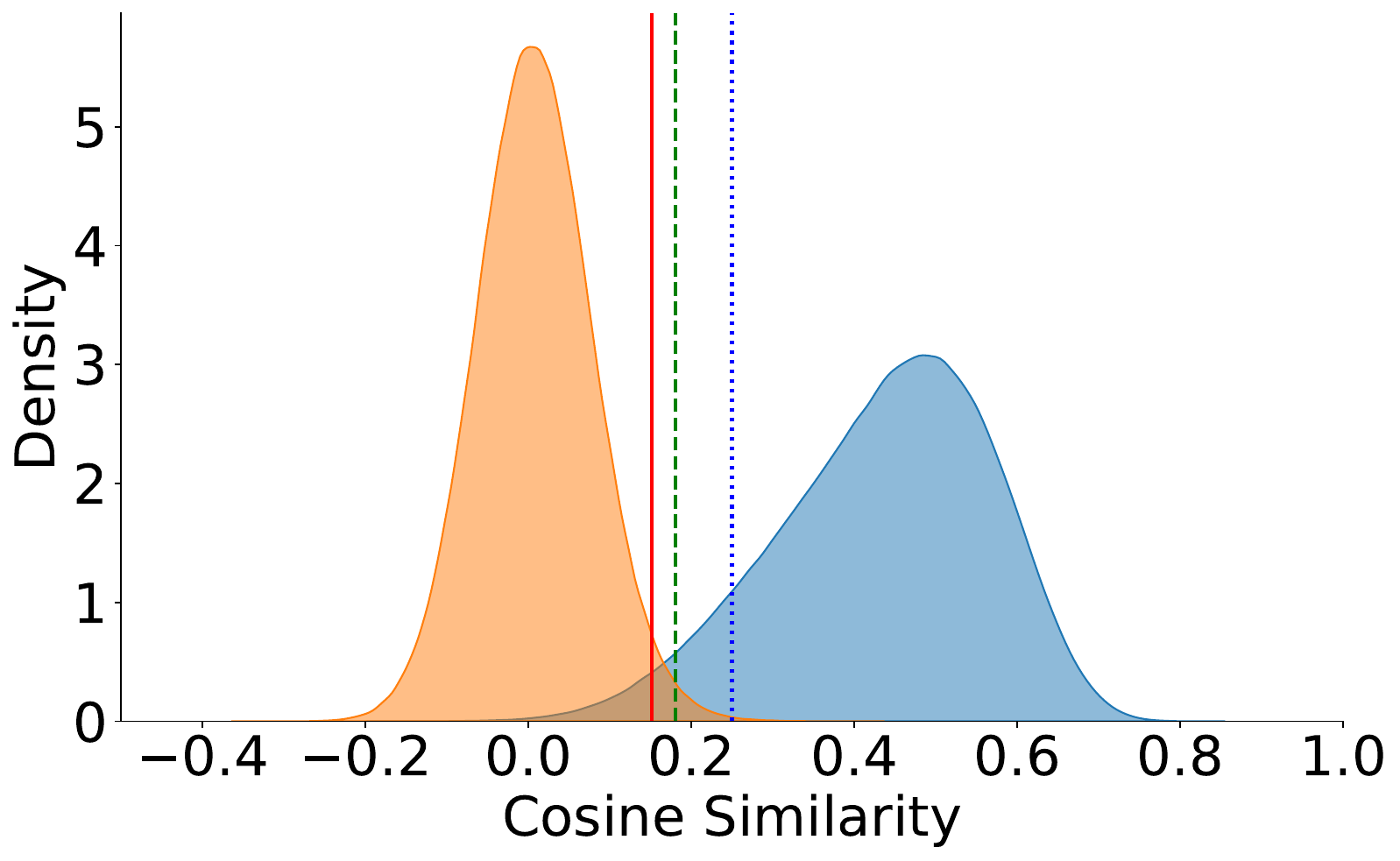} 
        \caption{MorphFace}
        \label{fig:MorphFace}
    \end{subfigure}
    \hfill
        \begin{subfigure}[b]{0.33\textwidth}
        \centering               \includegraphics[width=1.0\textwidth]{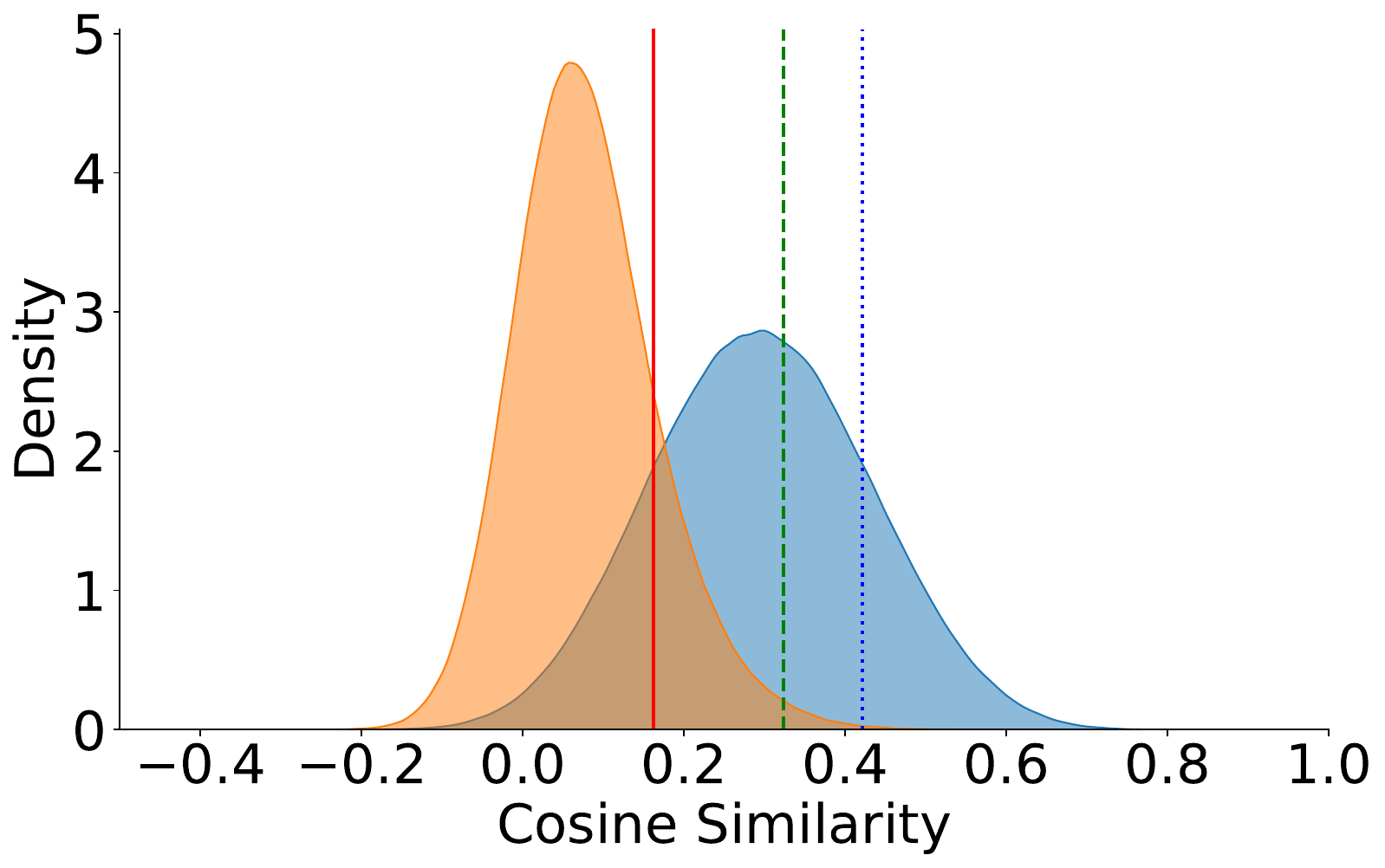} 
        \caption{SFace2}
        \label{fig:SFace2}
    \end{subfigure}
    \vfill
        \begin{subfigure}[b]{0.33\textwidth}
        \centering               \includegraphics[width=1.0\textwidth]{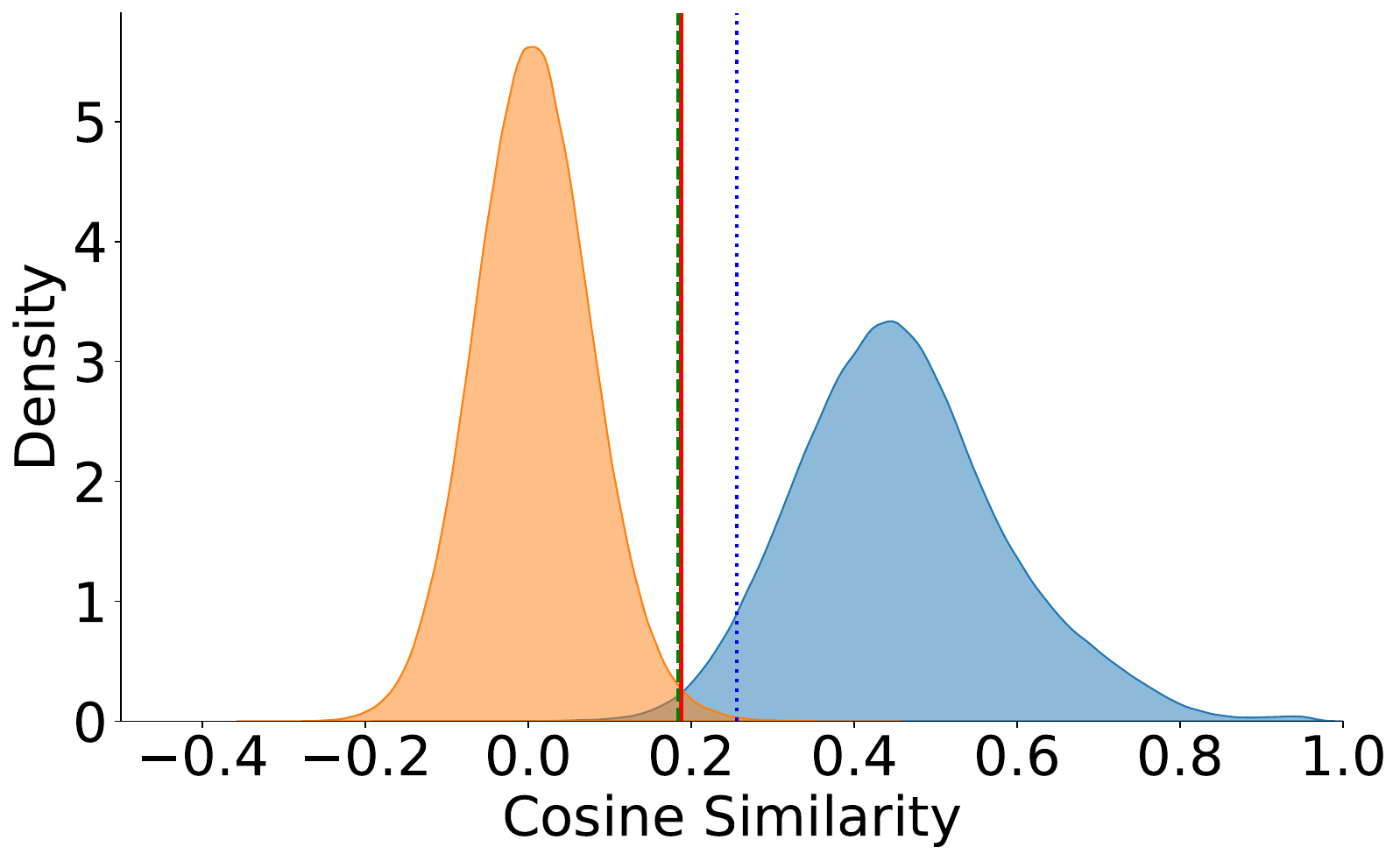} 
        \caption{Vec2Face}
        \label{fig:Vec2Face}
    \end{subfigure}
    \hfill
        \begin{subfigure}[b]{0.33\textwidth}
        \centering               \includegraphics[width=1.0\textwidth]{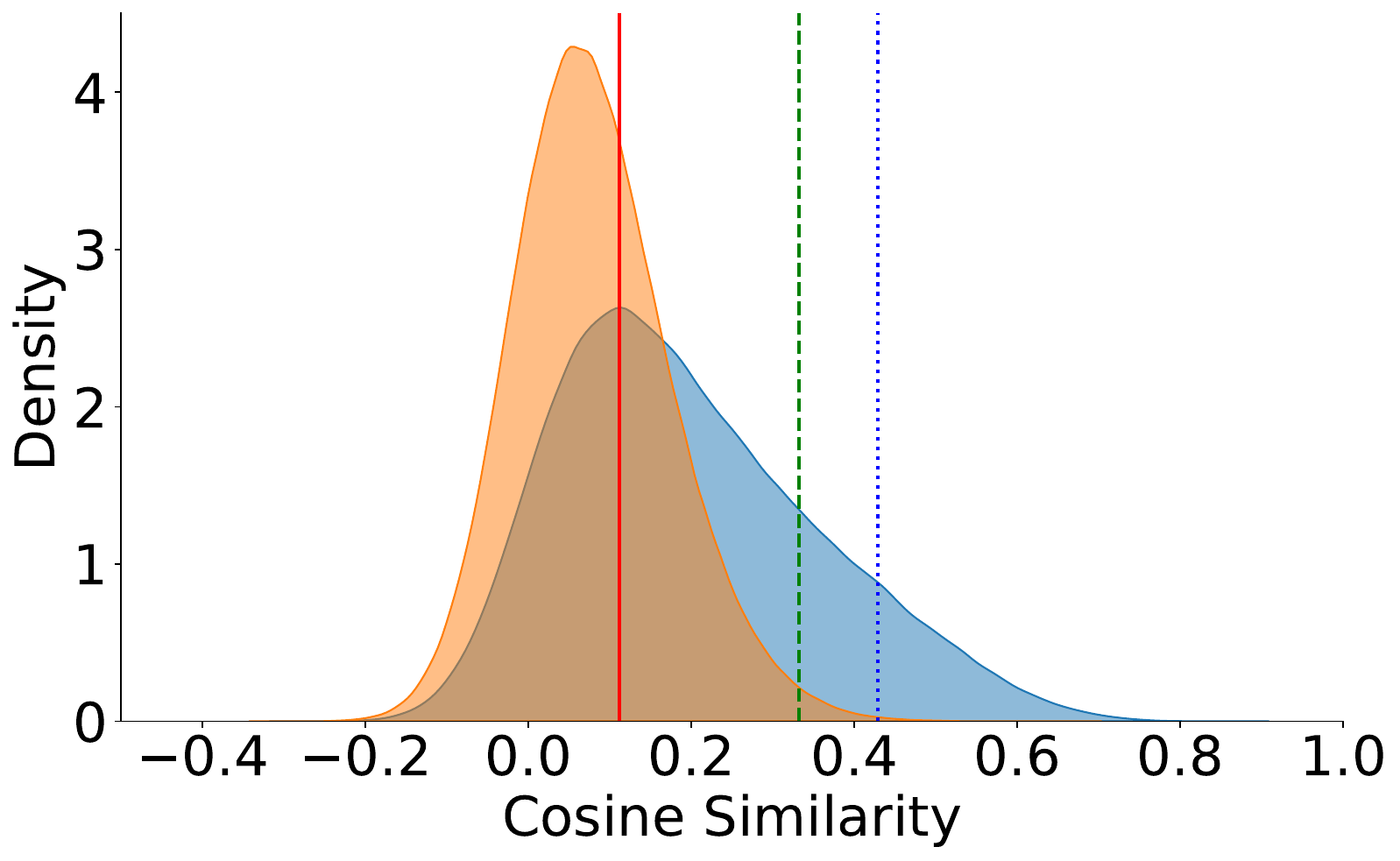} 
        \caption{SynFace}
        \label{fig:SynFace}
    \end{subfigure}
    \hfill
        \begin{subfigure}[b]{0.33\textwidth}
        \centering               \includegraphics[width=1.0\textwidth]{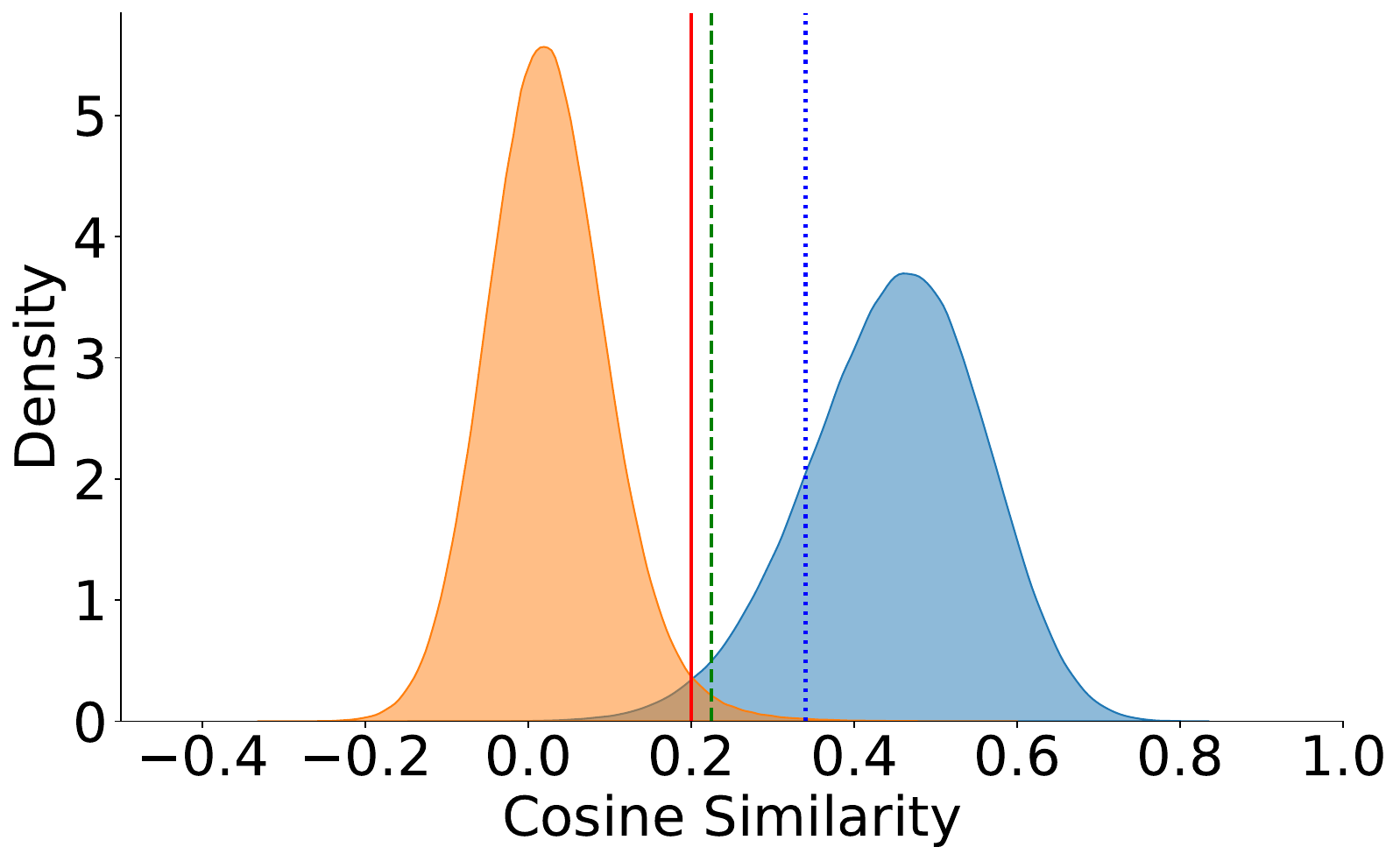} 
        \caption{IDiffFace}
        \label{fig:IDiffFace}
    \end{subfigure}
    \vfill
        \begin{subfigure}[b]{0.33\textwidth}
        \centering               \includegraphics[width=1.0\textwidth]{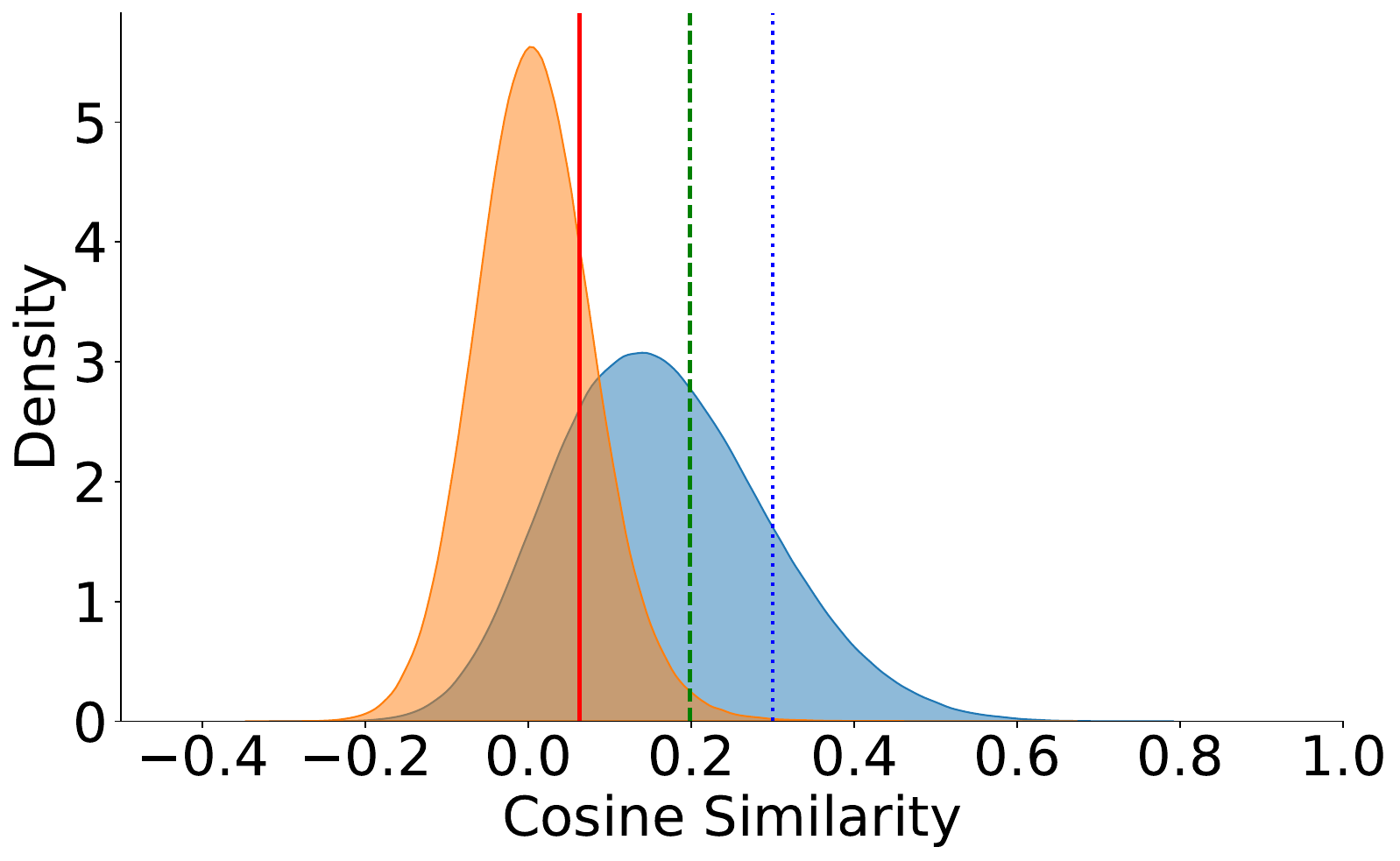} 
        \caption{SFace}
        \label{fig:SFace}
    \end{subfigure}
    \hfill
        \begin{subfigure}[b]{0.33\textwidth}
        \centering               \includegraphics[width=1.0\textwidth]{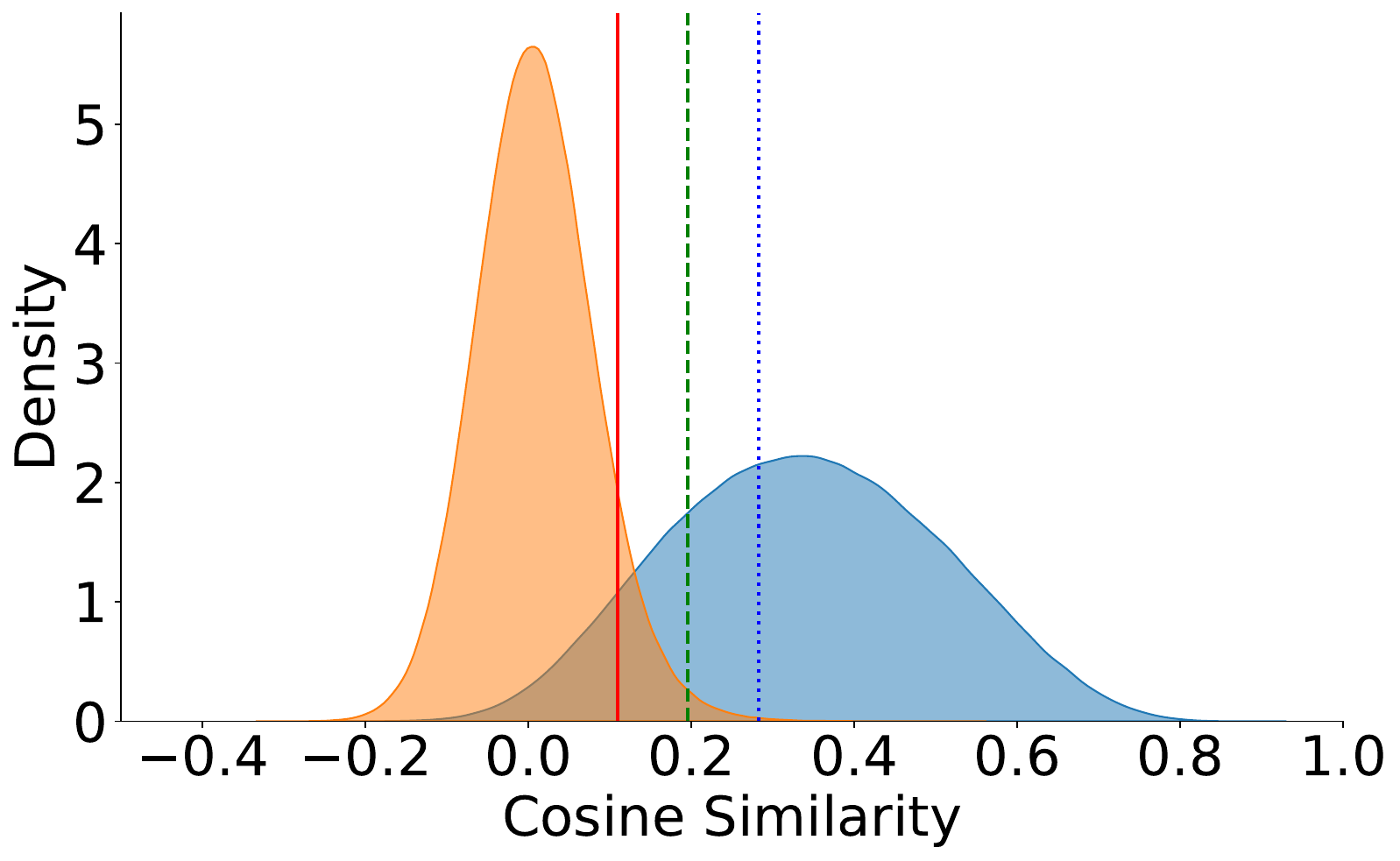} 
        \caption{IDNet}
        \label{fig:IDNet}
    \end{subfigure}
    \hfill
        \begin{subfigure}[b]{0.33\textwidth}
        \centering               \includegraphics[width=0.7\textwidth]{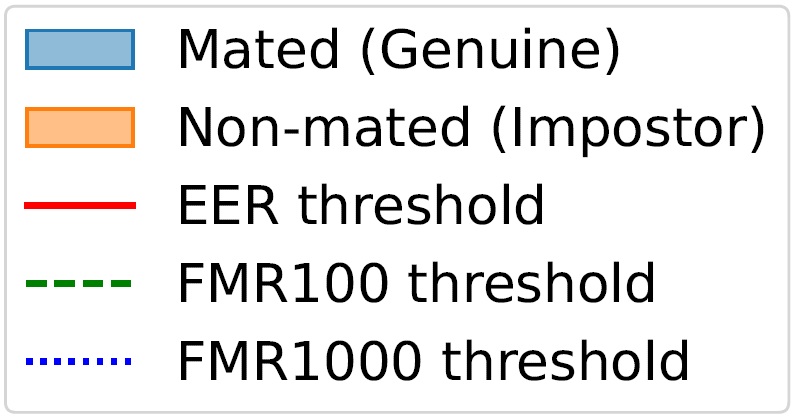}
        \caption*{}
    \end{subfigure}
    \caption{Mated vs non-mated distribution of cosine similarities across datasets with thresholds of key biometric indicators marked}
    \label{fig:MatedVsNonmated}
\end{figure}

To assess the intra-class variance and identity separability in the next subsection, we have evaluated mated (comparison of samples of the same identity) and non-mated (comparison of samples of different identities) distributions based on 1 million comparisons each, across publicly available synthetic datasets (\Cref{fig:MatedVsNonmated}).
Intra-class variance itself can be observed based on how distributed the similarity scores are within mated comparisons. For instance, in case of Digi2Real~\cite{george2025digi2real} and HyperFace~\cite{HyperFace_datasetURL} (Figures~\ref{fig:Digi2Real} and~\ref{fig:HyperFace} respectively), one can observe that the samples belonging to the same identity tend to achieve only very high similarity scores, suggesting the risk that the mated samples may be too similar to each other and as a consequence not constitute as a proper challenge in the learning process~\cite{sun2024cemiface}. On the other hand, if the samples are too diverse, one risks the problem of identity separability, as can be seen (Figures~\ref{fig:GanDiffFace} and~\ref{fig:Langevin}) in the case of GANDiffFace~\cite{melzi2023gandiffface} and Langevin-DisCo~\cite{geissbuhler2024Langevin}. While the dataset consists of a wide variety of samples in terms of similarity to one another, many of the mated samples seem to be too distinct from one another to a degree that they are regarded as a different identity. 
Further, one can observe a common pattern across the best-performing publicly available datasets MorphFace~\cite{mi2025morphFace}, Vec2Face~\cite{Vec2Face_datasetURL}, and CemiFace~\cite{sun2024cemiface} (Figures~\ref{fig:MorphFace},~\ref{fig:Vec2Face}, and~\ref{fig:CemiFace} respectively). Namely, mated samples are quite diverse with relatively low peak values of cosine similarities around 0.4. At the same time, they only slightly overlap with non-mated distributions that focus around 0 and do not tend to produce high value, which can be most easily observed in case of GANDiffFace~\cite{melzi2023gandiffface} (\Cref{fig:GanDiffFace}).

\find{
{\bf\ding{45}  Key take-aways of [R2]}\begin{itemize}[leftmargin=*]
    \item Researchers strive to design datasets with a proper intra-class variability, however, the difficulty lies in the ability to choose an adequate level of diversity --- not too high and not too low.
    \item Best-performing publicly available datasets MorphFace, Vec2Face, and CemiFace showcase diverse mated samples.
    \item Our experiments have shown that GANDiffFace and Langevin-DisCo struggle with too much diversity, while Digi2Real and HyperFace seem to be too biased towards high similarity mated samples.
\end{itemize}
}

\subsection{R3 - Identity Separability}
The earliest versions of privacy-preserving datasets struggled with identity separability, as shown in ~\cite{wu2024vec2face} with respect to IDiff-Face~\cite{boutros2023idiff}, SFace~\cite{boutros2022sface}, DigiFace1-M~\cite{bae2023digiface} and SynFace~\cite{qiu2021synface}. 
For instance, the number of identities in SFace~\cite{boutros2022sface} and IDNet~\cite{kolf2023idnet} datasets cannot be larger than in the training dataset~\cite{boutros2023exfacegan} due to the choice of class-conditional GAN as a generation method.
What is more, works based on DiscoFaceGAN~\cite{deng2020discofacegan}, such as SynFace~\cite{qiu2021synface}, are unable to produce more than 500 unique identities~\cite{kim2024vigface}. Further, IDiffFace~\cite{boutros2023idiff} was also reported to be unable to ensure synthetic identity uniqueness~\cite{kim2024vigface}. 
Our own experiments have confirmed issues in SFace~\cite{boutros2022sface} (\Cref{fig:SFace}) and SynFace~\cite{qiu2021synface} (\Cref{fig:SynFace}), and also uncovered separability problems within GANDiffFace~\cite{GANDiffFace_datasetURL} (\Cref{fig:GanDiffFace}), Langevin-DisCo~\cite{geissbuhler2024Langevin} (\Cref{fig:Langevin}) and to a lesser extent SFace2~\cite{boutros2024sface2}. As shown in the figures, the overlap areas for mated and non-mated samples take significant parts of the distributions, suggesting that many of the samples outside of the given identity are more similar than mated images. 

On the other hand, thanks to novel advances Arc2Face~\cite{papantoniou2024arc2face} and DCFace~\cite{kim2023dcface} have been reported to achieve almost an identical level of separability as the real dataset WebFace4M~\cite{zhu2022webface260m}, while Vec2Face~\cite{wu2024vec2face} has even exceeded it. The superior performance in that matter is attributed to greater control over used identities, since the real dataset contains images of some closely related people and minor label noise. Within the scope of our assessment, we have confirmed that the DCFace~\cite{kim2023dcface} and Vec2Face~\cite{wu2024vec2face} achieve proper separability (Figures~\ref{fig:DCFace} and~\ref{fig:Vec2Face} respectively). Additionally, we have observed similar positive patterns for CemiFace~\cite{sun2024cemiface} (\Cref{fig:CemiFace}), Digi2Real~\cite{george2025digi2real} (\Cref{fig:Digi2Real}), HyperFace~\cite{shahreza2024hyperface} (\Cref{fig:HyperFace}), IDiffFace~\cite{boutros2023idiff} (\Cref{fig:IDiffFace}), IDNet~\cite{boutros2023idiff} (\Cref{fig:IDNet}) and MorphFace~\cite{mi2025morphFace} (\Cref{fig:MorphFace}). Contrary to previous findings~\cite{wu2024vec2face}, identities in DigiFace-1M~\cite{bae2023digiface} seem to be properly separated.

\begin{table}[!ht]

    \centering
    \small
    \caption{Genuine vs impostor comparison of synthetic facial recognition datasets. The best results are indicated by bold font, the second best by underlining.}
    \label{tab:biometric_metrics}
    \begin{tabular}{ccccccccc}
    \toprule
        Dataset & \makecell{Genuine \\ $\text{Mean}_{\pm \text{Std}}$} & \makecell{Impostor \\ $\text{Mean}_{\pm \text{Std}}$} & EER & FMR100 & FMR1000 & FDR \\ \midrule
        CemiFace~\cite{sun2024cemiface} & $0.296_{\pm0.136}$ & $0.004_{\pm0.071}$ & 0.073 & 0.183 & 0.355 & 3.632
 \\ 
        ControlFace10k~\cite{nzalasse2025sig} & $0.667_{\pm 0.094}$ & $0.624_{\pm 0.105}$ & 0.408 & 0.980 & 0.999 & 0.093 \\ 
        DCFace~\cite{kim2023dcface} & $0.333_{\pm 0.137}$ & $0.013_{\pm 0.073}$ & 0.060 & 0.146 & 0.303 & 4.280 \\ 
        DigiFace-1M~\cite{bae2023digiface} & $0.516_{\pm 0.136}$ & $0.100_{\pm 0.085}$ & 0.039 & 0.078 & 0.185 & 6.705 \\ 
        Digi2Real~\cite{george2025digi2real} & $0.644_{\pm 0.139}$ & $0.024_{\pm 0.080}$ & 0.011 & 0.011 & \underline{0.022} & \underline{15.043} \\ 
        GANDiffFace~\cite{melzi2023gandiffface} & $0.226_{\pm 0.206}$ & $0.068_{\pm 0.127}$ & 0.315 & 0.895 & 0.975 & 0.423 \\ 
        HyperFace~\cite{shahreza2024hyperface} & $0.643_{\pm 0.125}$ & $0.023_{\pm 0.074}$ & \textbf{0.007} & \textbf{0.006} & \textbf{0.012} & \textbf{18.204} \\ 
        IDiffFace~\cite{melzi2023gandiffface} & $0.448_{\pm 0.108}$ & $0.025_{\pm 0.075}$ & 0.017 & 0.027 & 0.161 & 10.310 \\ 
        IDNet~\cite{kolf2023idnet} & $0.340_{\pm 0.164}$ & $0.010_{\pm 0.073}$ & 0.084 & 0.207 & 0.379 & 3.367 \\ 
        Langevin-DisCo~\cite{geissbuhler2024Langevin} & $0.200_{\pm 0.134}$ & $0.035_{\pm 0.081}$ & 0.226 & 0.670 & 0.859 & 1.101 \\ 
        MorphFace~\cite{mi2025morphFace} & $0.436_{\pm 0.131}$ & $0.008_{\pm 0.071}$ & 0.025 & 0.039 & 0.096 & 8.236 \\ 
        SFace~\cite{boutros2022sface} & $0.165_{\pm 0.128}$ & $0.009_{\pm 0.074}$ & 0.222 & 0.622 & 0.849 & 1.115 \\ 
        SFace2~\cite{boutros2024sface2} & $0.296_{\pm 0.135}$ & $0.080_{\pm 0.089}$ & 0.166 & 0.581 & 0.819 & 1.795 \\ 
        SynFace~\cite{qiu2021synface} & $0.198_{\pm 0.163}$ & $0.079_{\pm 0.097}$ & 0.343 & 0.792 & 0.898 & 0.392 \\ 
        Syn-Multi-PIE~\cite{SynMulti-PIE_datasetURL} & $0.537_{\pm 0.185}$ & $0.119_{\pm 0.101}$ & 0.078 & 0.238 & 0.438 & 3.942 \\ 
        Vec2Face~\cite{wu2024vec2face} & $0.458_{\pm 0.129}$ & $0.007_{\pm 0.072}$ & \underline{0.009} & \underline{0.008} & 0.044 & 9.372 \\
    \bottomrule
    \end{tabular}
\end{table}

To enrich our analysis of the separability patterns, we have also collected statistical data concerning the distributions shown in Figure~\ref{fig:MatedVsNonmated}, specifically useful in assessing the applicability of datasets in real-life biometric contexts.
Those include Equal Error Rate (EER), False Match Rate (FMR), False Non-Match Rate (FNMR) and Fisher Discriminant Ratio (FDR). FMR describes the percentage of impostor images misclassified, while FNMR measures the opposite scenario when positive samples are predicted as impostors. FMR100 and FMR1000 measure FNMR for a situation when FMR is 1/100 and 1/1000, respectively - in other words, how many genuine samples are rejected when 1 in 100/1000 impostor samples is misclassified. EER is the error rate in the situation when FMR and FNMR are equal. The lower the values of the aforementioned metrics, the better the performance of the underlying model. 
Finally, FDR measures separation between genuine and impostor classes by dividing the squared difference of means by the sum of squares of variances. The higher the FDR value, the better the separability.  

According to our experiments (\Cref{tab:biometric_metrics}), the most impressive results were achieved by HyperFace~\cite{shahreza2024hyperface} with EER of 0.007, FMR100 of 0.006, FMR1000 of 0.012 and FDR of 18.204, proving excellent separability. Vec2Face~\cite{Vec2Face_datasetURL} and Digi2Real~\cite{george2025digi2real} come close behind, with Vec2Face performing slightly better in terms of EER and FMR100, but showing worse results for FMR1000 and FDR. It is worth noting that Vec2Face displays a significantly lower genuine mean than Digi2Real and HyperFace (0.46 vs 0.64), and comparable impostor mean (0.01 vs 0.02), achieving proper separability with less diverse samples. Exceptionally high EER values, above 0.2, have been observed for ControlFace10k~\cite{nzalasse2025sig}, GANDiffFace~\cite{melzi2023gandiffface}, Langevin-DisCo~\cite{geissbuhler2024Langevin}, SFace~\cite{boutros2022sface} and SynFace~\cite{SynFace_datasetURL}. Those datasets also display FMR100 values that would make them unreliable in real-life applications. 

\find{
{\bf\ding{45}  Key take-aways of [R3]}\begin{itemize}[leftmargin=*]
    \item Considering real-life biometric applications of synthetic datasets, HyperFace presents the most impressive results in terms of EER (0.007), FMR100 (0.006), FMR1000 (0.012) and FDR (18.204) metrics.
    \item Vec2Face and Digi2Real follow closely behind, proving proper identity separability.
    \item Datasets such as ControlFace10k, GANDiffFace, SynFace and SFace do not provide sufficient difference between mated and non-mated samples.
\end{itemize}
}

\subsection{R4 - High Image Count}
The majority of publicly available synthetic datasets offer from 0.5M to 1M images of around 10k identities, while the largest variant of Vec2Face contains 20M of 400k images. These numbers match or even exceed the size of CASIA-WebFace~\cite{yi2014casia}, yet are clearly smaller than WebFace260M~\cite{zhu2022webface260m}, consisting of 260M images and 4M identities. However, label noise is a significant problem that real datasets struggle with, especially at that scale, pointing our attention towards the notion of quality vs quantity. While synthetic datasets are less prone to label noise, throughout our own assessment of datasets, we have observed instances of deliberate label noise in CemiFace (\Cref{fig:CemiFace}) and DCFace 
(\Cref{fig:DCFace}). 
As shown in the subfigures, in both cases, there is a sudden spike of cosine similarities equal to 1. It is caused by the existence of multiple samples of the same image within the same identity folder, but under different filenames. In the case of DCFace, the authors intentionally introduced 5 repeated samples per identity to address lower label consistency with respect to real data, while CemiFace mimicked this approach.



\find{
{\bf\ding{45}  Key take-aways of [R4]}\begin{itemize}[leftmargin=*]
    \item With synthetic datasets frequently  exceeding 1M samples, high image count requirement can be considered as fulfilled.
    \item The available sizes of synthetic datasets are sufficient to compete with and in some cases even surpass the performance of models trained on CASIA-WebFace.
    \item Synthetic datasets allow to eliminate the problem of label noise affecting real data. Exceptionally, DCFace and CemiFace introduce deliberate label noise by repeating specific samples.
\end{itemize}
}

\subsection{R5 - Ethical Data Sourcing}
Synthetic datasets are most commonly created using previously mentioned methods trained on real data from datasets such as CASIA-WebFace~\cite{yi2014casia}, WebFace4M~\cite{zhu2022webface260m} and FFHQ~\cite{karras2019ffhq}. 
CASIA-WebFace~\cite{yi2014casia} contains 494,414 face images of 10,575 identities, collected from IMDb without individuals' consent via semi-automated crawling. As such the data is legally and ethically questionable.
WebFace4M~\cite{zhu2022webface260m} is a cleaned, significantly smaller subset of WebFace260M. It contains 4M images of celebrities downloaded from Google image search engine based on names retrieved from IMDb and Freebase~\footnote{https://developers.google.com/freebase/} knowledge graph. Similarly to CASIA-WebFace, data was acquired without consent.
FFHQ~\cite{karras2019ffhq} consists of 70000 images, under permissive licenses, collected from Flickr~\footnote{https://www.flickr.com/}. Since image authors have agreed to redistribution and modification by their choice of a license, any ethical and legal claims made against the dataset are significantly less substantial. 
While FFHQ can be considered as a significant step forward in terms of usage of data acquired with consent, an interesting research direction would be collection of a real demographically balanced dataset with an explicit consent, as in case of face scans used to train DigiFace-1M~\cite{bae2023digiface}, or even invention of a method that would not require any real data at all.

\find{
{\bf\ding{45}  Key take-aways of [R5]}\begin{itemize}[leftmargin=*]
    \item Real data is still needed to produce synthetic samples, with many datasets used for that purpose being assembled via unconsented web crawling.
    \item Commonly used, more ethical, alternative in a form of FFHQ provides images under permissive licenses.
    \item To fully address ethical concerns, there is a need for a demographically balanced dataset acquired fully with consent, or to create an image synthesis method fully independent from real data.
\end{itemize}
}

\subsection{R6 - Bias Mitigation}

It has been suggested~\cite{yeung2024variface, leyva2024demographic} that the synthetic datasets inherit biases from the real data they are based on. While some researchers have decided to explicitly address the issue, unfortunately, the open-source datasets that achieve the best benchmark results -- Vec2Face~\cite{wu2024vec2face} and CemiFace~\cite{sun2024cemiface} -- do not. On the other hand, VariFace~\cite{yeung2024variface}, presenting the best results so far, actively reduces demographic biases, but it is not available to the public. 
While the current standard testing benchmark set contains datasets such as AgeDB~\cite{moschoglou2017agedb} and CALFW~\cite{zheng2017calfw} that can help assess age biases, only some of the datasets are tested on RFW~\cite{wang2019rfw} against racial biases and gender biases tend to be overlooked.

Age biases can be analyzed using already introduced benchmarks in Tables~\ref{tab:benchmark_unrestricted} and ~\ref{tab:comparison_with_real}. It can be observed that all models (based on both real and synthetic data), in relative terms, achieve significantly lower performance than on base LFW~\cite{huang2008lfw} when evaluated on AgeDB~\cite{moschoglou2017agedb} and CALFW~\cite{zheng2017calfw}. This drop is especially evident for older synthetic datasets. When the best-performing synthetic dataset is considered, one can observe that VariFace~\cite{yeung2024variface} surpasses results obtained by real GlintAsian~\cite{an2021glint360k}, Casia-WebFace~\cite{yi2014casia} and VGGFace2~\cite{cao2018vggface2}, however, it comes short when compared with multiple newer datasets. As other well-performing synthetic datasets offer slightly lower resistance to temporal changes, it is important to pay more attention to this issue in the future.

\begin{table}[h]
    \caption{Accuracy comparison of IResNet50 models trained on synthetic or real data, evaluated on RFW. The best results are indicated by bold font, the second best by underlining.}
    \label{tab:rfw}
    \centering
    \begin{tabular}{c|c|c|c|c|c|c}
    \toprule
        Dataset & Data & African & Asian & Caucasian & Indian & Average \\ \midrule
        MSCeleb-1M~\cite{guo2016ms} & Real & \textbf{0.9832} & \textbf{0.9785} & \textbf{0.9912} & \textbf{0.9790} & \textbf{0.9829} \\ 
        WebFace-4M~\cite{zhu2022webface260m} & Real & \underline{0.9763} & \underline{0.9710} & \underline{0.9898} & \underline{0.9772} & \underline{0.9786} \\ 
        VariFace~\cite{yeung2024variface} & Synthetic & 0.9363 & 0.9180 & 0.9590 & 0.9332 & 0.9366 \\
        CASIA-WebFace~\cite{yi2014casia} & Real & 0.8835 & 0.8632 & 0.9420 & 0.8912 & 0.8949 \\ 
        Vec2Face~\cite{wu2024vec2face} [\ding{70}] & Synthetic & 0.8415 & 0.8535 & 0.9028 & 0.8750 & 0.8682 \\ 
        DCFace~\cite{kim2023dcface} [\ding{72}] & Synthetic & 0.8073 & 0.8403 & 0.8950 & 0.8720 & 0.8537 \\ 
        LangevinDisCo~\cite{geissbuhler2024Langevin} & Synthetic & 0.7953 & 0.8328 & 0.8865 & 0.8083 & 0.8307 \\ 
        Digi2Real-20K~\cite{george2025digi2real}  & Synthetic & 0.8067 & 0.7995 & 0.8533 & 0.8140 & 0.8184 \\ 
        IDiffFace~\cite{boutros2023idiff} [\ding{72}] & Synthetic & 0.7523 & 0.7987 & 0.8520 & 0.8068 & 0.8025  \\ 
        GANDiffFace~\cite{melzi2023gandiffface} [\ding{72}] & Synthetic & 0.6460 & 0.7072 & 0.7538 & 0.7155 & 0.7056 \\ 
        DigiFace-1M~\cite{bae2023digiface} [\ding{72}] & Synthetic & 0.6555 & 0.6960 & 0.7258 & 0.7000 & 0.6943 \\ 
        SFace~\cite{boutros2022sface} [\ding{72}] & Synthetic & 0.6460 & 0.6968 & 0.7432 & 0.6787 & 0.6912 \\ 
        IDNet~\cite{kolf2023idnet} [\ding{71}] & Synthetic & 0.5930 & 0.6422 & 0.7003 & 0.6577 & 0.6483 \\ 
        SynFace~\cite{qiu2021synface} [\ding{71}] & Synthetic & 0.5727 & 0.6448 & 0.6560 & 0.6148 & 0.6221 \\ 
        ExFaceGAN~\cite{boutros2023exfacegan} [\ding{72}] & Synthetic & 0.5563 & 0.6382 & 0.6462 & 0.6218 & 0.6156 \\ \bottomrule
    \end{tabular}

Indicated results obtained from reproduced experiments~\cite{yeung2024variface}[\ding{70}],~\cite{geissbuhler2024Langevin}[\ding{71}] and~\cite{george2025digi2real}[\ding{72}] due to data unavailability in respective original studies.

\end{table}

To assess the racial bias in synthetic-based face recognition, we gathered the results on Racial Faces in-the-wild (RFW) benchmark (\Cref{tab:rfw}), available in~\cite{shahreza2024sdfr, geissbuhler2024Langevin, george2025digi2real, yeung2024variface}. The RFW dataset contains four testing subsets corresponding to African, Asian, Caucasian, and Indian groups and includes approximately 80k images from 12k identities.
Analyzing the results, it can be observed that the majority of synthetic datasets achieve inferior performance on the RFW, across all ethnicities, when compared with models trained on real data. However, VariFace~\cite{yeung2024variface} has managed to outperform CASIA-WebFace~\cite{yi2014casia}, proving the remarkable potential of the synthetic data. Regardless of the used dataset, a bias towards Caucasian individuals can always be observed. Models trained on synthetic datasets typically struggle the most with recognizing African individuals, while the ones trained on real data underperform the most for Asian samples. Interestingly, VariFace~\cite{yeung2024variface} is the only synthetic dataset to exhibit the property of real datasets by underperforming for Asian individuals.

While an effort to address biases can be observed in multiple datasets, the requirement fulfillment level varies from dataset to dataset. Nonetheless, there is a need for a publicly available dataset that reaches state-of-the-art performance and, at the same time, addresses underlying biases. Consequently, all newly created datasets should not only strive to achieve superior performance, but also ensure that the biases are properly addressed.

\find{
{\bf\ding{45}  Key take-aways of [R6]}\begin{itemize}[leftmargin=*]
    \item Some methods, such as VariFace, address demographic biases explicitly. Nonetheless, many of existing synthetic datasets do not consider them as a part of their design.
    \item Evaluations on RFW indicate that, similarly to real datasets, synthetic counterparts are the most effective in recognizing Caucasian individuals.
    \item Evaluated real datasets struggle the most with recognizing Asian individuals, while synthetic tend to underperform the most for African individuals.
\end{itemize}
}

\subsection{R7 - Synthetic Benchmarks}
In recent years three synthetic benchmark datasets have been proposed -
ControlFace10k~\cite{nzalasse2025sig}, Syn-Multi-PIE~\cite{colbois2021synmultipie} and SDFD~\cite{baltsou2024sdfd}. While these works lead us towards a fully synthetic evaluation of facial recognition models, they have one common problem. The main shortcoming we have identified is an insufficient amount of experiments to determine the real-life performance of proposed datasets in terms of their benchmarking capabilities on multiple models and against relevant benchmarking datasets containing real data. Such extensive tests should always be performed before one can be sure of their applicability. Since no relevant benchmarking procedure has been established yet, we propose the following approach:

\begin{enumerate}[leftmargin=*]
\item[\ding{182}] \textit{Model selection for evaluation:} It is essential to consider a wide range of models to be evaluated, differing in architectures and loss functions to reflect their diversity.
Architectures should include vision transformers (e.g., SwinFace~\cite{qin2023swinface}) and CNN-based backbones (e.g., ResNet~\cite{he2016resnet}). Loss functions should consider margin-based softmax (e.g., ArcFace~\cite{deng2019iresnet50}, CosFace~\cite{wang2018cosface}), metric-learning losses (e.g., triplet loss~\cite{schroff2015facenet}) and possibly hybrid approaches.
Widespread availability and use of pre-trained models within so-called model zoos~\cite {InsightFaceZoo} or easily implementable frameworks such as DeepFace~\cite{serengil_deepface} allows for an ease of access and execution of this step. Additionally, models present in such repositories are often already assessed on relevant benchmarks, further reducing the procedure's complexity in the next steps.

\item[\ding{183}] \textit{Selection of real benchmark datasets for comparison:} In order to assess a synthetic dataset, we would like to know if it produces results comparable with real benchmarks. 
To align with the specific aim of a given dataset, we can consider selecting relevant among the following categories of benchmark datasets: general-purpose (e.g., IARPA Janus Benchmark–B (IJB-B)~\cite{whitelam2017iarpa}, IARPA Janus Benchmark–C (IJB-C)~\cite{maze2018iarpa}, LFW~\cite{huang2008lfw}), cross-pose (e.g., CFP-FP~\cite{sengupta2016cfp_fp}, CPLFW~\cite{zheng2018cplfw}), cross-age (e.g., AgeDB~\cite{moschoglou2017agedb}, CALFW~\cite{zheng2017calfw}), racial bias assessment (e.g., RFW~\cite{wang2019rfw}), low-resolution imagery (e.g., TinyFace~\cite{cheng2018low}).


\item[\ding{184}] \textit{Models evaluation on benchmark datasets:}
For real evaluation datasets, we should follow their respective predefined evaluation procedures. An exemplary approach for LFW~\cite{huang2008lfw} uses 6,000 image pairs (3,000 mated, 3,000 non-mated), divided into 10 cross-validation folds to output accuracy for each of the assessed models.
For synthetic model evaluation, we mimic the evaluation procedures established for real datasets. 
While benchmarks usually focus on accuracy, one cannot forget about metrics relevant in the field and prioritize them. In the context of biometrics, alignment with ISO standards and evaluation of FMR100 or FMR1000 can be significantly more important than raw accuracy.

\item[\ding{185}] \textit{Benchmark assessment:} In the last step, we evaluate how reliable synthetic datasets are compared to the real data in terms of metrics computed in the previous step. It is important to note that many sets are serving a particular purpose, and consequently, it is more important for a synthetic dataset to be well aligned with its intended purpose rather than with datasets serving multiple distinct purposes at once. 
Whenever applicable, we propose to compute the Pearson Correlation Coefficient (PCC)~\cite{pearson1896vii} to establish a correlation with each real dataset evaluated. Additionally, one should consider assessing data distribution, as intra-class variability and identity separability, in order to compare it with patterns observed in real data.
Finally, we suggest re-running the evaluation procedure multiple times on different dataset segments to assess consistency and to report the scale of deviations.
\end{enumerate}

Synthetic benchmark datasets are not only an ethical alternative to real data, but also a necessity in times when available data is scarce. An outbreak of COVID-19 resulted in an extensive use of face masks, having a significant impact on the recognition performance rates. For that reason, NIST resorted to evaluations on synthetic data~\cite{ngan2020ongoing, damer2021facemask2021NIST}. Evaluation procedures for scenarios where real data is not available are a complex problem and we believe that they should draw more attention from the scientific community.

\begin{figure}[htbp]
    \centering
    \begin{subfigure}[b]{0.49\textwidth}
        \centering               \includegraphics[width=1.0\textwidth]{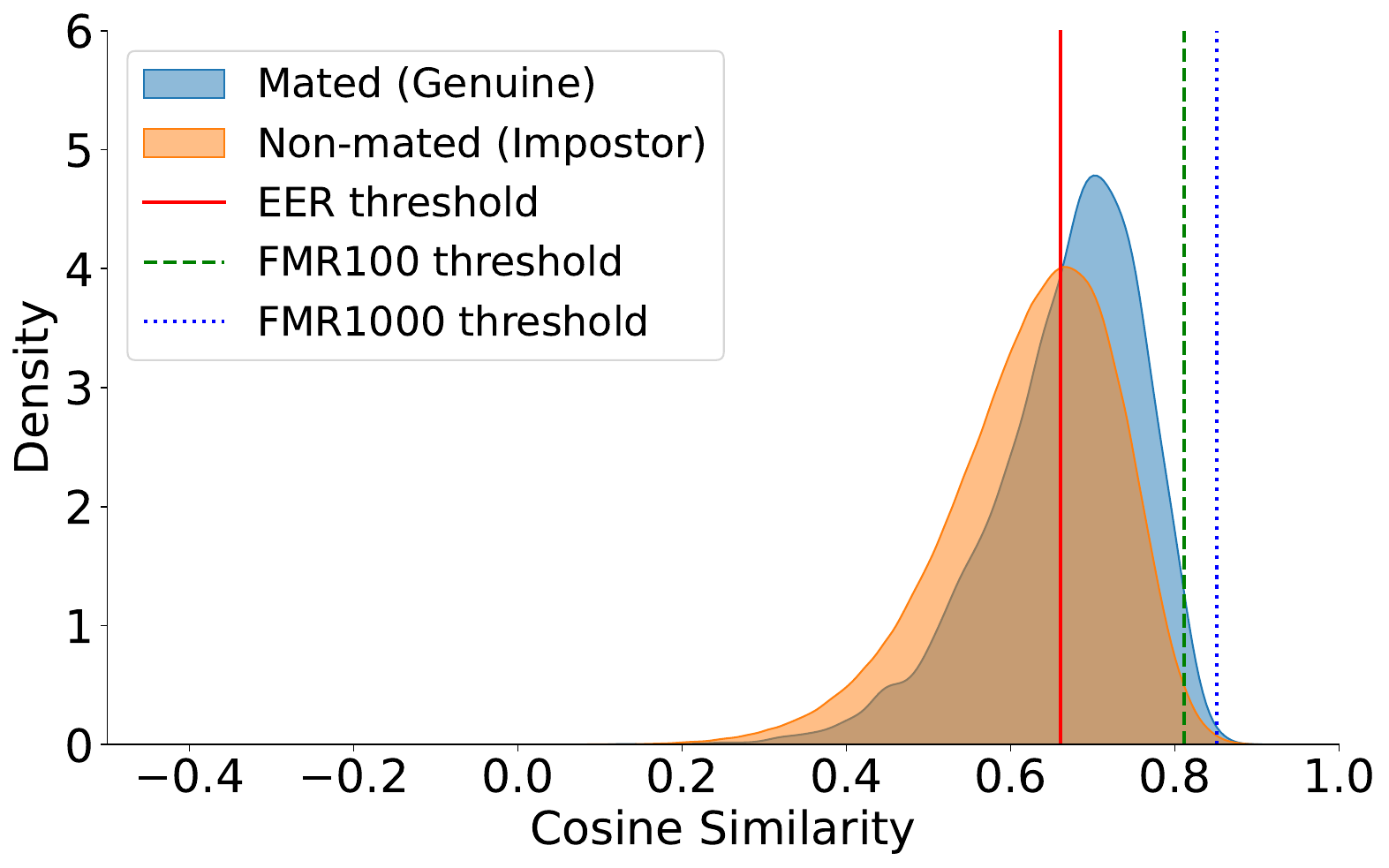} 
        \caption{ControlFace10k}
        \label{fig:ControlFace10k_MvsNM}
    \end{subfigure}
    \hfill
    \begin{subfigure}[b]{0.49\textwidth}
        \centering
        \includegraphics[width=1.0\textwidth]{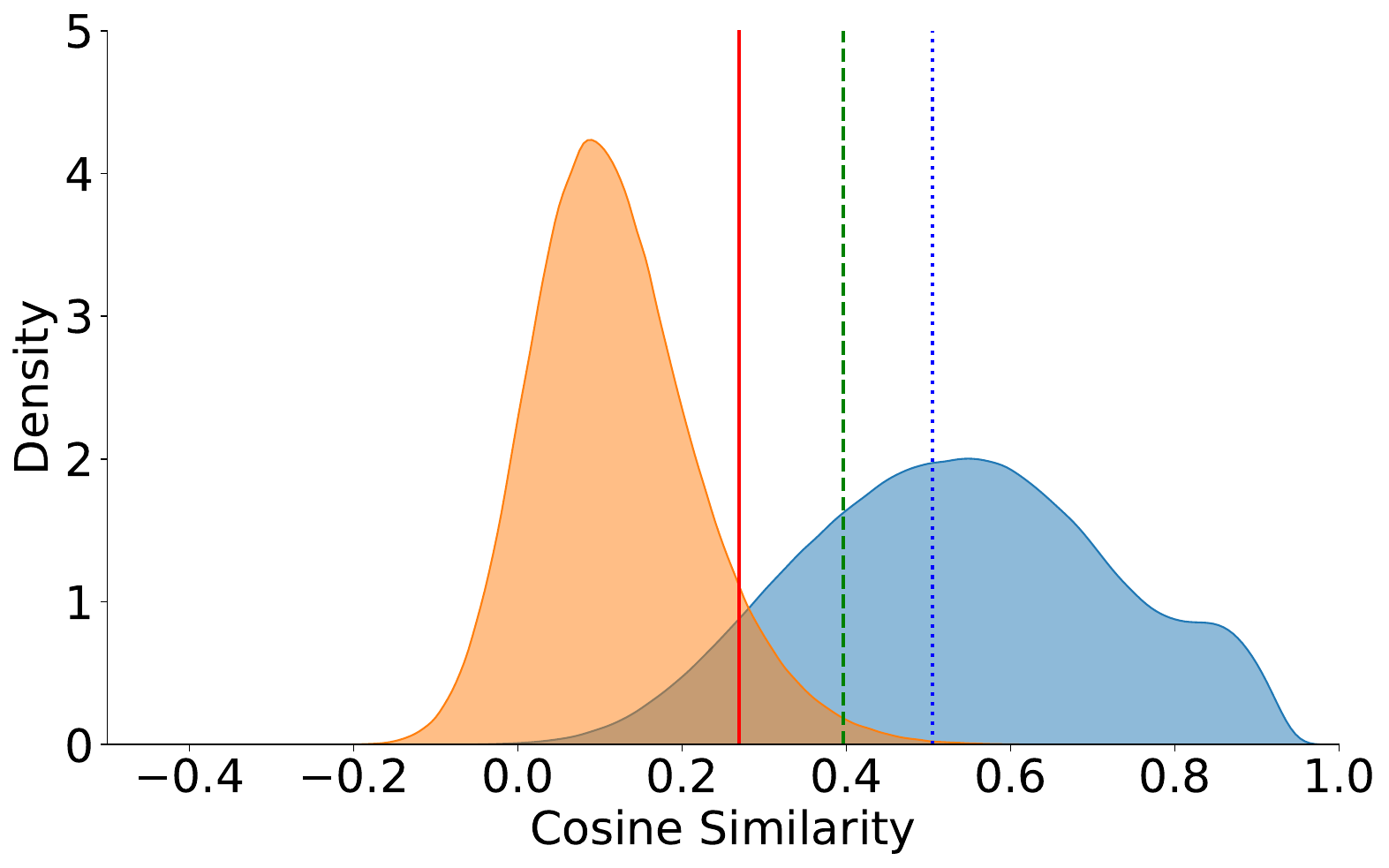}
        \caption{Syn-Multi-PIE}
        \label{fig:Syn-Multi-PIE_MvsNM}
    \end{subfigure}
    \caption{Mated vs non-mated distribution of cosine similarities across benchmark datasets containing more than one sample per identity with thresholds of key biometric indicators marked}
    \label{fig:benchmark_comparison}
\end{figure}

\begin{figure}[htbp]
    \centering
    \begin{subfigure}[b]{0.3\textwidth}
        \centering               \includegraphics[width=1.0\textwidth]{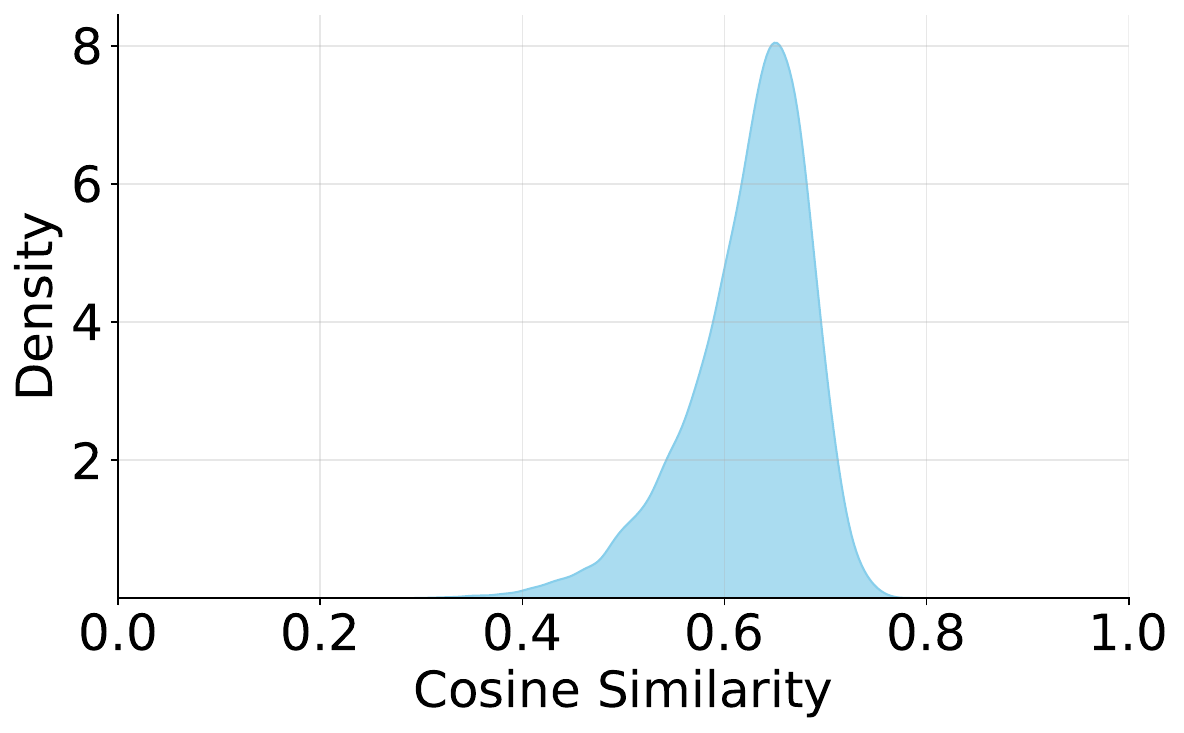} 
        \caption{ControlFace10k}
        \label{fig:CASIA_ControlFace10k}
    \end{subfigure}
    \hfill
    \begin{subfigure}[b]{0.3\textwidth}
        \centering
        \includegraphics[width=1.0\textwidth]{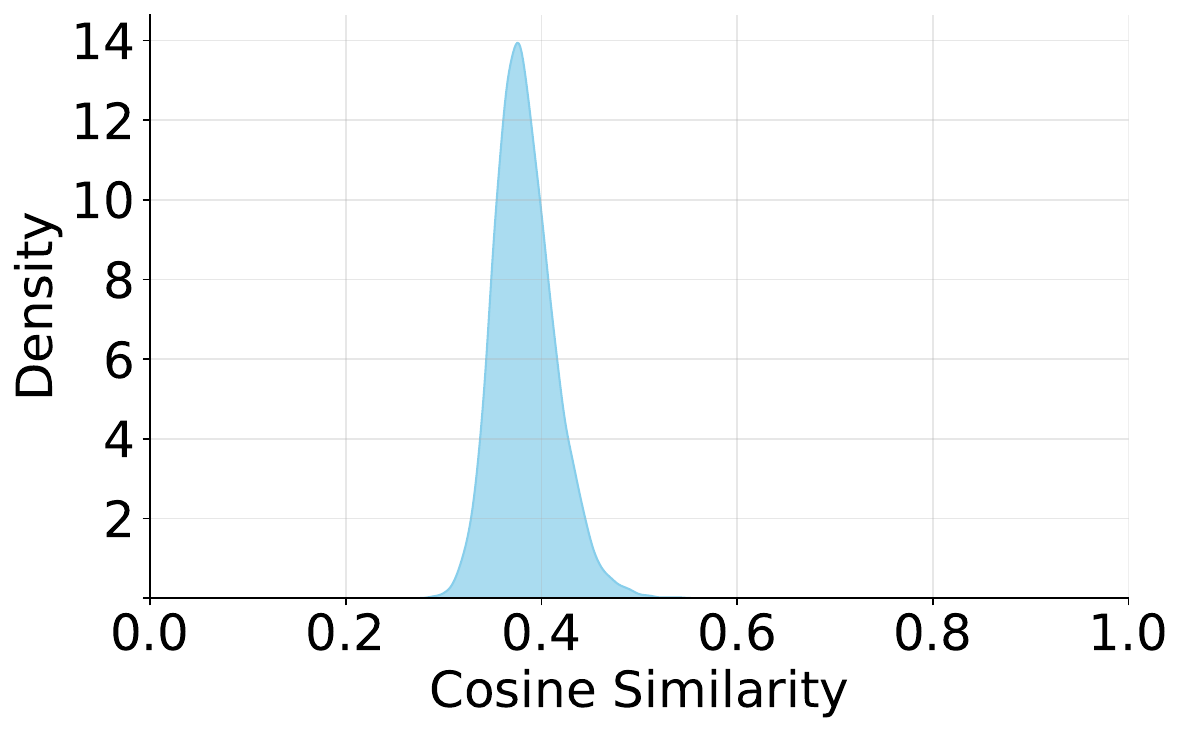}
        \caption{Syn-Multi-PIE}
        \label{fig:CASIA-SynMultiPIE}
    \end{subfigure}
    \hfill
    \begin{subfigure}[b]{0.3\textwidth}
        \centering
        \includegraphics[width=1.0\textwidth]{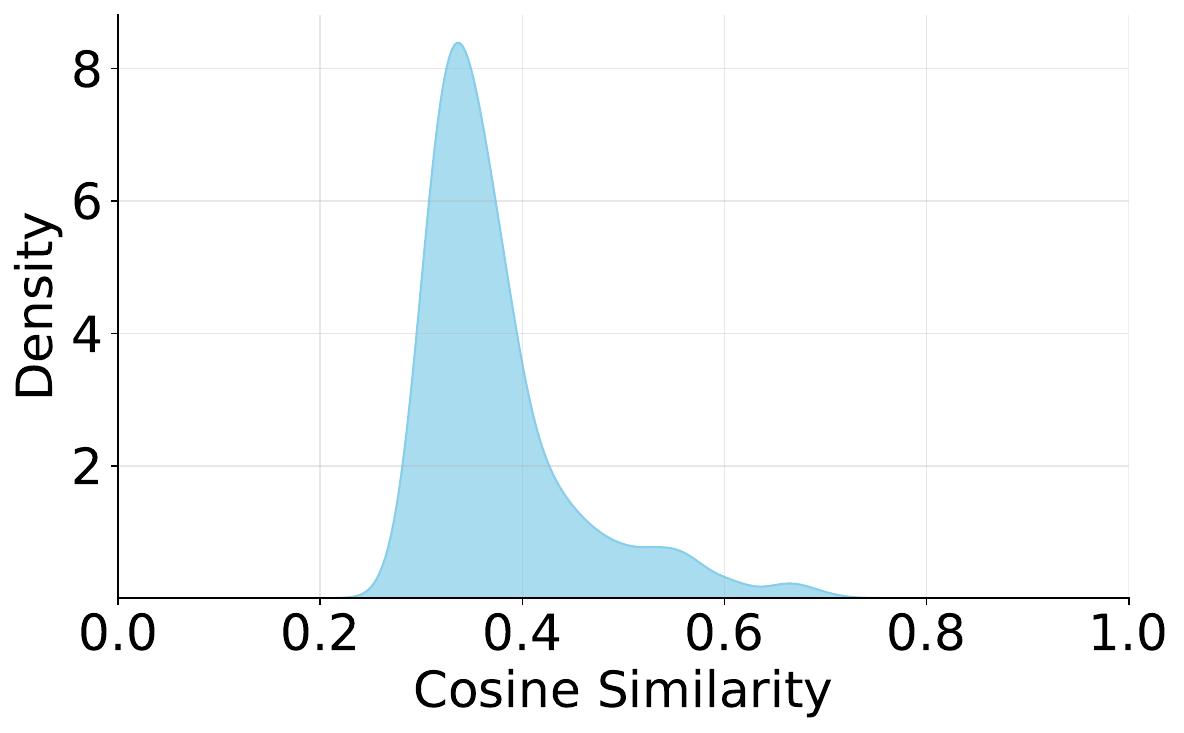}
        \caption{SDFD}
        \label{fig:CASIA_SDFD}
    \end{subfigure}
    \caption{Cosine similarity distribution of closest samples with respect to CASIA-WebFace dataset}
    \label{fig:benchmark_comparison_CASIA}
\end{figure}

Further, we would like to address individual limitations of each of the synthetic benchmark facial recognition datasets. SDFD offers only one image per identity, making it applicable in a limited scope of demographic attribute prediction. It also suffers from long-tail distribution (\Cref{fig:CASIA_SDFD}) when its similarity is compared to CASIA-WebFace~\cite{yi2014casia}, suggesting potential identity leakage. While ControlFace10k claims to address demographic biases, it has an acute problem with identity separability (\Cref{fig:ControlFace10k_MvsNM}), low intra-class variations and it achieves high similarity scores when compared with CASIA-WebFace~\cite{yi2014casia} (\Cref{fig:CASIA_ControlFace10k}).
Syn-Multi-PIE, on the other hand, suggests proper identity separability and intra-class variability (\Cref{fig:Syn-Multi-PIE_MvsNM}), positioning itself as the most reasonable benchmark choice across the 3 available options. Additionally, it offers the lowest risk of identity leakage (\Cref{fig:CASIA-SynMultiPIE}) among benchmark datasets. Nonetheless, the dataset generation procedure fails to explicitly address demographic biases. Overall, synthetic datasets dedicated to bias assessment are another research direction to be explored.

\find{
{\bf\ding{45}  Key take-aways of [R7]}\begin{itemize}[leftmargin=*]
    \item 
    Among currently existing synthetic benchmark datasets, based on evaluated properties, only Syn-Multi-PIE exhibits potential to be used in real-life scenarios. 
    \item To assess the value of each new synthetic evaluation dataset, a proper benchmarking procedure should be defined, allowing for extensive comparison with existing solutions based on real data before release.
    \item The creation of synthetic datasets dedicated to assessing demographic biases 
    presents itself as a compelling research direction.
\end{itemize}
}
\section{Conclusions}
\label{sec:conclusion}
In recent years, we have observed a sudden increase in the interest in synthetic facial recognition datasets. Creation of stricter legal norms concerning one's identity, such as GDPR, and greater interest in ethical aspects of personal data use have prompted researchers in the direction of artificial data.
Properly designed synthetic datasets additionally help to address the shortcomings of real data in the form of label noise and demographic biases.

Importantly, the performance of models trained on synthetic data has risen significantly in recent years, enabling some of the most recent ones - VariFace~\cite{yeung2024variface} and VIGFace~\cite{kim2024vigface} - to outperform a model trained on real CASIA-WebFace~\cite{yi2014casia} and GlintAsian~\cite{an2021glint360k} data, marking an important milestone. 
While synthetic data struggles with problems of low intra-class variation, insufficient identity separability and identity leakage, the highest ranked datasets have introduced measures to actively address those issues. Their effectiveness for MorphFace~\cite{mi2025morphFace}, Vec2Face~\cite{Vec2Face_datasetURL}, and CemiFace~\cite{sun2024cemiface} has been confirmed by our experiments. Consequently, based on the benchmark results and our findings, we are inclined to promote the usage of the aforementioned datasets for facial recognition training.

The weaker branch of synthetic facial recognition datasets, in terms of progress and dataset quality, is evaluation datasets. The scarcity of solutions and the lack of proper testing methodology do not provide enough empirical proof to justify their usage over ethically problematic real data. As such, to enable a fully synthetic recognition training and evaluation pipeline, the domain requires more research into evaluation datasets. Since synthetic generation methods can provide a lot of flexibility in terms of demographic parameters, they can have a potential advantage over real datasets in terms of finding hidden biases. Nonetheless, at the moment we are not convinced that the available synthetic solutions should replace well-established benchmarks such as LFW~\cite{huang2008lfw}, CFP-FP~\cite{sengupta2016cfp_fp}, CPLFW~\cite{zheng2018cplfw}, AgeDB~\cite{moschoglou2017agedb} and CALFW~\cite{zheng2017calfw}.

While we admire the progress made over the past years, our analysis has allowed us to distinguish the following future research directions:
\begin{itemize}
    \item Development of publicly available synthetic dataset with state-of-the-art performance on par with or exceeding benchmark accuracies obtained with real datasets, at the same time actively addressing demographic biases, identity leakage, and using only data acquired with consent during training.
    \item Research into the creation of synthetic facial recognition samples without the use of any real data at all stages of dataset generation.
    \item Research into the viability of retrieving real data from synthetic images via member inference attacks or reverse-engineering techniques.
    \item Generation of synthetic benchmark datasets, either general or targeting specific bias, for which proper performance is confirmed by extensive experiments on actual facial recognition models.
\end{itemize}

Finally, we would like to stress again that synthetic datasets are a unique opportunity to address ethical challenges present in mass-scraped real facial recognition datasets. Currently achieved benchmark and experimental results confirm the possibility of developing viable solutions based on artificial data. 

\section*{Data availability}
Datasets used for the experiments presented in this paper are available for download in the original repositories as indicated by references in~\Cref{tab:datasets_synth}. We do not redistribute those datasets due to their sizes reaching even 200GB, and the fact that some of the datasets require permission to be accessed.

\section*{Code availability}
The code used for the experiments presented in this paper, with explanations regarding its usage, is made publicly available for open science under the following link: 
\begin{center}
\url{https://anonymous.4open.science/r/SyntheticDatasetsReview-925B/README.md}
\end{center}

\section*{Acknowledgments}
\begin{itemize}[leftmargin=*]
    \item The authors acknowledge the use of AI-based tools in this work. Their usage was limited to correcting typos and grammatical errors as well as to rephrasing and refining the final text for readability and clarity.
    \item The experiments presented in this paper were partially carried out using the HPC\footnote{https://hpc.uni.lu} facilities of the University of Luxembourg.
    \item This research was supported by the Luxembourg Army.
\end{itemize}






\section*{Author Contributions}
Concept and experiment design were done by P.B., F.B. and T.F.B.; data collection by P.B. and F.B.;
programming and analysis by P.B.; 
writing and visualization by P.B. and F.B.;
project administration by T.F.B. and J.K.;
P.B., F.B., I.E.O., C.B., W.C.O., T.F.B. and J.K. contributed to editing and review.

\section*{Competing Interests}
The authors declare no competing interests.

\section*{Additional Informaton}
Correspondence and requests for materials should be addressed to
Tegawendé F. Bissyandé.

\end{document}